\documentclass[preprint,12pt]{elsarticle}

\usepackage{amssymb}
\usepackage{amsfonts}       
\usepackage{amsmath, bm, amsthm}
\usepackage{algorithm, algpseudocode}

\usepackage{booktabs}       
\usepackage{multirow}
\usepackage{subfigure}

\makeatletter
\renewcommand{\@thesubfigure}{\hskip\subfiglabelskip}
\makeatother

\graphicspath{{figures/}}

\journal{Elsevier}

\begin{document}

\begin{frontmatter}



\title{Discriminatively Boosted Image Clustering \\ with Fully Convolutional Auto-Encoders}



\author[amss]{Fengfu Li}
\author[ia,brain]{Hong Qiao}
\author[amss]{Bo Zhang} 
\author[ia]{Xuanyang Xi}

\address[amss]{Academy of Mathematics and Systems Science, Chinese Academy of Sciences\\
Beijing 100190, China\\
and\\
School of Mathematical Sciences, University of Academy of Sciences\\
Beijing 100049, China}
\address[ia]{Institute of Automation, Chinese Academy of Sciences, Beijing 100190, China\\
and \\
University of Academy of Sciences, Beijing 100049, China\\}
\address[brain]{CAS Centre for Excellence in Brain Science and Intelligence Technology\\
Shanghai 200031, China}


\begin{abstract}
Traditional image clustering methods take a two-step approach, feature learning and clustering, sequentially. However, recent research results demonstrated that combining the separated phases in a unified framework and training them jointly can achieve a better performance. In this paper, we first introduce fully convolutional auto-encoders for image feature learning and then propose a unified clustering framework to learn image representations and cluster centers jointly based on a fully convolutional auto-encoder and soft $k$-means scores. At initial stages of the learning procedure, the representations extracted from the auto-encoder may not be very discriminative for latter clustering. We address this issue by adopting a boosted discriminative distribution, where high score assignments are highlighted and low score ones are de-emphasized. With the gradually boosted discrimination, clustering assignment scores are discriminated and cluster purities are enlarged. Experiments on several vision benchmark datasets show that our methods can achieve a state-of-the-art performance.
\end{abstract}

\begin{keyword}
image clustering \sep fully convolutional auto-encoder \sep representation learning \sep discriminatively boosted clustering



\end{keyword}

\end{frontmatter}


\section{Introduction}
\label{sec:intro}

Clustering methods are very important techniques for exploratory data analysis with wide applications ranging from
data mining \cite{Han2011DataMining,Berkhin2006DataMiningSurvery}, dimension reduction \cite{Boutsidis2015RDR},
segmentation \cite{Shi2000Ncuts} and so on.
Their aim is to partition data points into clusters so that data in the same cluster are similar
to each other while data in different clusters are dissimilar.
Approaches to achieve this aim include partitional methods such as $k$-means and $k$-medoids, hierarchical
methods like agglomerative clustering and divisive clustering, methods based on density estimation such as
DBSCAN \cite{ester1996density}, and recent methods based on finding density peaks such as
CFSFDP \cite{Rodriguez2016CFSFDP} and LDPS \cite{Li2016LDPS}.

Image clustering \cite{Ahmed2016review} is a special case of clustering analysis that seeks to find compact,
object-level models from many unlabeled images.
Its applications include automatic visual concept discovery \cite{Lee2011easy}, content-based image retrieval
and image annotation. However, image clustering is a hard task mainly owning to the following two reasons:
1) images often are of high dimensionality, which will significantly affect the performance of clustering methods
such as $k$-means \cite{Ding2007AdaptiveDR}, and 2) objects in images usually have two-dimensional or three-dimensional
local structures which should not be ignored when exploring the local structure information of the images.

To address these issues, many representation learning methods have been proposed for image feature extractions
as a preprocessing step. Traditionally, various hand-crafted features such as SIFT \cite{Lowe1999SIFT},
HOG \cite{Dalal2005HOG}, NMF \cite{Hong2016jointNMF}, and (geometric) CW-SSIM
similarity \cite{Sampat2009CWSSIM,Li2016GCWSSIM} have been used to encode the visual information.
Recently, many approaches have been proposed to combine clustering methods with deep neural networks (DNN),
which have shown a remarkable performance improvement over hand-crafted features \cite{Krizhevsky2012DCNN}.
Roughly speaking, these methods can be categorized into two groups: 1) sequential methods that apply clustering
on the learned DNN representations, and 2) unified approaches that jointly optimize the deep representation
learning and clustering objectives.

In the first group, a kind of deep (convolutional) neural networks, such as deep belief network (DBN) \cite{Hinton2006DBN}
and stacked auto-encoders \cite{Tian2014graph}, is first trained in an unsupervised manner to approximate
the non-linear feature embedding from the raw image space to the embedded feature space (usually being low-dimensional).
And then, either $k$-means or spectral clustering or agglomerative clustering can be applied to partition
the feature space. However, since the feature learning and clustering are separated from each other,
the learned DNN features may not be reliable for clustering.

There are a few recent methods in the second group which take the separation issues into consideration.
In \cite{Xie2015DEC}, the authors proposed deep embedded clustering that simultaneously learns feature representations
with stacked auto-encoders and cluster assignments with soft $k$-means by minimizing a joint loss function.
In \cite{Yang2016JULE}, joint unsupervised learning was proposed to learn deep convolutional representations
and agglomerative clustering jointly using a recurrent framework.
In \cite{Liu2016IEC}, the authors proposed an infinite ensemble clustering framework that integrates
deep representation learning and ensemble clustering.
The key insight behind these approaches is that good representations are beneficial for clustering and conversely
clustering results can provide supervisory signals for representation learning.
Thus, two factors, designing a proper representation learning model and designing a suitable unified
learning objective will greatly affect the performance of these methods.

In this paper, we follow recent advances to propose a unified clustering method named Discriminatively Boosted
Clustering (DBC) for image analysis based on fully convolutional auto-encoders (FCAE). See Fig. \ref{fig:dbc}
for a glance of the overall framework. We first introduce a fully convolutional encoder-decoder network for fast
and coarse image feature extraction. We then discard the decoder part and add a soft $k$-means model on top of
the encoder to make a unified clustering model. The model is jointly trained with gradually boosted discrimination
where high score assignments are highlighted and low score ones are de-emphasized. The our main contributions are
summarized as follows:

\begin{itemize}
\item We propose a fully convolutional auto-encoder (FCAE) for image feature learning. The FCAE is composed of
convolution-type layers (convolution and de-convolution layers) and pool-type layers (pooling and un-pooling layers).
By adding batch normalization (BN) layers to each of the convolution-type layers, we can train the FCAE in
an end-to-end way. This avoids the tedious and time-consuming layer-wise pre-training stage adopted in the
traditional stacked (convolutional) auto-encoders. To the best of our knowledge, this is the first attempt to learn
a deep auto-encoder in an end-to-end manner.
\item We propose a discriminatively boosted clustering (DBC) framework based on the learned FCAE and an additional
soft $k$-means model. We train the DBC model in a self-paced learning procedure, where deep representations of
raw images and cluster assignments are jointly learned. This overcomes the separation issue of the traditional
clustering methods that use features directly learned from auto-encoders.
\item We show that the FCAE can learn better features for clustering than raw images on several image datasets
include MNIST, USPS, COIL-20 and COIL-100. Besides, with discriminatively boosted learning, the FCAE based DBC
can outperform several state-of-the-art analogous methods in terms of $k$-means and deep auto-encoder based clustering.
\end{itemize}

The remaining part of this paper is organized as follows. Some related work including stacked (convolutional)
auto-encoders, deconvolutional neural networks, and joint feature learning and clustering are briefly reviewed
in Section \ref{sec:related-work}. Detailed descriptions of the proposed FCAE and DBC are presented in Section \ref{sec:FCAE-DBC}.
Experimental results on several real datasets are given in Section \ref{sec:experiments} to validate the proposed methods.
Conclusions and future works are discussed in Section \ref{sec:conclusion}.

\section{Related work}
\label{sec:related-work}

Stacked auto-encoders \cite{Vincent2010SDAE,Baldi2012auto-encoders,Bengio2013review,Hinton2006DBN,Hinton2006RBM,Bengio2007layerwise}
have been studied in the past years for unsupervised deep feature extraction and nonlinear dimension reduction.
Their extensions for dealing with images are convolutional stacked auto-encoders \cite{Masci2011SCAE,Lee2009CDBN}.
Most of these methods contain a two-stage training procedure \cite{Bengio2007layerwise}: one is layer-wise pre-training
and the other is overall fine-tuning. One of the significant drawbacks of this learning procedure is that the layer-wise
pre-training is time-consuming and tedious, especially when the base layer is a Restricted Boltzmann Machine (RBM)
rather than a traditional auto-encoder or when the overall network is very deep.

Recently, there is an attempt to discard the layer-wise pre-training procedure and train a deep auto-encoder type
network in an end-to-end way. In \cite{Noh2015deconvolution}, a deep deconvolution network is learned for image
segmentation. The input of the architecture is an image and the output is a segmentation mask. The network achieves
the state-of-the-art performance compared with analogous methods thanks to three factors: 1) introducing a
deconvolution layer and a unpooling layer \cite{Zeiler2014visualizing,Mohan2014deconvolution,Zeiler2011deconvolution}
to recover the original image size of the segmentation mask, 2) applying the batch normalization \cite{Ioffe2015BN}
to each convolution layer and each deconvolution layer to reduce the internal covariate shifts, which not only makes
an end-to-end training procedure possible but also speeds up the process, and 3) adopting a pre-trained encoder on
large-scale datasets such as VGG-16 model \cite{Simonyan2015VGG}. The success of the architecture motivates us
that it is possible to design an end-to-end training procedure for fully convolutional auto-encoders.

Clustering has also been studied in the past years based on independent features extracted from auto-encoders
(see, e.g. \cite{Ding2007AdaptiveDR,Tian2014graph,Huang2014DEN,Song2013AEC}). Recently, there are attempts to
combine the auto-encoders and clustering in a unified framework \cite{Xie2015DEC,Yang2016DCN}. In \cite{Xie2015DEC},
the authors proposed Deep Embedded Clustering (DEC) that learns deep representations and cluster assignments jointly.
DEC uses a deep stacked auto-encoder to initialize the feature extraction model and a Kullback-Leibler divergence
loss to fine-tune the unified model. In \cite{Yang2016DCN}, the authors proposed Deep Clustering Network (DCN),
a joint dimensional reduction and $k$-means clustering framework. The dimensional reduction model is based on
deep neural networks. Although these methods have achieved some success, they are not suitable for dealing with
high-dimensional images due to the use of stacked auto-encoders rather than convolutional ones.
This motivates us to design a unified clustering framework based on convolutional auto-encoders.

\section{Proposed methods}
\label{sec:FCAE-DBC}

\begin{figure}[!t]
\centering
\includegraphics[width=1\linewidth]{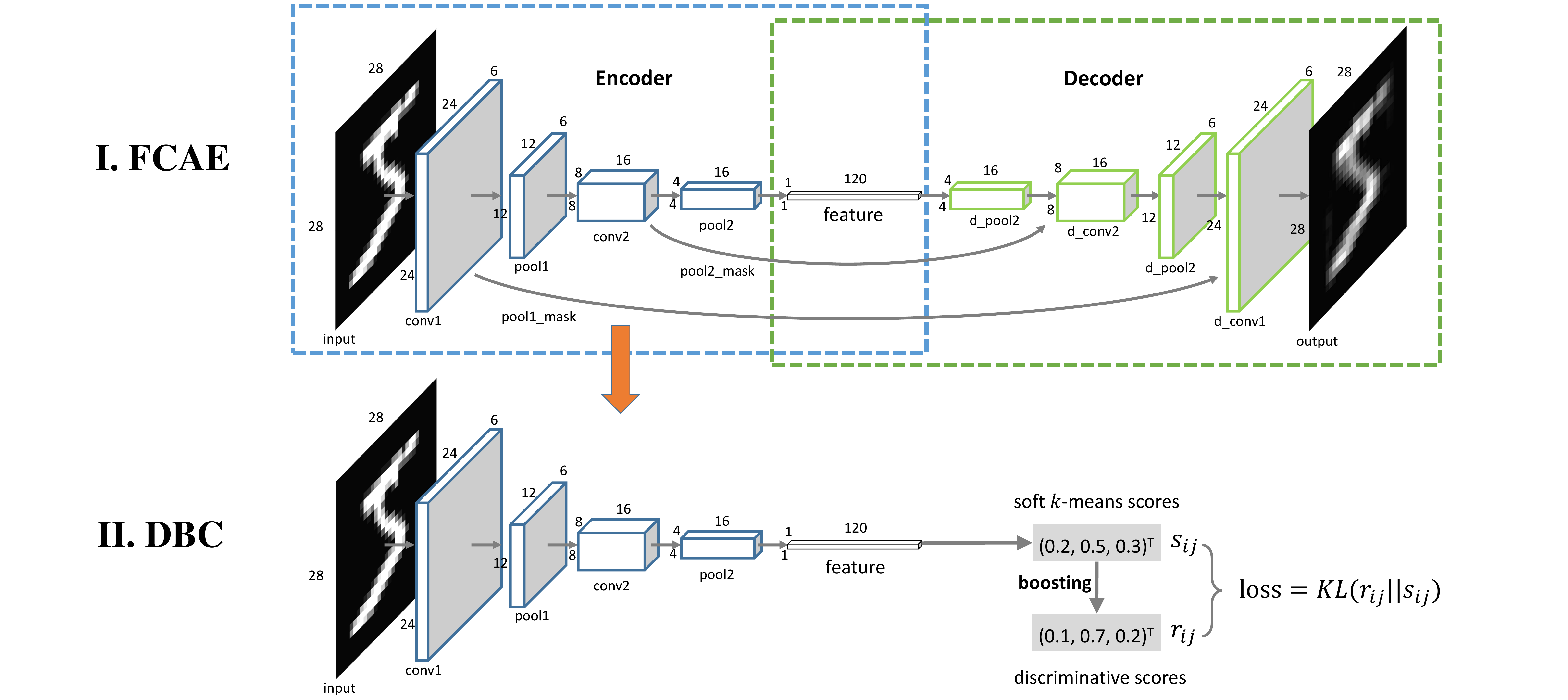}
\caption{A unified image clustering framework. Part I is a fully convolutional auto-encoder (FCAE), and
Part II is a discriminatively boosted clustering (DBC) framework based on the FCAE.}
\label{fig:dbc}
\end{figure}

In this section, we propose a unified image clustering framework with fully convolutional auto-encoders and
a soft $k$-means clustering model (see Fig. \ref{fig:dbc}). The framework contains two parts: part I is a fully
convolutional auto-encoder (FCAE) for fast and coarse image feature extraction, and part II is a discriminatively
boosted clustering (DBC) method which is composed of a fully convolutional encoder and a soft $k$-means categorizer.
The DBC takes an image as input and exports soft assignments as output. It can be jointly trained with a
discriminatively boosted distribution assumption, which makes the learned deep representations more suitable
for the top categorizer. Our idea is very similar to self-paces learning \cite{Lee2011easy}, where easiest
instances are first focused and more complex objects are expanded progressively. In the following subsections,
we will explain the detailed implementation of the idea.

\subsection{Fully convolutional auto-encoder for image feature extraction}

Traditional deep convolutional auto-encoders adopt a greedy layer-wise training procedure for feature transformations.
This could be tedious and time-consuming when dealing with very deep neural networks. To address this issue,
we propose a fully convolutional auto-encoder architecture which can be trained in an end-to-end manner.
Part I of Fig. \ref{fig:dbc} shows an example of FCAE on the MNIST dataset. It has the following features:

\begin{description}
\item[Fully Convolutional]~As pointed out in \cite{Masci2011SCAE}, the max-pooling layers are very crucial
for learning biologically plausible features in the convolutional architectures. Thus, we adopt convolution layers
along with max-pooling layers to make a fully convolutional encoder (FCE). Since the down-sampling operations
in the FCE reduce the size of the output feature maps, we use an unpooling layer introduced
in \cite{Noh2015deconvolution} to recover the feature maps. As a result, the unpooling layers along
with deconvolution layers (see \cite{Noh2015deconvolution}) are adopted to make a fully convolutional decoder (FCD).
\item[Symmetric]~The overall architecture is symmetric around the feature layer. In practice,
it is suggested to design layers of an odd number. Otherwise, it will be ambiguous to define the feature layer.
Besides, fully connected layers (dense layers) should be avoided in the architecture since they destroy the
local structure of the feature layer.
\item[Normalized]~The depth of the whole network grows in $\log$-magnitude as the input image size increases.
This could make the network very deep if the original image has a very large width or height. To overcome this
problem, we adopt the batch normalization (BN) \cite{Ioffe2015BN} strategy for reducing the internal covariate
shift and speeding up the training. The BN operation is performed after each convolutional layer and each
deconvolutional layer except for the last output layer. As pointed out in \cite{Noh2015deconvolution},
BN is critical to optimize the fully convolutional neural networks.
\end{description}

FCAE utilizes the two-dimensional local structure of the input images and reduces the redundancy in parameters
compared with stacked auto-encoders (SAEs). Besides, FCAE differs from conventional SAEs as its weights are
shared among all locations within each feature map and thus preserves the spatial locality.

\subsection{Discriminatively boosted clustering}

\label{sec:dbc}

\begin{figure}[!htb]
\centering
\includegraphics[width=0.9\linewidth]{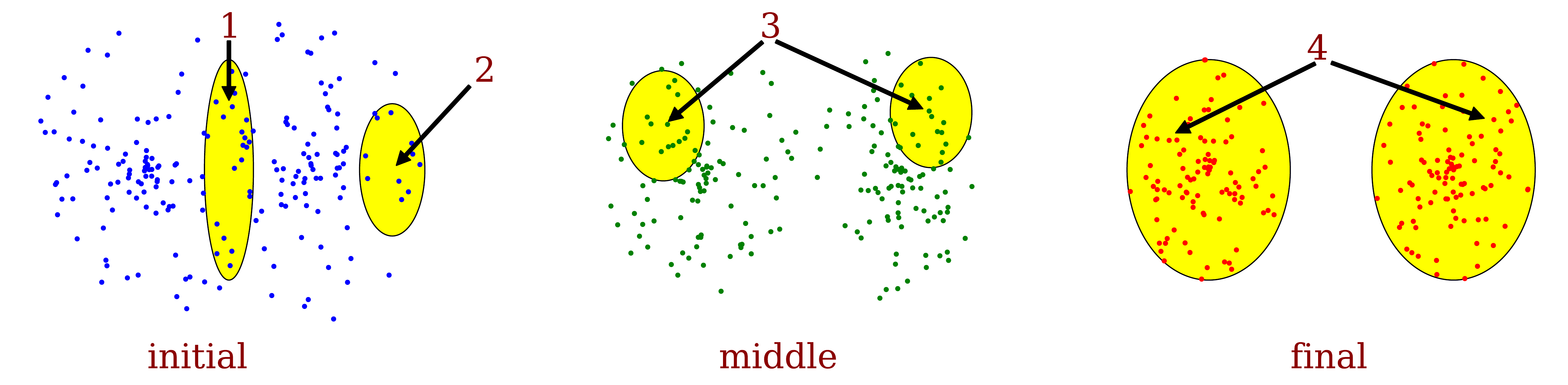}
\caption{Learning procedure of DBC.}
\label{fig:procedure}
\end{figure}

Once FCAE has been trained, we can extract features with the encoder part to serve as the input of a categorizer.
This strategy is used in many clustering methods based on auto-encoders, such as GraphEncoder \cite{Tian2014graph},
deep embedding networks \cite{Huang2014DEN}, and auto-encoder based clustering \cite{Song2013AEC}.
These approaches treat the auto-encoder as a preprocessing step which is separately designed from the latter
clustering step. However, the representations learned in this way could be amphibolous for clustering,
and the clusters may be unclear (see the initial stage in Fig. \ref{fig:procedure})).

To address this issue, we propose a self-paced approach to make feature learning and clustering in a unified
framework (see Part II in Fig. \ref{fig:dbc}). We throw away the decoder of the FACE and add a soft $k$-means
model on top of the feature layer. To train the unified model, we trust easier samples first and then gradually
utilize new samples with the increasing complexity. Here, an \textit{easier} sample (see the regions labelled 2, 3
and 4 in Fig. \ref{fig:procedure}) is much certain to belong to a specific cluster, and a \textit{harder} sample
(see the region 1 in Fig. \ref{fig:procedure}) is very likely to be categorized to multiple clusters.
Fig. \ref{fig:procedure} describes the difference between these samples at a different learning stage of DBC.

There are three challenging questions in the learning problem of DBC which will be answered in the following subsections:
\begin{enumerate}
\item How to choose a proper criterion to determine the easiness or hardness of a sample?
\item How to transform harder samples into easier ones?
\item How to learn from easier samples?
\end{enumerate}

\subsubsection{Easiness measurement with the soft $k$-means scores}

We follow DEC \cite{Xie2015DEC} to adopt the $t$-distribution-based soft assignment to measure the easiness of a sample.
The $t$-distribution is investigated in \cite{Maaten2008tSNE} to deal with the crowding problem of low-dimensional data distributions. Under the $t$-distribution kernel, the soft score (or similarity) between the
feature $z_i$ ($i\in 1,2,\ldots,m$) and the cluster center $\mu_j$ ($j\in 1,2,\ldots,k$) is
\begin{eqnarray}
s_{ij} &\propto& \big(1+\frac{||z_i-\mu_j||^2}{v}\big)^{-\frac{v+1}{2}} \label{eq:S} \\
&\text{s.t.}& \sum_{j=1}^k s_{ij} = 1 \nonumber
\end{eqnarray}
Here, $v$ is the degree of freedom of the $t$-distribution and set to be $1$ in practice. The most important
reason for choosing the $t$-distribution kernel is that it has a longer tail than the famous heat kernel
(or the Gaussian-distribution kernel). Thus, we do not need to pay much attention to the parameter estimation (see \cite{Maaten2008tSNE}), which is a hard task in unsupervised learning.

\subsubsection{Boosting easiness with discriminative target distribution}

We transform the harder examples to the easier ones by boosting the higher score assignments
and, meanwhile, bring down those with lower scores. This can be achieved by constructing an underlying
target distribution $r_{ij}$ from $s_{ij}$ as follows:
\begin{eqnarray}\label{eq:R}
r_{ij} &\propto& s_{ij}^{\alpha},~\alpha > 1 \\
&\text{s.t.}& \sum_{j=1}^k r_{ij} = 1 \nonumber
\end{eqnarray}

Suppose we can ideally learn from the soft scores (denoted as $S$) to the assumptive distribution
(denoted as $R$) each time. Then we can generate a learning chain as follows:
$$S^{(0)} \rightarrow R^{(0)} = S^{(1)} \rightarrow R^{(1)} = S^{(2)}\rightarrow \cdots.$$
The following two properties can be observed from the chain:

\textbf{Property 1} If $s_{ij}^{(0)}=s_{ij'}^{(0)}$ for any $j$ and $j'$, then $s_{ij}^{(t)}\equiv{1}/{k}$
for all $j$ and all time step $t$.

\textbf{Proof} Under the condition, and by (\ref{eq:R}) we can deduce that $r_{ij}^{(0)}\equiv r_{ij'}^{(0)}$.
By the chain this is equivalent to the fact that $s_{ij}^{(1)}\equiv s_{ij'}^{(1)}$.
Thus, the conclusion $s_{ij}^{(t)}\equiv s_{ij'}^{(t)}$ follows recursively for all $t$. \qed

\textbf{Property 2} If there exists an $l$ such that $s_{il}^{(0)}>\mathop{\max}\limits_{j\neq l}s_{ij}^{(0)}$, then
\[
\mathop{\text{limit}}_{t\rightarrow\infty} s_{ij}^{(t)} =
\begin{cases}
1 & \text{if}~ j = l \\
0 & \text{if}~ j \neq l
\end{cases}
\]

\textbf{Proof} By (\ref{eq:R}) we have
$$
\frac{s_{ij}^{(t)}}{s_{il}^{(t)}}=\Big[\frac{s_{ij}^{(t-1)}}{s_{il}^{(t-1)}}\Big]^{\alpha}
=\Big[\frac{s_{ij}^{(t-2)}}{s_{il}^{(t-2)}}\Big]^{\alpha^2}
=\cdots=\Big[\frac{s_{ij}^{(0)}}{s_{il}^{(0)}}\Big]^{\alpha^t}.
$$
By the assumption $s_{il}^{(0)}>\mathop{\max}\limits_{j\neq l}s_{ij}^{(0)}$, it is seen that
${s_{ij}^{(0)}}/{s_{il}^{(0)}}<1$ for any $j\neq l$. On the other hand, since $\alpha>1$, we have $\mathop{\text{limit}}\limits_{t\rightarrow\infty}\alpha^t = \infty$. Thus,
$$
\mathop{\text{limit}}_{t\rightarrow\infty}\frac{s_{ij}^{(t)}}{s_{il}^{(t)}}
=\mathop{\text{limit}}_{\alpha^t\rightarrow\infty}\Big[\frac{s_{ij}^{(0)}}{s_{il}^{(0)}}\Big]^{\alpha^t}
=0,\;\;\forall j\neq l.
$$
Since $s_{il}^{(t)}$ is finite, we have $\mathop{\text{limit}}\limits_{t\rightarrow\infty}s_{ij}^{(t)}=0,$
$\forall j\neq l$. Finally, with the constrains $\sum_{j=1}^k s_{ij}^{(t)} = 1$, we obtain
$$
\mathop{\text{limit}}_{t\rightarrow\infty}s_{il}^{(t)}= 1
-\sum_{j\neq l}\mathop{\text{limit}}_{t\rightarrow \infty} s_{ij}^{(t)} = 1.
\qed
$$

\textbf{Property 1} tells us that the \textit{hardest} sample (which has the equal probability to be assigned
to different clusters) would always be the hardest one. However, in practical applications, there can hardly
exist such examples. \textbf{Property 2} shows that the initial non-discriminative samples could be boosted
gradually to be definitely discriminative. As a result, we get the desired features for $k$-means clustering.

Note that the boosting factor $\alpha$ controls the speed of the learning process. A larger $\alpha$ can make
the learning process more quickly than smaller ones. However, it may boost some falsely categorized samples
too quickly at initial stages and thus makes their features irrecoverable at later stages.

Besides, it can be helpful to balance the data distribution at different learning stages. In \cite{Xie2015DEC},
the authors proposed to normalize the boosted assignments to prevent large clusters from distorting the hidden
feature space. This problem can be overcome by dividing a normalization factor $n_j=\sum_{i=1}^m r_{ij}$ for
each of the $r_{ij}$.

\subsubsection{Learning with the Kullback-Leibler divergence loss}

\begin{figure}[!bht]
\centering
\includegraphics[width=0.5\linewidth]{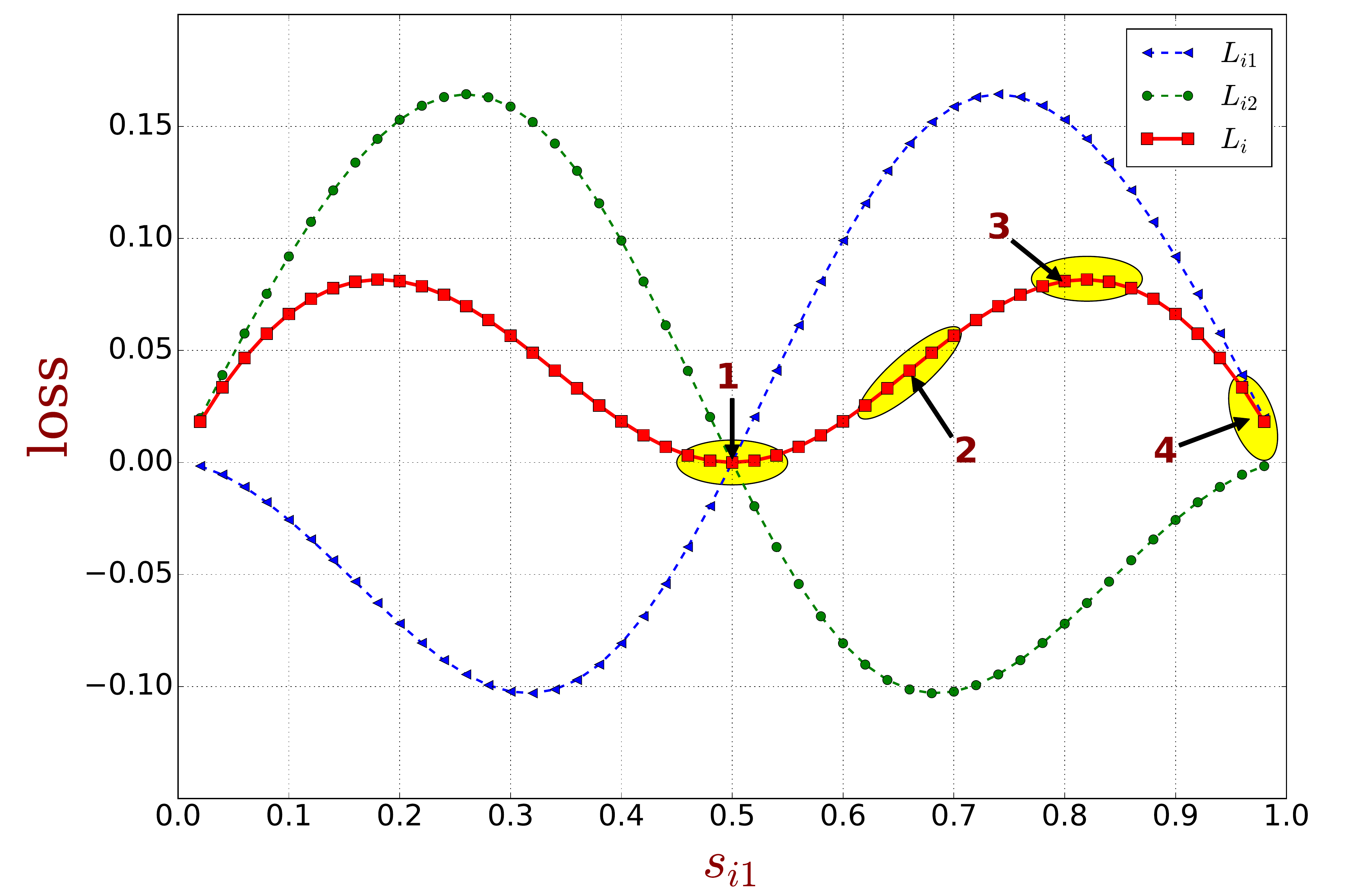}
\caption{KL divergence loss with respect to the soft scores assigned to the first cluster.
Here we assume that there are $2$ clusters, so $0.5$ is a random guess probability.
}\label{fig:kl}
\end{figure}

In the last subsection, it was assumed that we could learn from $s_{ij}$ to the boosted target distribution $r_{ij}$.
This aim can be achieved with a joint Kullback-Leibler (KL) divergence loss, that is,
\begin{equation}\label{eq:kl-loss}
(\theta^*,\mu^*)=\mathop{\arg\min}\limits_{\theta,~\mu}L=\text{KL}(R||S)=\sum_{i=1}^m\sum_{j=1}^k r_{ij}\log \frac{r_{ij}}{s_{ij}}.
\end{equation}
Fig. \ref{fig:kl} gives an example of the joint loss when $k=2$, where $L_{ij}=r_{ij}\log({r_{ij}}/{s_{ij}})$
is the loss generated by the sample $x_i$ with respect to the $j$th cluster ($j=1$ or $2$). Regions marked
in Fig. \ref{fig:kl} roughly correspond to the regions marked in Fig. \ref{fig:procedure}.

Intuitively, the loss has the following main features:
\begin{itemize}
\item For an ambiguous (or hard) sample (i.e., $s_{ij}\approx s_{il}, \forall j,l$), its loss
$L_i=\sum_{j=1}^{k}r_{ij}\log({r_{ij}}/{s_{ij}})\approx 0$ according to \textbf{Property 1}.
Therefore, it will not be seriously treated in the learning process. (Region 1)
\item For a good categorized sample (i.e., there exists an $l$ such that $1\gg s_{il}>\max_{j\neq l}s_{ij}$),
its loss will be much greater than zero, and thus it will be treated more seriously. (Regions 2 and 3)
\item For a definitely well categorized sample (i.e., there exists an $l$ such that
$1\approx s_{il}\gg\max_{j\neq l}s_{ij}$), its loss will be near zero. This means that its features do not
need to be changed much more. (Region 4)
\end{itemize}

By (\ref{eq:S})-(\ref{eq:kl-loss}), the gradients of the KL divergence loss w.r.t. $z_i$ and $\mu_j$ can be
deduced as follows:
\begin{eqnarray}\label{eq:gradient-z}
\frac{\partial L}{\partial z_i}=\frac{1+v}{v}\sum_{j=1}^k(r_{ij}-s_{ij})\frac{z_i-\mu_j}{1+||z_i-\mu_j||^2/v}
\end{eqnarray}
and
\begin{eqnarray}\label{eq:gradient-mu}
\frac{\partial L}{\partial\mu_j}=\frac{1+v}{v}\sum_{i=1}^m (r_{ij}-s_{ij})\frac{\mu_j-z_i}{1+||\mu_j-z_i||^2/v}
\end{eqnarray}

The derivation of (\ref{eq:gradient-z}) and (\ref{eq:gradient-mu}) can be found in the appendix.

\subsubsection{Training algorithm}

In this section, we summarize the overall training procedure of the proposed method in Algorithm \ref{alg:DBC-I} and Algorithm \ref{alg:DBC-II}. They implement the framework showed in Fig. \ref{fig:dbc}. Here $T$ is the maximum learning
epochs, $B$ is the maximum updating iterations in each epoch and $m_b$ is the mini-batch size.
The encoder part of FCAE is $f$:~$x\rightarrow z$, which is parameterized by $\theta_\text{e}$ and
the decoder part of FCAE is $g$:~$z\rightarrow\hat{x}$, which is parameterized by $\theta_\text{d}$.

\begin{algorithm}[!tbh]
\caption{Discriminatively Boosted Clustering (DBC)}
\label{alg:DBC-I}
\begin{algorithmic}[1]
\Require $X$, $T$, $B$, $m_b$, $\alpha$, $k$
\Ensure $\theta$ and $\mu$
\Statex //\textbf{Stage I: Train a FCAE and clustering with its features}
\State Train a deep fully convolutional auto-encoder
\begin{equation}\label{eq:CAE}
x_i\overset{\theta_{\text{e}}}{\longrightarrow}z_i(\text{features})
\overset{\theta_{\text{d}}}{\longrightarrow}\hat{x}_i\tag{M1}
\end{equation}
with the Euclidian loss
$$
(\theta_{\text{e}}^*,\theta_{\text{d}}^*)=\mathop{\arg\min}\limits_{\theta_{\text{e}},~\theta_{\text{d}}}
\sum_{i=1}^m||x_i-\hat{x}_i||^2_2=\sum_{i=1}^m ||x_i-g_{\theta_{\text{d}}}(f_{\theta_{\text{e}}}(x_i))||^2_2
$$
by using the traditional error back-propagation algorithm.
\State Extract features: $Z\leftarrow f_{\theta_e^*}(X)$
\State Clustering with the features: $\mu_z\leftarrow$ $k$-means centers
\algstore{dbc}
\end{algorithmic}
\end{algorithm}

\begin{algorithm}[!tbh]
\caption{DBC (Continued)}
\label{alg:DBC-II}
\begin{algorithmic}[1]
\algrestore{dbc}
\Statex //\textbf{Stage II: Jointly learn the FCE and cluster centers}
\State Construct a unified clustering model with encoder parameters $\theta$ and cluster centers $\mu$
\begin{equation}\label{eq:DBCM}
x_i \overset{\theta}{\longrightarrow} z_i \overset{\mu_j}{\longrightarrow} s_{ij}\tag{M2}
\end{equation}
\State Initialization: $\theta\leftarrow \theta_{\text{e}}^*$, $\mu\leftarrow \mu_z$
\For{$t$ = 1 to $T$}
\State Forward propagate (\ref{eq:DBCM}) and update the \textit{soft} assignments
$$
s_{ij}\leftarrow \frac{(1+||z_i-\mu_j||^2)^{-1}}{\sum_{j=1}^{k}(1+||z_i-\mu_j||^2)^{-1}},~\text{where}~z_i=f_\theta(x_i).
$$
\State Update the target distribution
$$
r_{ij}\leftarrow \frac{s_{ij}^\alpha/n_j}{\sum_{j=1}^{k}s_{ij}^\alpha/n_j},~\text{where}~n_j=\sum_{i=1}^{m}s_{ij}^\alpha.
$$
\For{$b$ = 1 to $B$}
\State Forward propagate (\ref{eq:DBCM}) with a mini-batch of $m_b$ samples.
\State Backward propagate (\ref{eq:DBCM}) from (\ref{eq:gradient-z}) and (\ref{eq:gradient-mu})
to get ${\partial L}/{\partial\theta}$ and ${\partial L}/{\partial \mu}$.
\State Update $\theta$ and $\mu$ with the gradients.
\EndFor
\State Stop if \textit{hard} assignments remain unchanged.
\EndFor
\end{algorithmic}
\end{algorithm}

\section{Experiments}
\label{sec:experiments}

In this section, we present experimental results on several real datasets to evaluate the proposed methods
by comparing with several state-of-the-art methods. To this end, we first introduce several evaluation benchmarks
and then present visualization results of the inner features, the learned FCAE weights, the frequency hist of
soft assignments during the learning process and the features embedded in a low-dimensional space.
We will also give some ablation studies with respect to the boosting factor $\alpha$, the normalization factor $n_j$
and the FCAE initializations.

\subsection{Evaluation benchmarks}

\textbf{Datasets} We evaluate the proposed FCAE and DBC methods on two hand-written digit image datasets
(MNIST \footnote{http://yann.lecun.com/exdb/mnist/} and USPS \footnote{http://www.cs.nyu.edu/~roweis/data.html})
and two multi-view object image datasets (COIL-20 \footnote{http://www.cs.columbia.edu/CAVE/software/softlib/coil-20.php}
and COIL-100 \footnote{http://www.cs.columbia.edu/CAVE/software/softlib/coil-100.php}).
The size of the datasets, the number of categories, the image sizes and the number of channels are summarized
in Table \ref{tab:dataset}.

\begin{table}[!bht]
\begin{scriptsize}
\caption{Datasets used in our experiments.}
\label{tab:dataset}
\begin{center}
\begin{tabular}{cccccc}
\toprule
Dataset & \#Samples & \#Categories & Image Size & \#Channels \\
\midrule
MNIST       & 70000 & 10 & 28$\times$28 & 1 \\
USPS        & 11000 & 10 & 16$\times$16 & 1 \\
COIL-20     & 1440  & 20 & 128$\times$128 & 1 \\
COIL-100    & 7200  & 100 & 128$\times$128 & 3 \\
\bottomrule
\end{tabular}
\end{center}
\end{scriptsize}
\end{table}

\textbf{Evaluation metrics} Two standard metrics are used to evaluate the experiment results explained as follows.
\begin{itemize}
\item Accuracy (ACC) \cite{Xie2015DEC}. Given the ground truth labels $\{c_i|1\leq i\leq m\}$ and the predicted
assignments $\{\hat{c}_i| 1\leq i\leq m\}$, ACC measures the average accuracy:
$$
\text{ACC}(\hat{c},c)=\max_g\frac{1}{m} \sum_{i=1}^{m} \textbf{1}\{c_i=g(\hat{c}_i)\}
$$
where $g$ ranges over all possible one-to-one mappings between the labels of the predicted clusters and
the ground truth labels. The optimal mapping can be efficiently computed using
the Hungarian algorithm \cite{Kuhn1995Hungarian}.
\item Normalized mutual information (NMI) \cite{Cai2011NMI}. From the information theory point of view
NMI can be interpreted as
$$
\text{NMI}(\hat{c}, c)=\frac{\text{MI}(\hat{c},c)}{\max(\text{H}(\hat{c}),\text{H}(c))}
$$
where $\text{H}(c)$ is the entropy of $c$ and $\text{NMI}(\hat{c},c)$ is the mutual information of $\hat{c}$ and $c$.
\end{itemize}

\textbf{Network architectures}
Table \ref{tab:net-architecture} shows the network architecture of the encoder parts with respect to different datasets.
The decoder parts are totally reversed by the encoder parts. We use max-pooling in all the experiments.
The size of all the feature layers is $1\times1$. No padding is used in the convolutional layers except
for the USPS dataset whose padding size is $1.$

\begin{table}[!htb]
\caption{Detailed configuration of the network architecture of the convolutional encoder.
The first rows are the filter sizes of the corresponding layer (filter size or pooling size, \#filters).
The second rows are the output sizes (feature map size, \#channels).
}\label{tab:net-architecture}
\begin{center}
\begin{scriptsize}
\begin{tabular}{cccccccccc}
\toprule
datasets & conv1 & pool1 & conv2 & pool2 & conv3 & pool3 & conv4 & pool4 & features \\ \midrule
\multirow{2}{*}{MNIST}  & 5,~6 & 2,~-  & 5,~16  & 2,~- & - & - & - & - & 4,~120 \\
 & 24,~6 & 12,~6 & 8,~16 & 4,~16 & - & - & - & - & 1,~120 \\ \midrule
\multirow{2}{*}{USPS} & 3,~20 & 2,~- & 3,~20  & 2,~- & - & - & - & - & 4,~160 \\
 & 16,20 & 8,~20 & 8,~20 & 4,~20 & - & - & - & - & 1,~160 \\ \midrule
\multirow{2}{*}{COIL}  & 9,~20 & 2,~- & 5,~20  & 2,~- & 5,~20 & 2,~- & 5,40 & 2,~- & 4,~320 \\
 & 120,~20 & 60,~20 & 56,~20 & 28,~20 & 24,~20 & 12,~20 & 8,~40 & 4,~40 & 1,~320\\
\bottomrule
\end{tabular}
\end{scriptsize}
\end{center}
\end{table}

\textbf{The comparing methods} To validate the effectiveness of FCAE and DBC, we compare them with the following
state-of-the-art methods in terms of the $k$-means and deep auto-encoders based clustering.
\begin{itemize}
\item \textbf{KMS} is the baseline method that applies the $k$-means algorithm on raw images.
\item \textbf{DAE-KMS} \cite{Xie2015DEC} uses deep auto-encoders for feature extraction and then
     applies $k$-means for later clustering.
\item \textbf{AEC} \cite{Song2013AEC} is a variant of DAE-KMS that simultaneously optimizes
   the data reconstruction error and representation compactness.
\item \textbf{IEC} \cite{Liu2016IEC} incorporates the deep representation learning and ensemble clustering.
\item \textbf{DEC} \cite{Xie2015DEC} simultaneously learns the feature representations and cluster centers
    using deep auto-encoders and soft $k$-means, respectively.
\item \textbf{DEN} \cite{Huang2014DEN} learns the clustering-oriented representations by utilizing
    deep auto-encoders and manifold constraints.
\item \textbf{DCN} \cite{Yang2016DCN} jointly applies dimensionality reduction and $k$-means clustering.
\item \textbf{FCAE-KMS} (our algorithm) adopts FCAE for feature extraction and applies $k$-means for
     the latter clustering.
\item \textbf{DBC} (our algorithm) uses Algorithm \ref{alg:DBC} for training a unified clustering method.
\end{itemize}

\begin{table}[!b]
\begin{center}
\begin{scriptsize}
\caption{Clustering performance on MNIST.}
\text{~~~~~~~~~~~~~~~~~~~~~~~~~~~~~~~~~~~~~~~~}
\label{tab:benchmarks-mnist}
\begin{tabular}{cccccccccc} \toprule
Metric & KMS & DAE-KMS  & AEC & IEC & DCN  & DEC & FCAE-KMS & DBC \\
\midrule
ACC & 0.535 &0.818 & 0.760 & 0.609 & 0.58 / 0.93\footnote{DCN with pre-processed MNIST.}&0.843 & 0.794&\textbf{0.964}\\
NMI & 0.531 & - & 0.669 & 0.542 & 0.63 / 0.85~ & - & 0.698 & \textbf{0.917} \\
\bottomrule
\end{tabular}
\end{scriptsize}
\end{center}
\noindent\rule{4cm}{0.4pt} \\
\begin{footnotesize}
\text{~~}${}^5$DCN with processed MNIST.
\end{footnotesize}
\end{table}

\textbf{Results and analysis} Table \ref{tab:benchmarks-mnist} summarizes the benchmark results on the MNIST dataset.
The $k$-means method performs badly on raw images. However, based on the end-to-end trained FACE features, $
k$-means can achieve comparative results compared with DAE-KMS which uses greedily layer-wise trained
deep auto-encoder features. Moreover, with an additional joint training, DBC outperforms FCAE-KMS and beats
all the other comparing methods in terms of ACC and NMI.

Tables \ref{tab:benchmarks-usps}-\ref{tab:benchmarks-coil-100} show the benchmarks on USPS, COIL-20 and COIL-100,
respectively. Similarly to the observations on the MNIST hand-writeen digits dataset, DBC outperforms FCAE-KMS
by a large margin on the USPS hand-written digits dataset. On the COIL sets, DBC obtained a little better results
than FCAE-KMS did.

On the hand-written digits datasets, the number of samples is much larger than the number of categories.
This results in the distribution of the FCAE features to be closely related, and lots of ambiguous samples may occur.
As a result, discriminatively boosting makes sense on these datasets. Thus, there is no doubt that DBC performs
much better than FCAE-KMS. On the COIL sets, DBC takes little advantage of the discriminatively boosting procedure
since the FCAE features are very helpful for clustering. Thus, there are very few ambiguous samples whose easiness
needs to be boosted.

\begin{table}[!ht]
\centering
\begin{scriptsize}
\caption{Clustering performance on USPS.}
\label{tab:benchmarks-usps}
\begin{center}
\begin{tabular}{cccccc} \toprule
Metric & KMS & AEC & IEC & FCAE-KMS & DBC \\
\midrule
ACC & 0.535 & 0.715 & \textbf{0.767} & 0.667 & 0.743 \\
NMI & 0.531 & 0.651 & 0.641 & 0.645 & \textbf{0.724} \\
\bottomrule
\end{tabular}
\end{center}
\end{scriptsize}
\end{table}

\begin{table}[!ht]
\centering
\begin{scriptsize}
\caption{Clustering performance on COIL-20.}
\label{tab:benchmarks-coil}
\begin{center}
\begin{tabular}{ccccc} \toprule
Metric & KMS & DEN  & FCAE-KMS & DBC \\
\midrule
ACC & 0.592 & 0.725 & 0.787 & \textbf{0.793} \\
NMI & 0.767 & 0.870 & 0.882 & \textbf{0.895} \\
\bottomrule
\end{tabular}
\end{center}
\end{scriptsize}
\end{table}

\begin{table}[!ht]
\centering
\begin{scriptsize}
\caption{Clustering performance on COIL-100.}
\label{tab:benchmarks-coil-100}
\begin{center}
\begin{tabular}{ccccc} \toprule
Metric & KMS & IEC  & FCAE-KMS & DBC \\
\midrule
ACC & 0.506 & 0.546 & 0.766 & \textbf{0.775} \\
NMI & 0.772 & 0.787 & 0.897 & \textbf{0.905} \\
\bottomrule
\end{tabular}
\end{center}
\end{scriptsize}
\end{table}

\subsection{Visualization}

One of the advantages of fully convolutional neural networks is that we can naturally visualize the inner activations (or features) and the trained weights (or filters) in a two-dimensional space \cite{Masci2011SCAE}. Besides, we can monitor
the learning process of DBC by drawing frequency hists of assignment scores. In addition, $t$-SNE can be applied to
the embedded features to visualize the manifold structures in a low-dimensional space. Finally, we show some typical
falsely categorized samples generated by our algorithm.

\subsubsection{Visualization of the inner activations and learned filters}

\begin{figure}[!htb]
\subfigure{\includegraphics[width=0.083\textwidth]{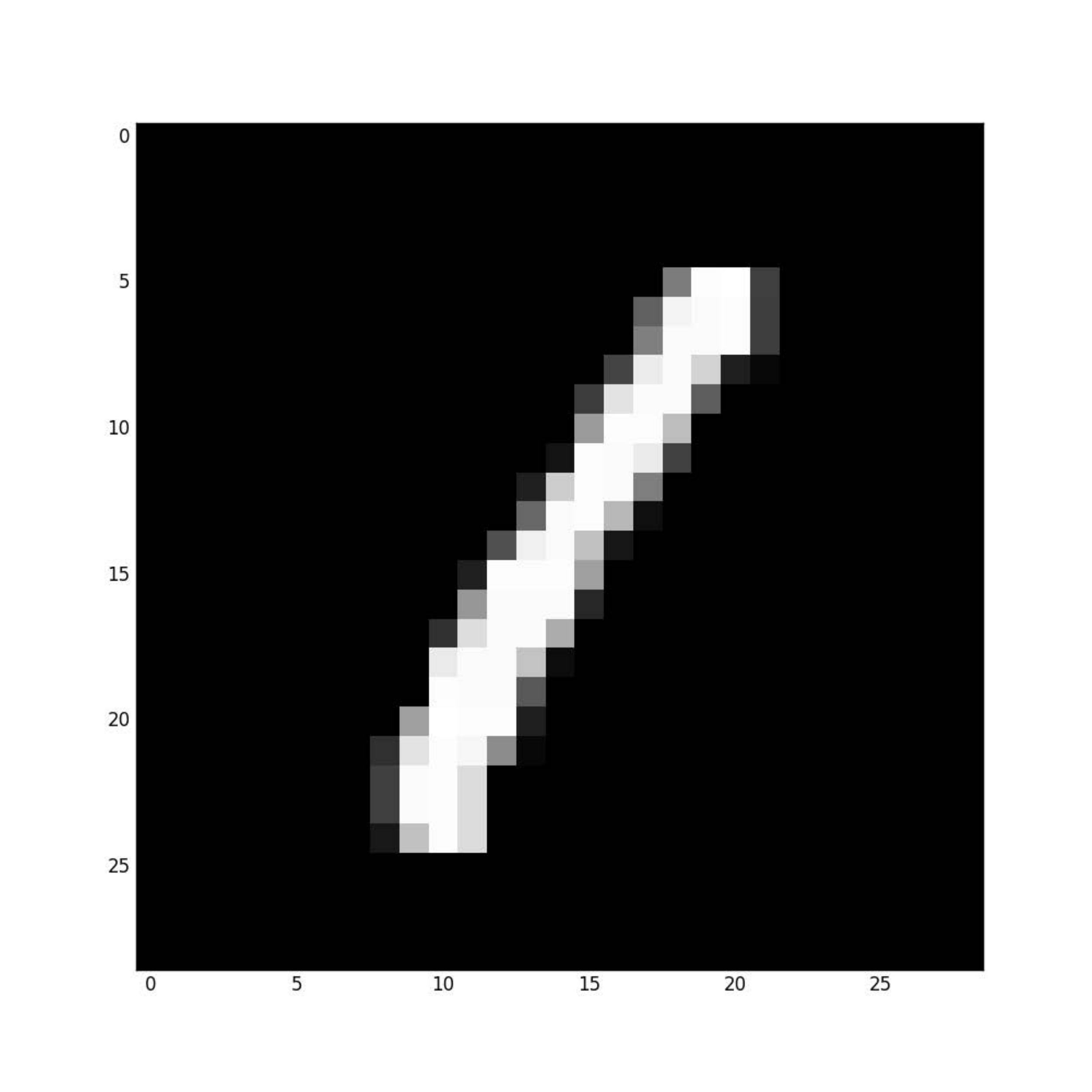}}
\subfigure{\includegraphics[width=0.083\textwidth]{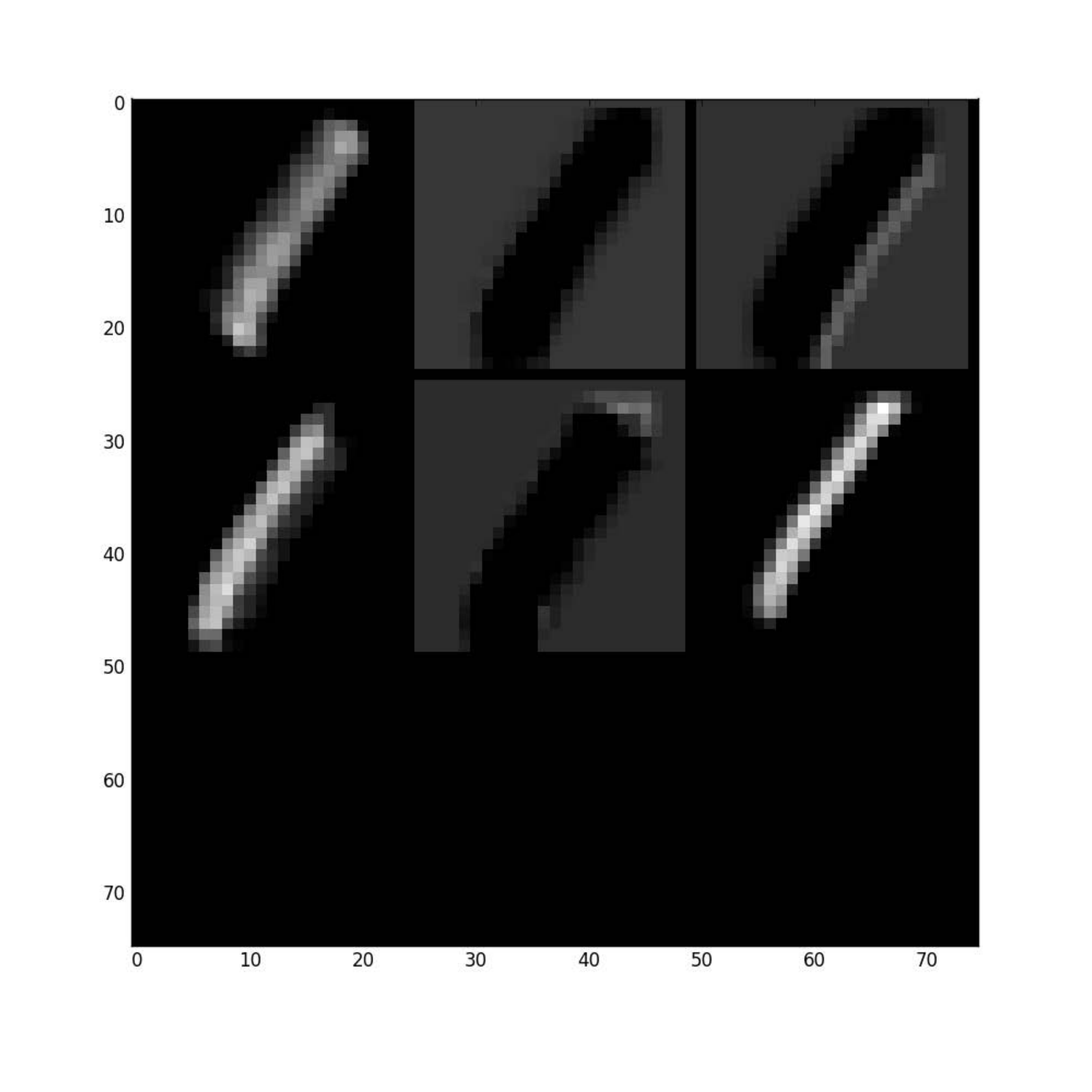}}
\subfigure{\includegraphics[width=0.083\textwidth]{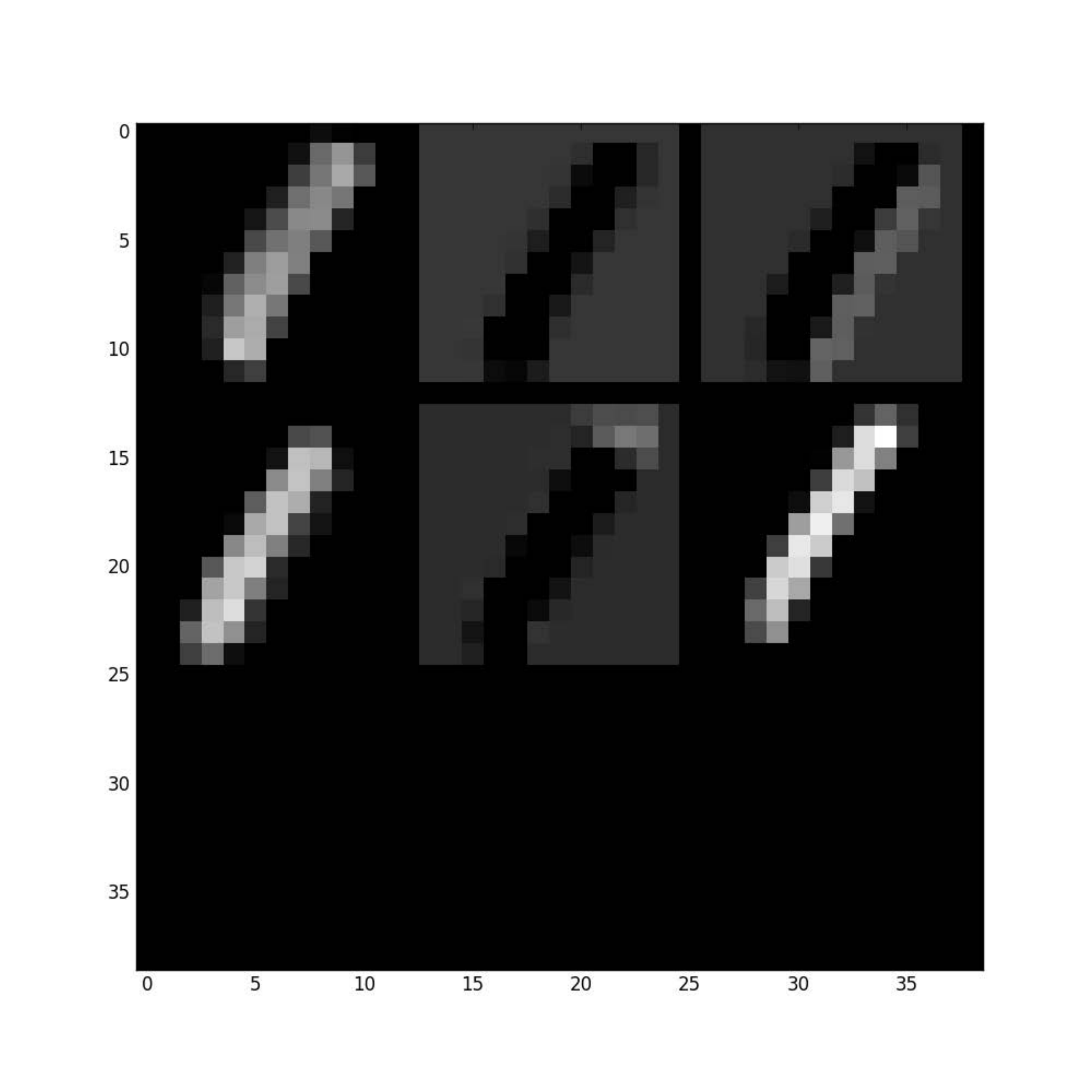}}
\subfigure{\includegraphics[width=0.083\textwidth]{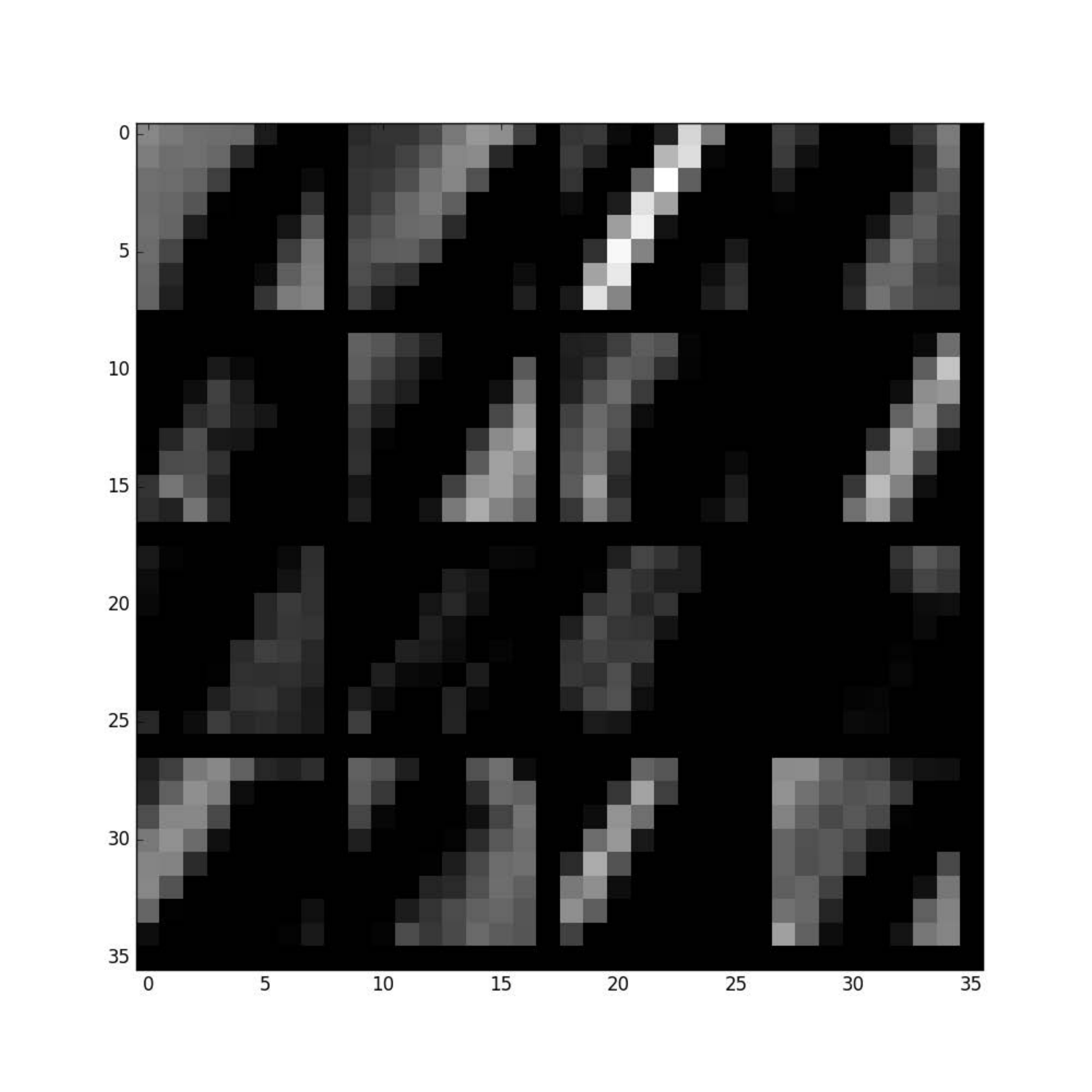}}
\subfigure{\includegraphics[width=0.083\textwidth]{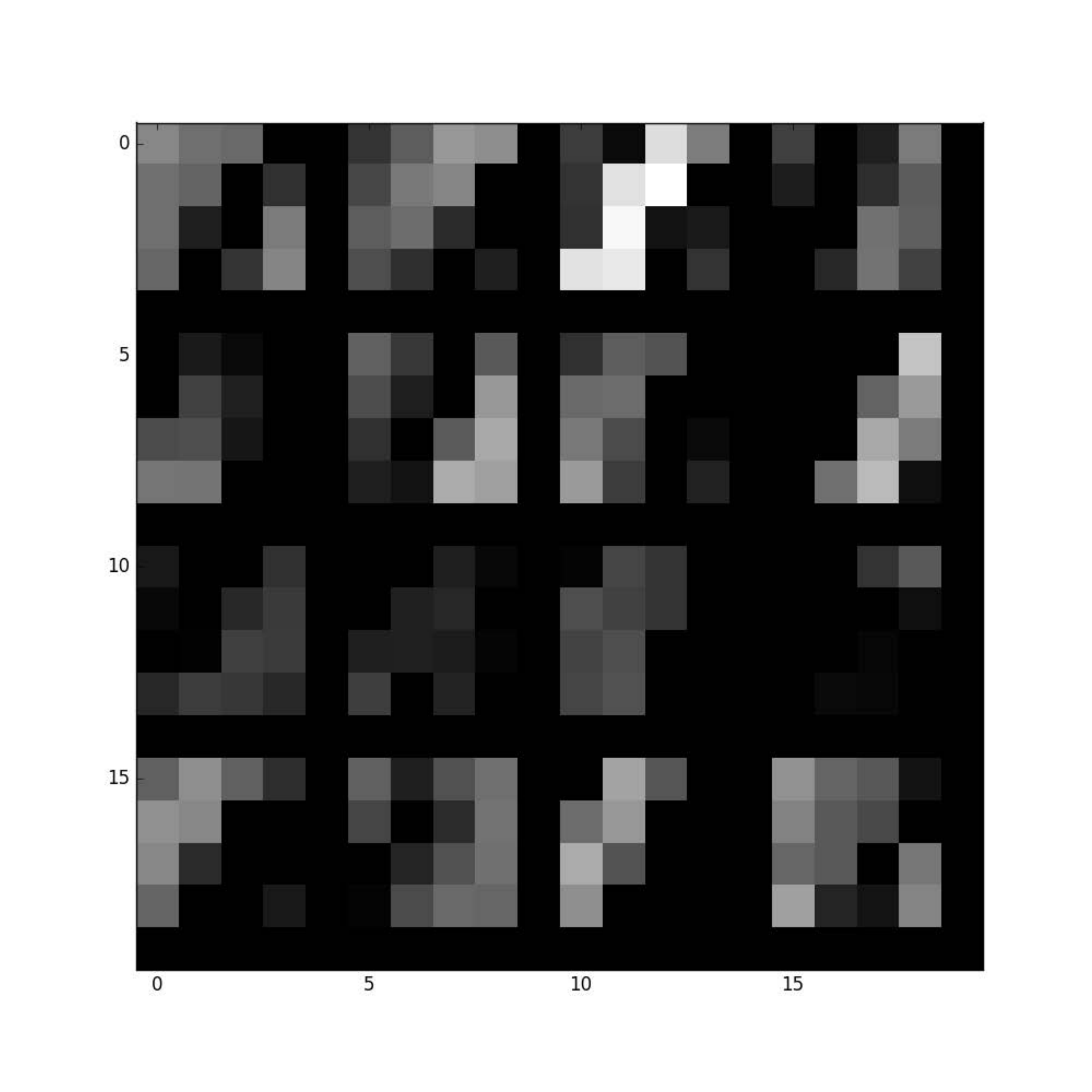}}
\subfigure{\includegraphics[width=0.083\textwidth]{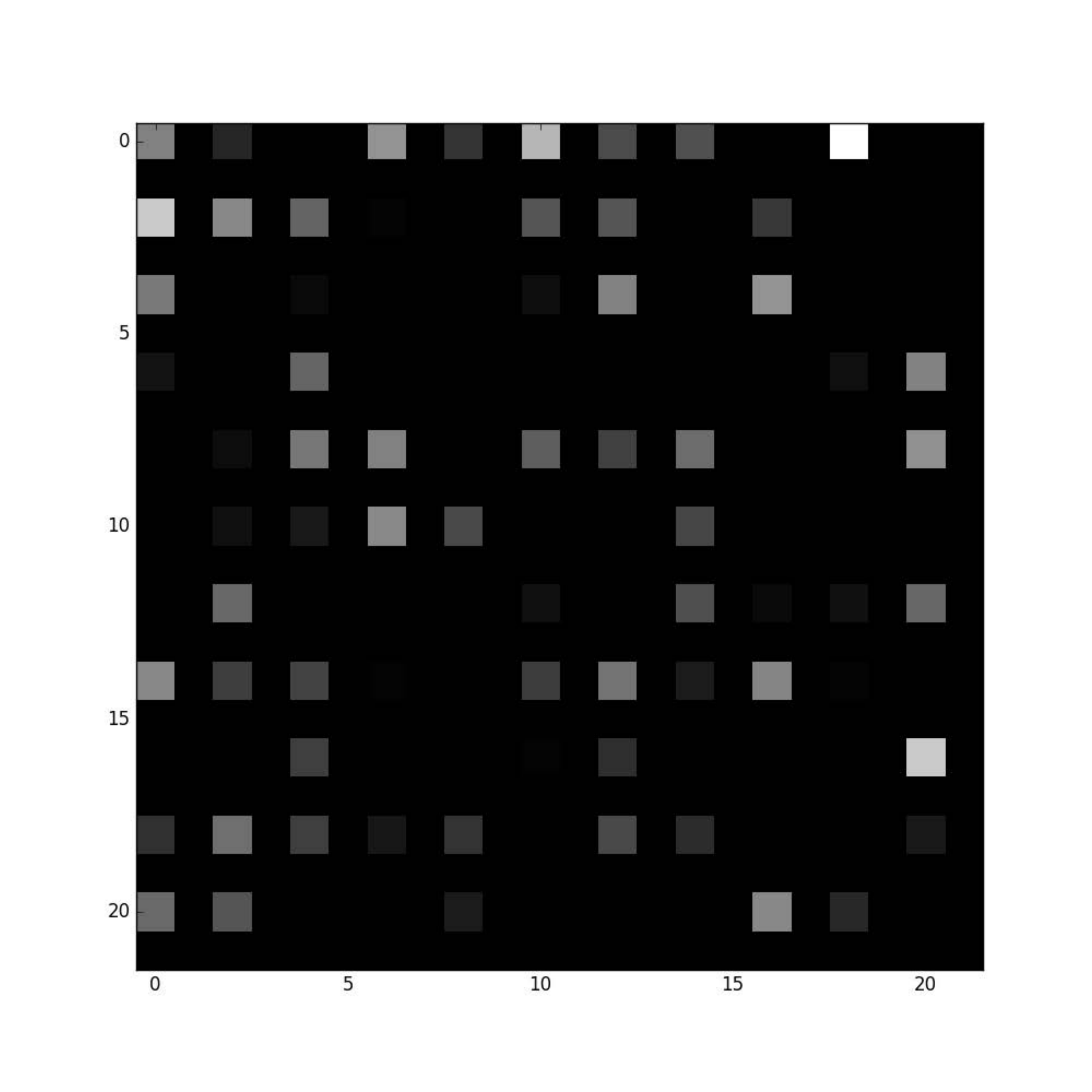}}
\subfigure{\includegraphics[width=0.083\textwidth]{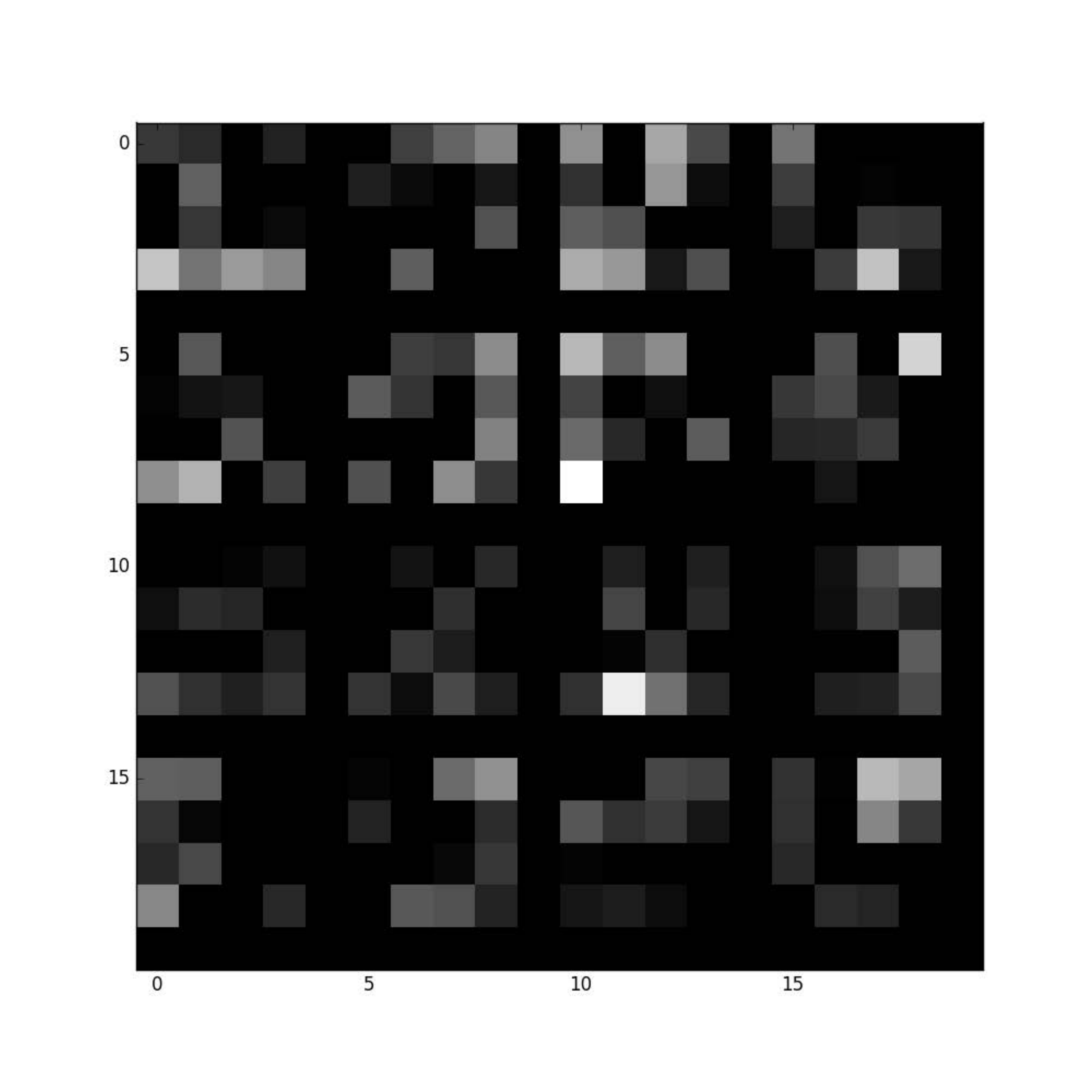}}
\subfigure{\includegraphics[width=0.083\textwidth]{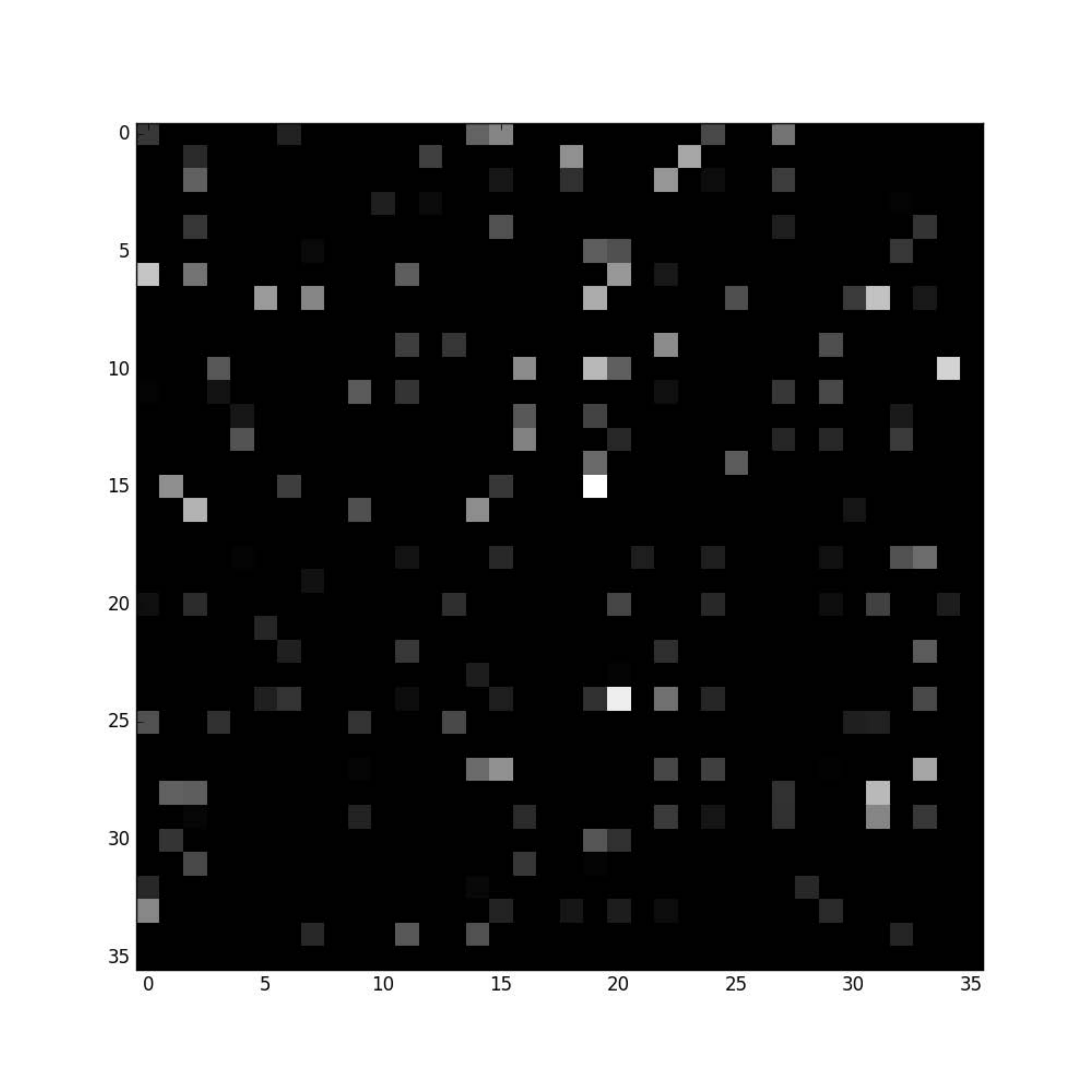}}
\subfigure{\includegraphics[width=0.083\textwidth]{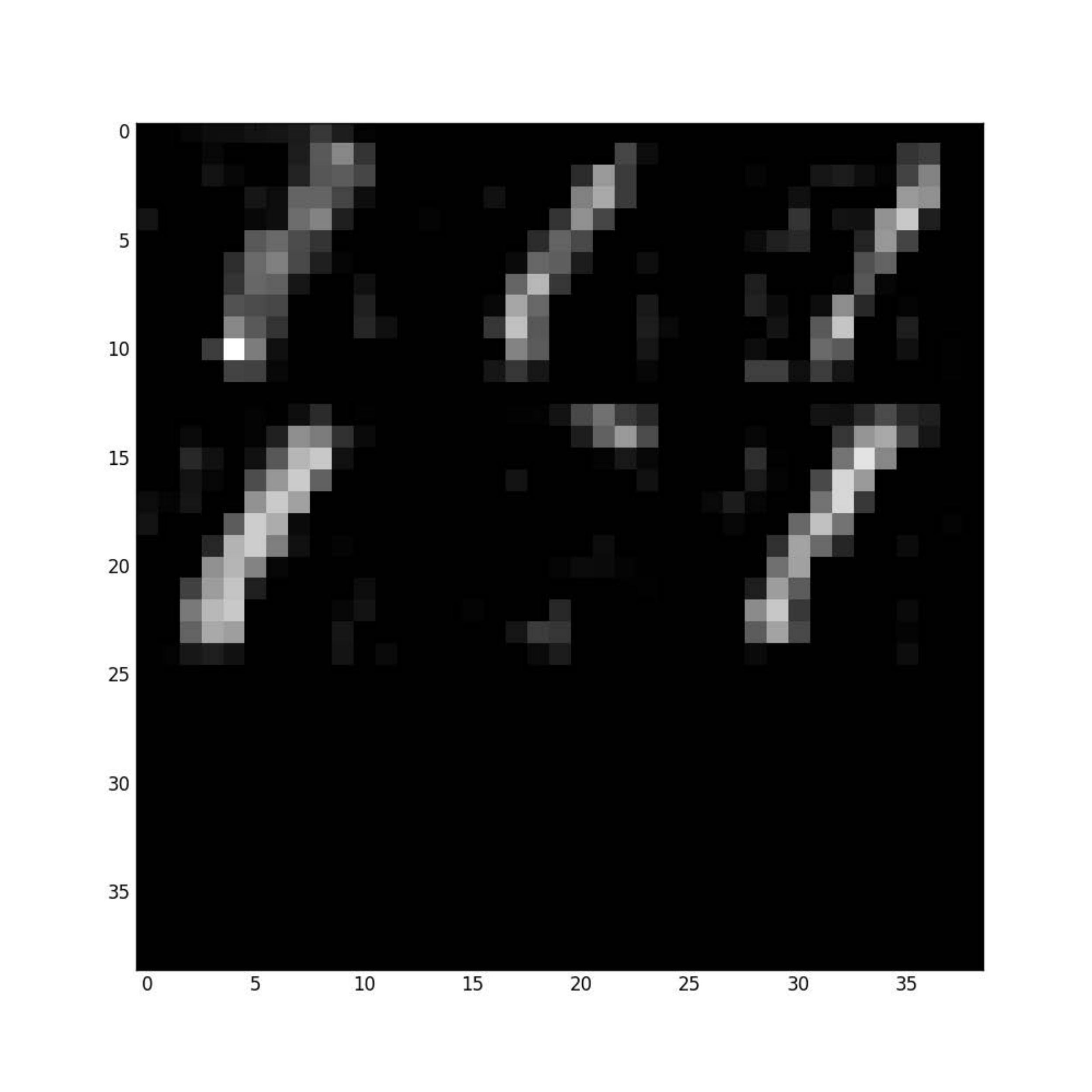}}
\subfigure{\includegraphics[width=0.083\textwidth]{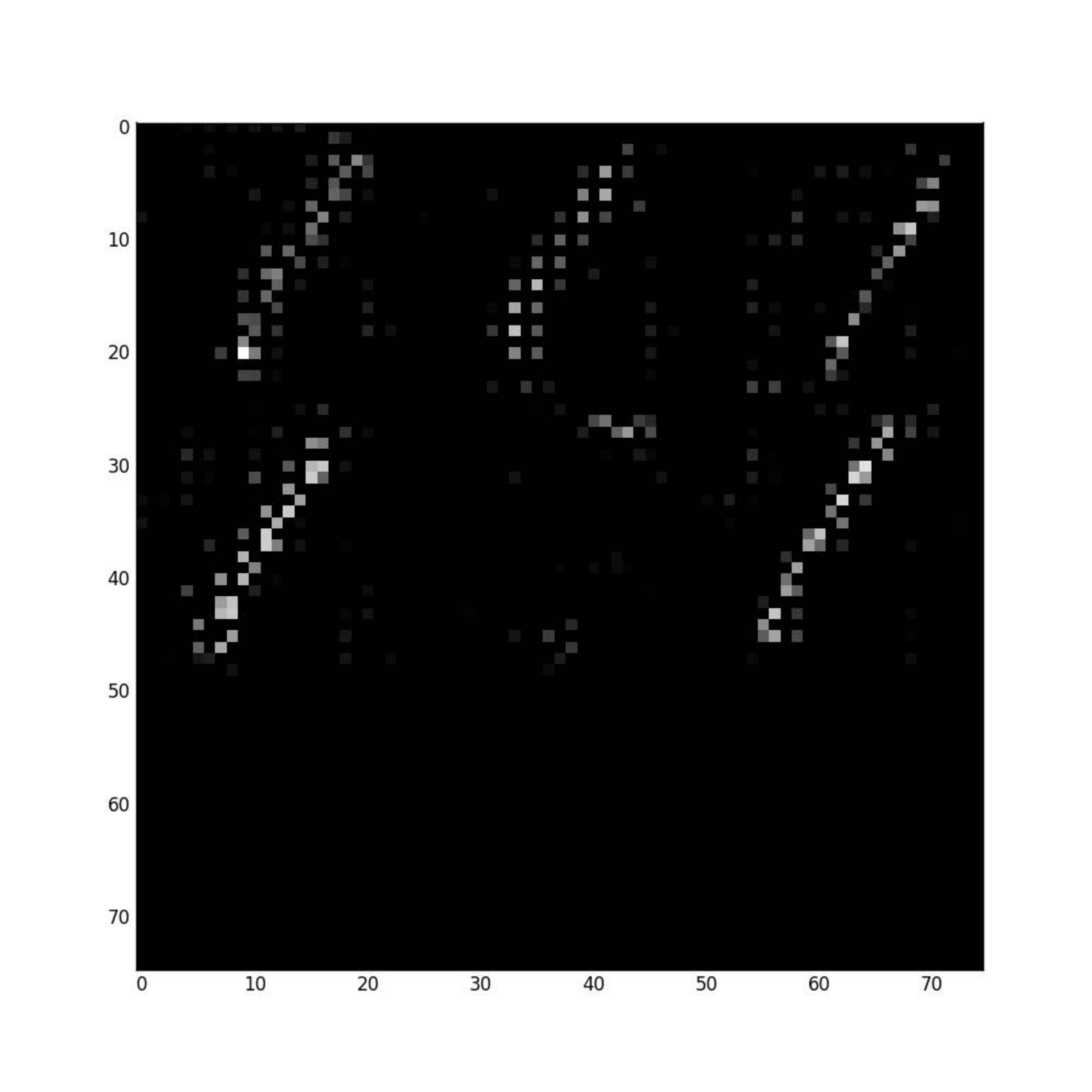}}
\subfigure{\includegraphics[width=0.083\textwidth]{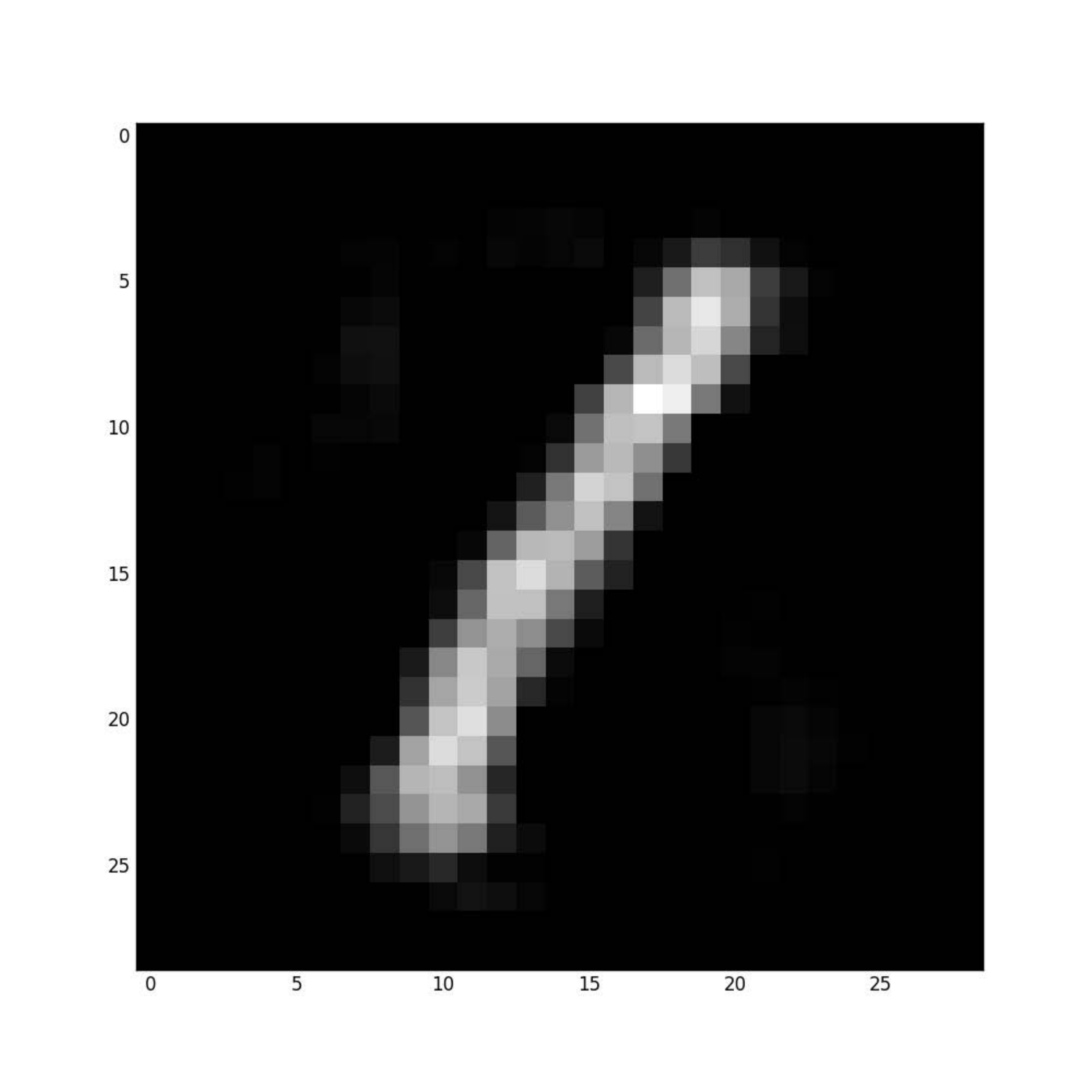}}\\
\subfigure{\includegraphics[width=0.083\textwidth]{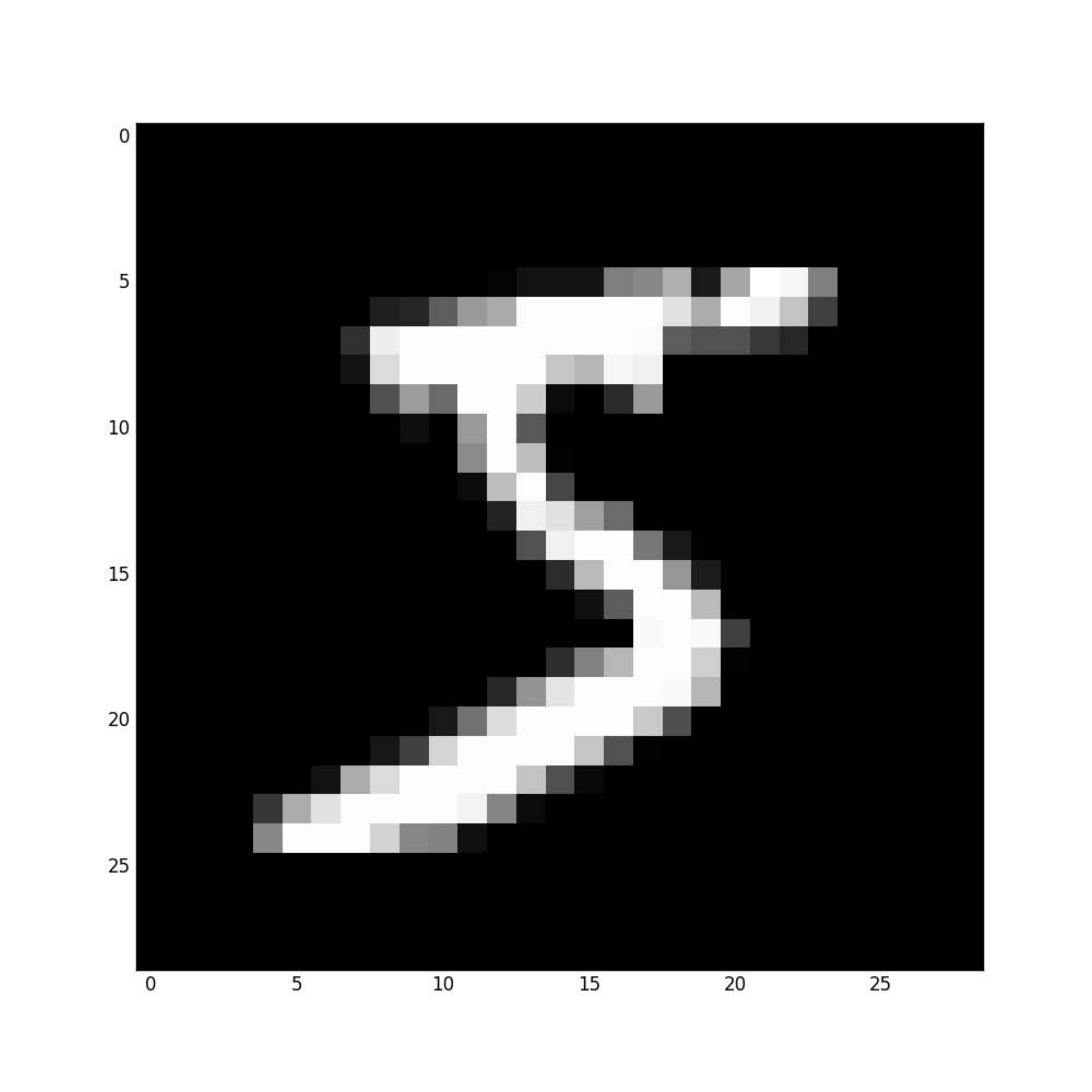}}
\subfigure{\includegraphics[width=0.083\textwidth]{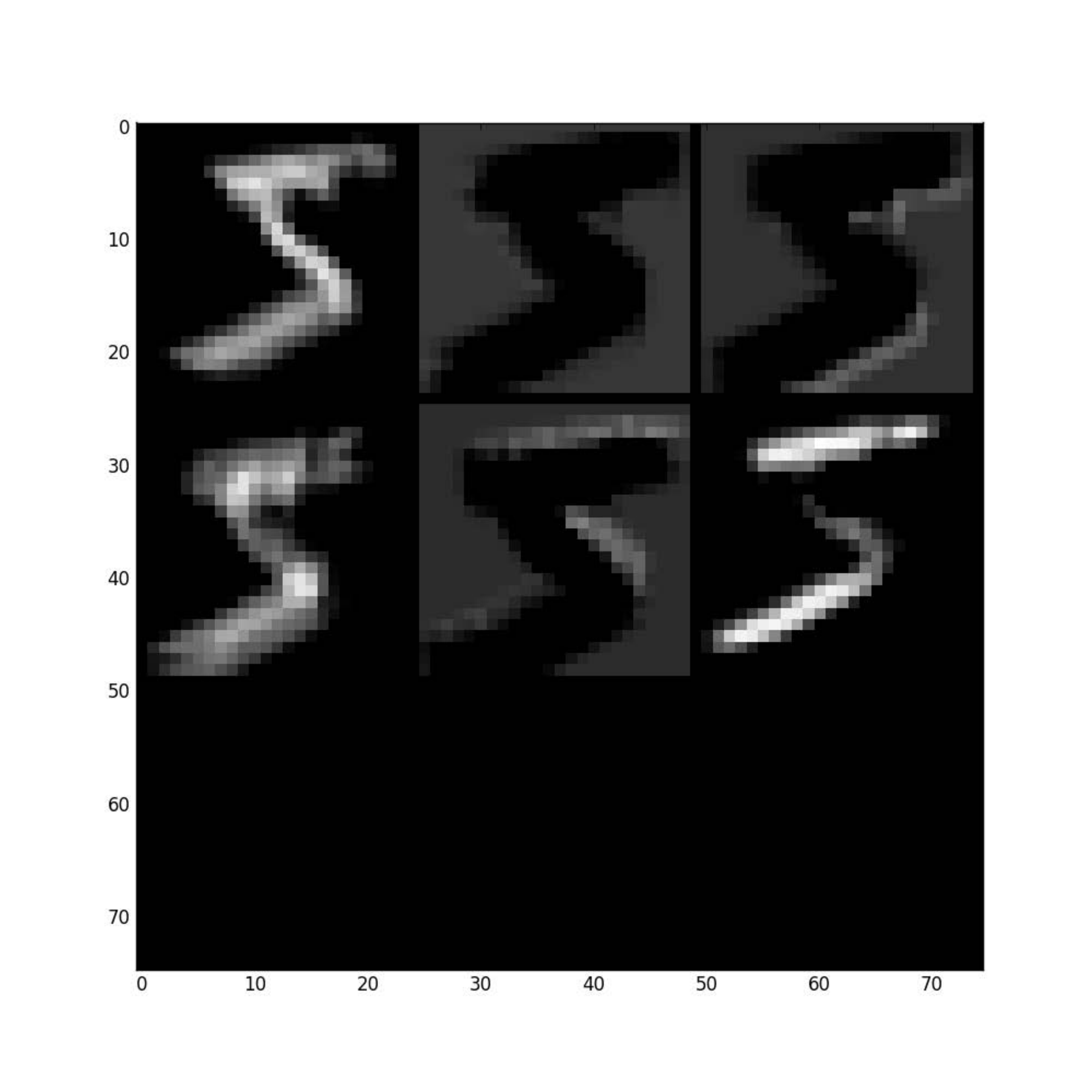}}
\subfigure{\includegraphics[width=0.083\textwidth]{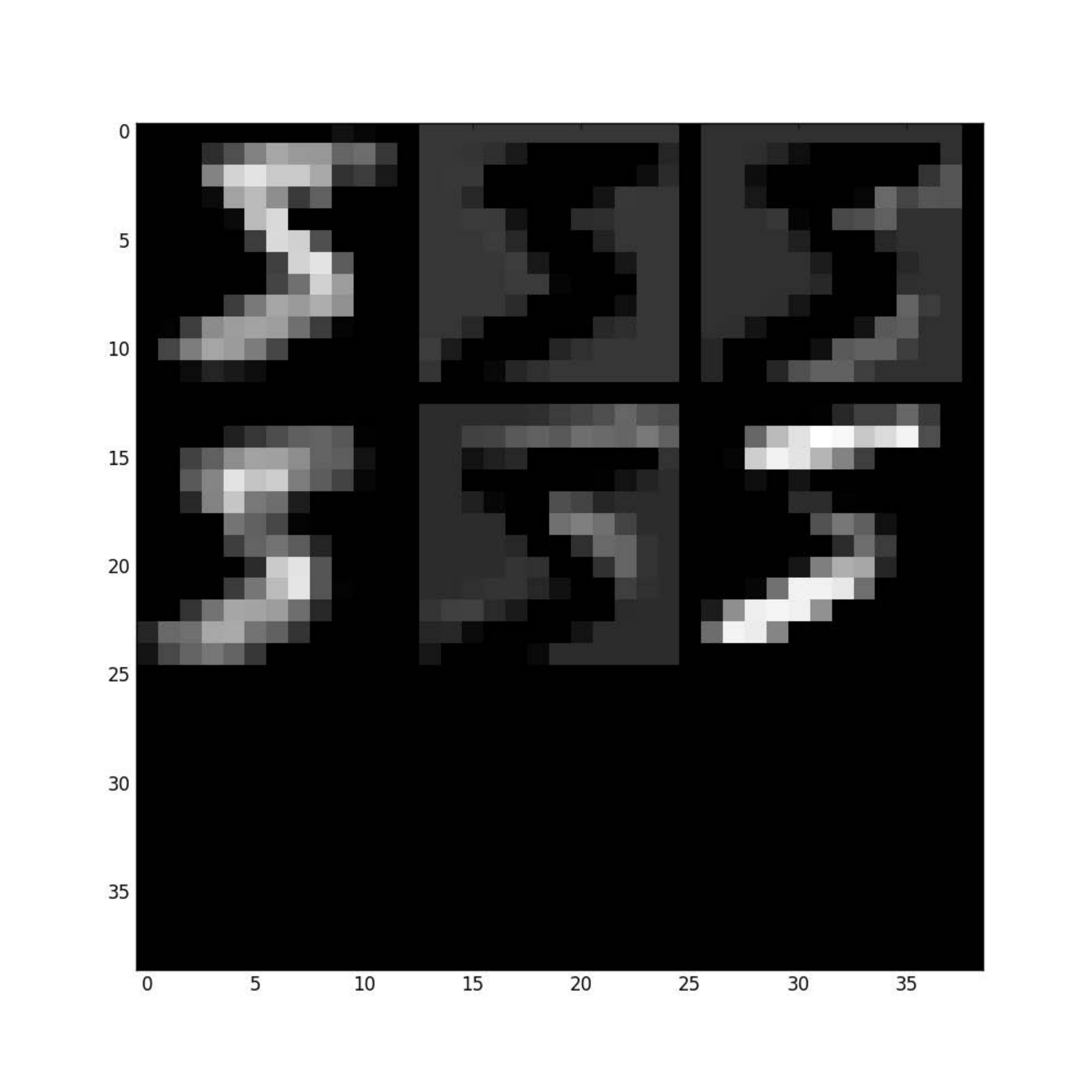}}
\subfigure{\includegraphics[width=0.083\textwidth]{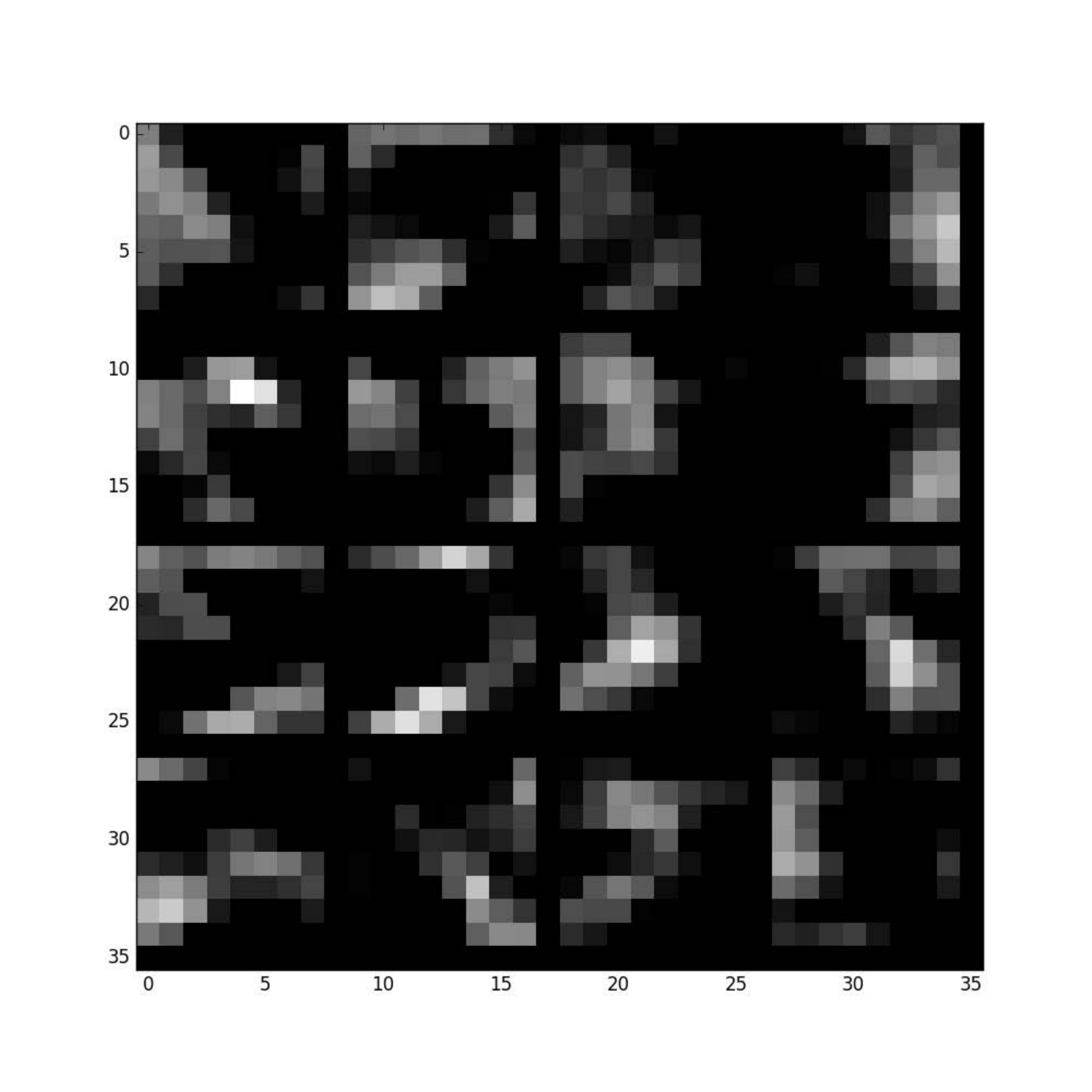}}
\subfigure{\includegraphics[width=0.083\textwidth]{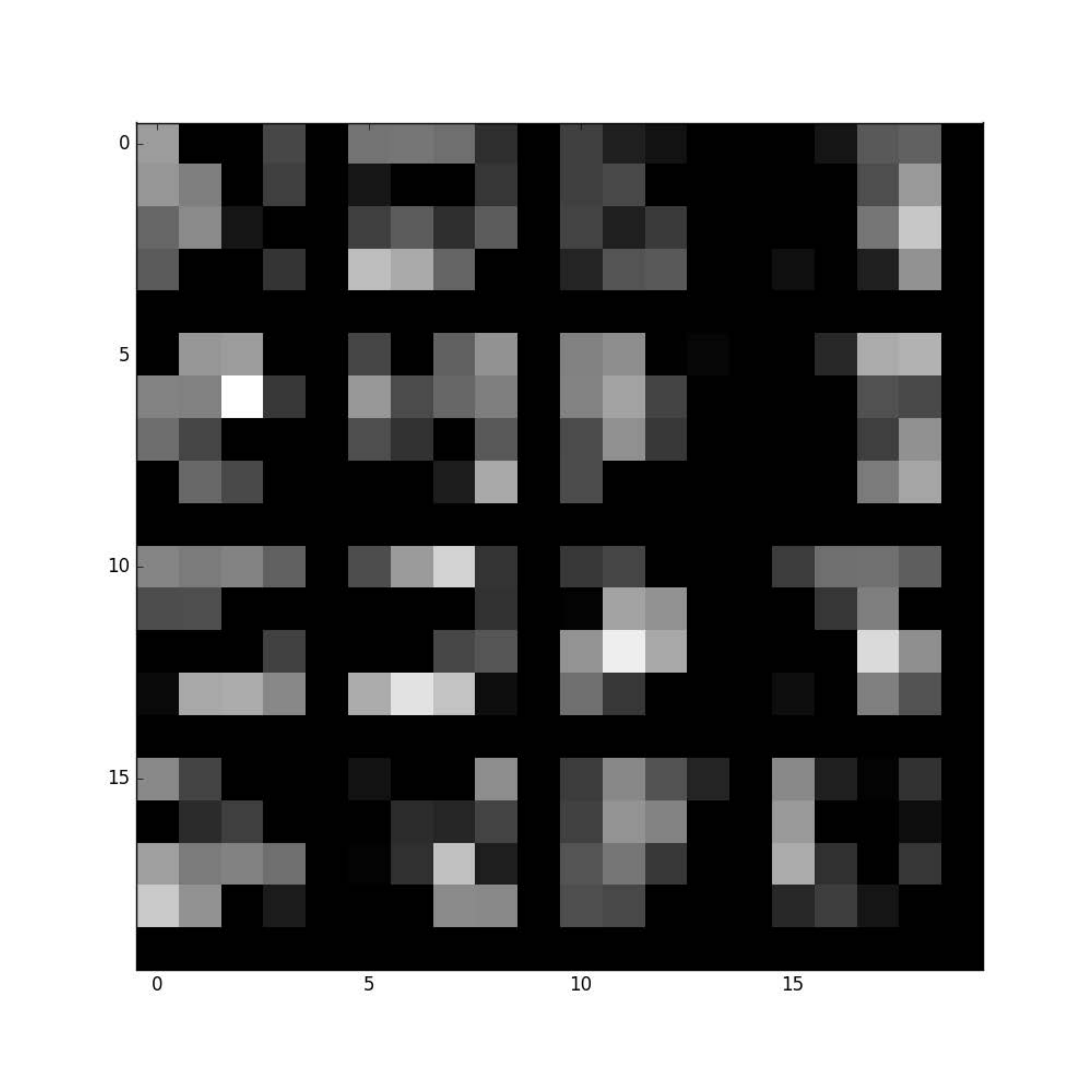}}
\subfigure{\includegraphics[width=0.083\textwidth]{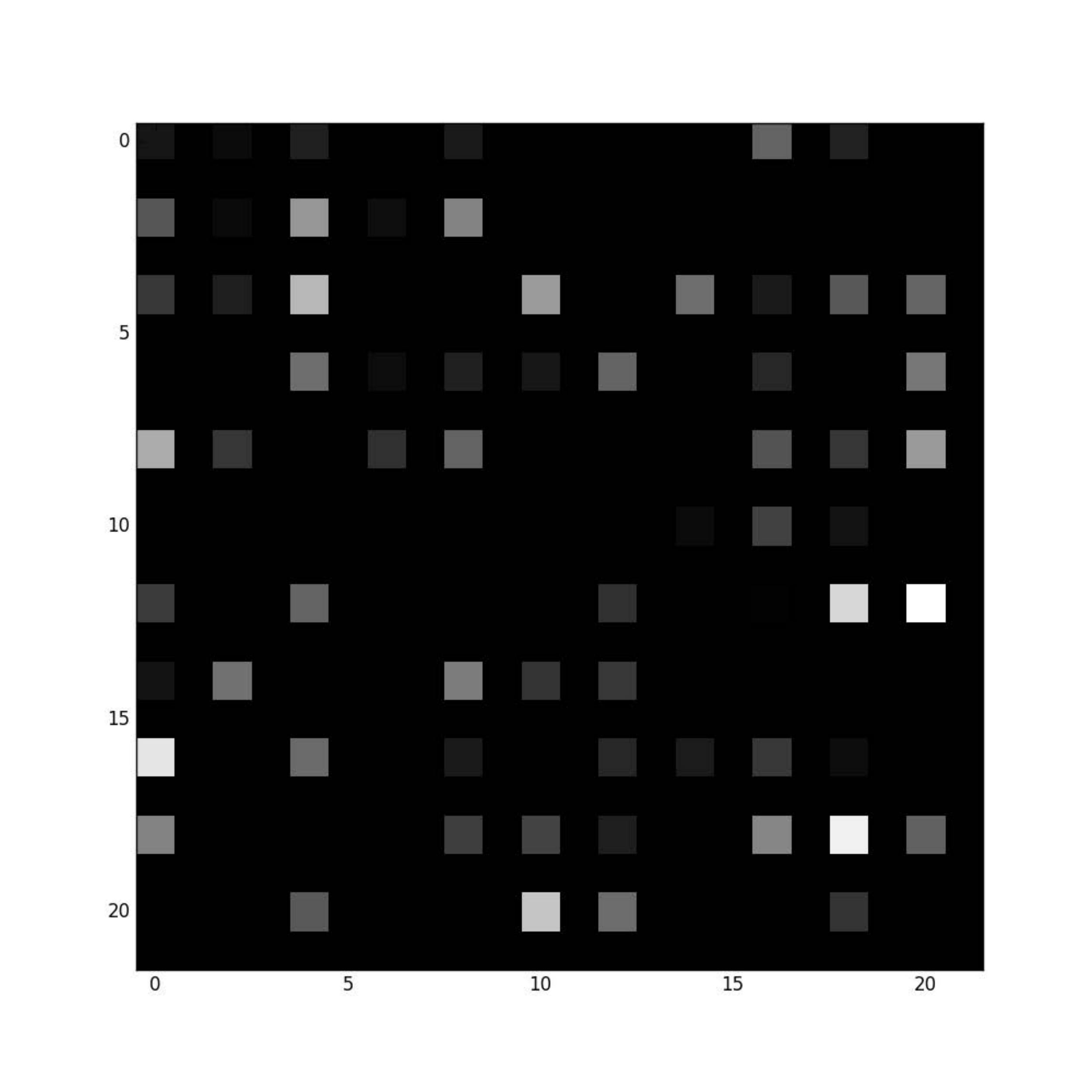}}
\subfigure{\includegraphics[width=0.083\textwidth]{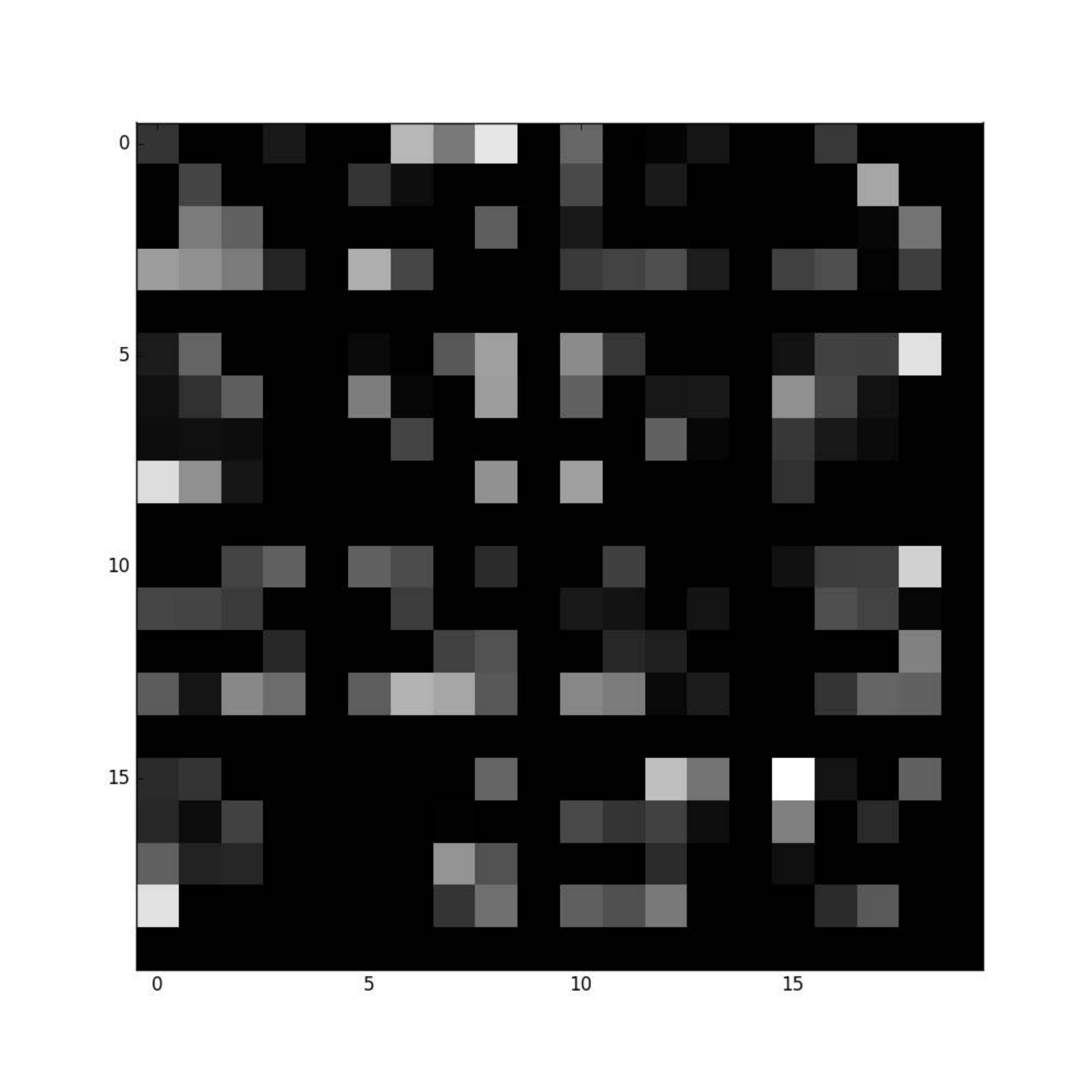}}
\subfigure{\includegraphics[width=0.083\textwidth]{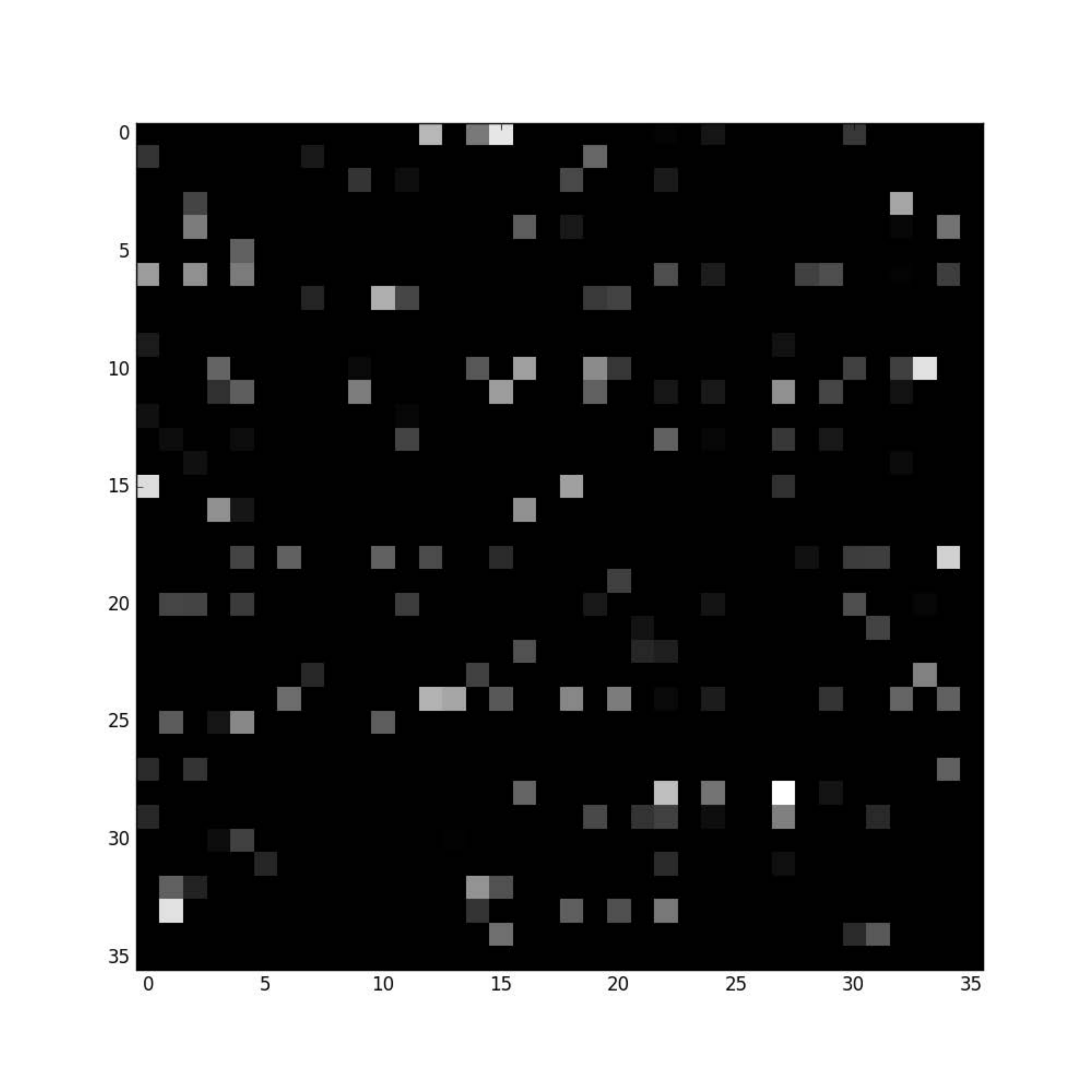}}
\subfigure{\includegraphics[width=0.083\textwidth]{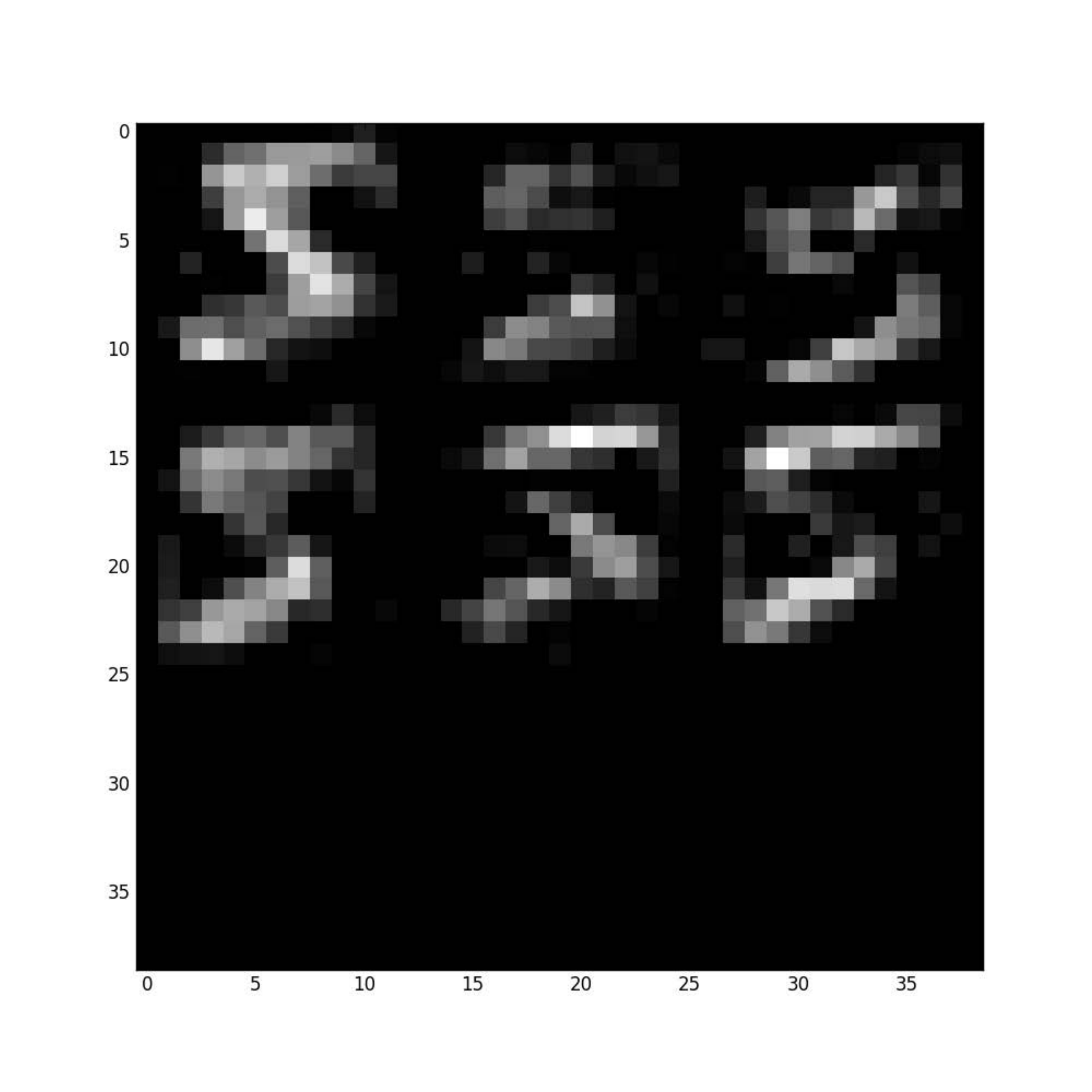}}
\subfigure{\includegraphics[width=0.083\textwidth]{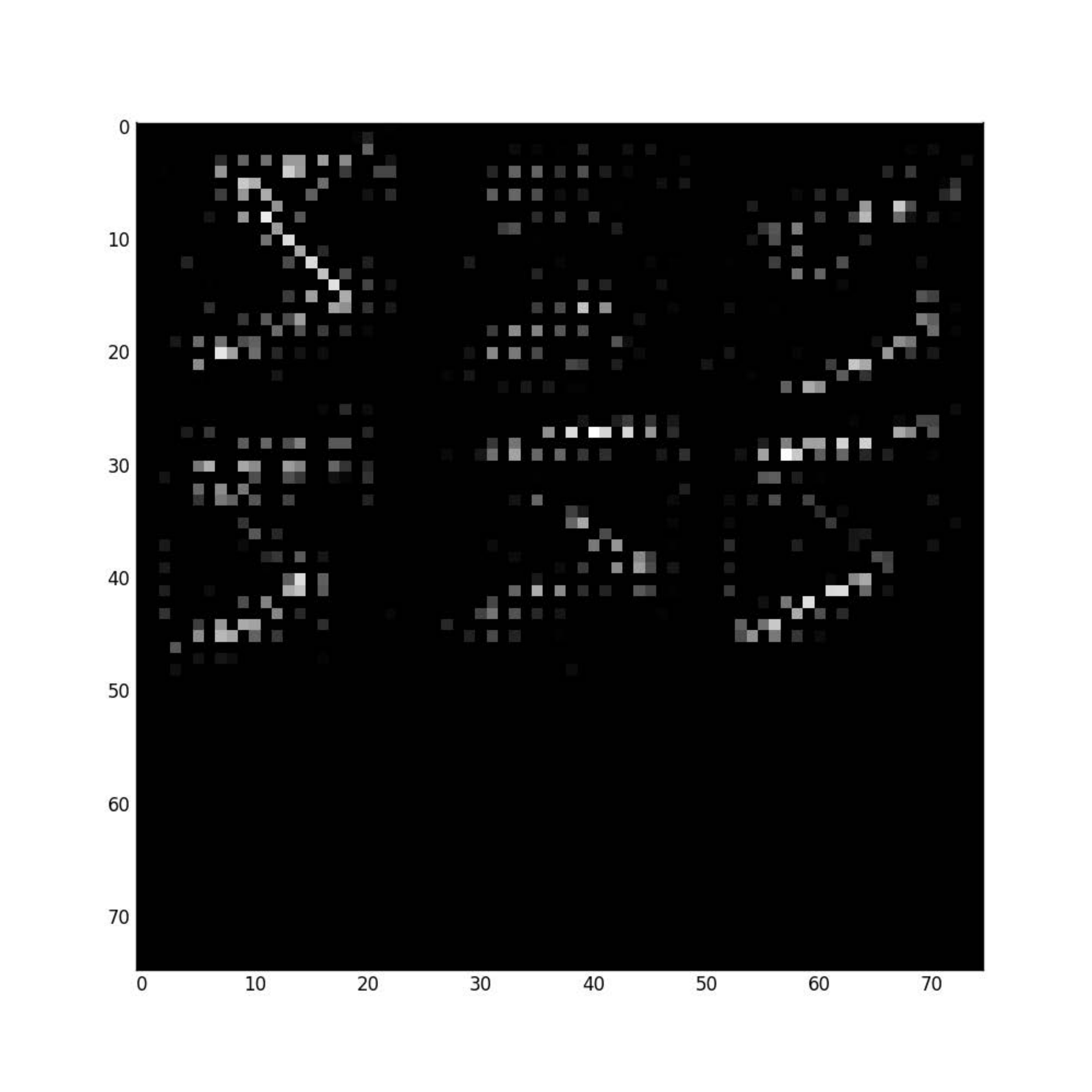}}
\subfigure{\includegraphics[width=0.083\textwidth]{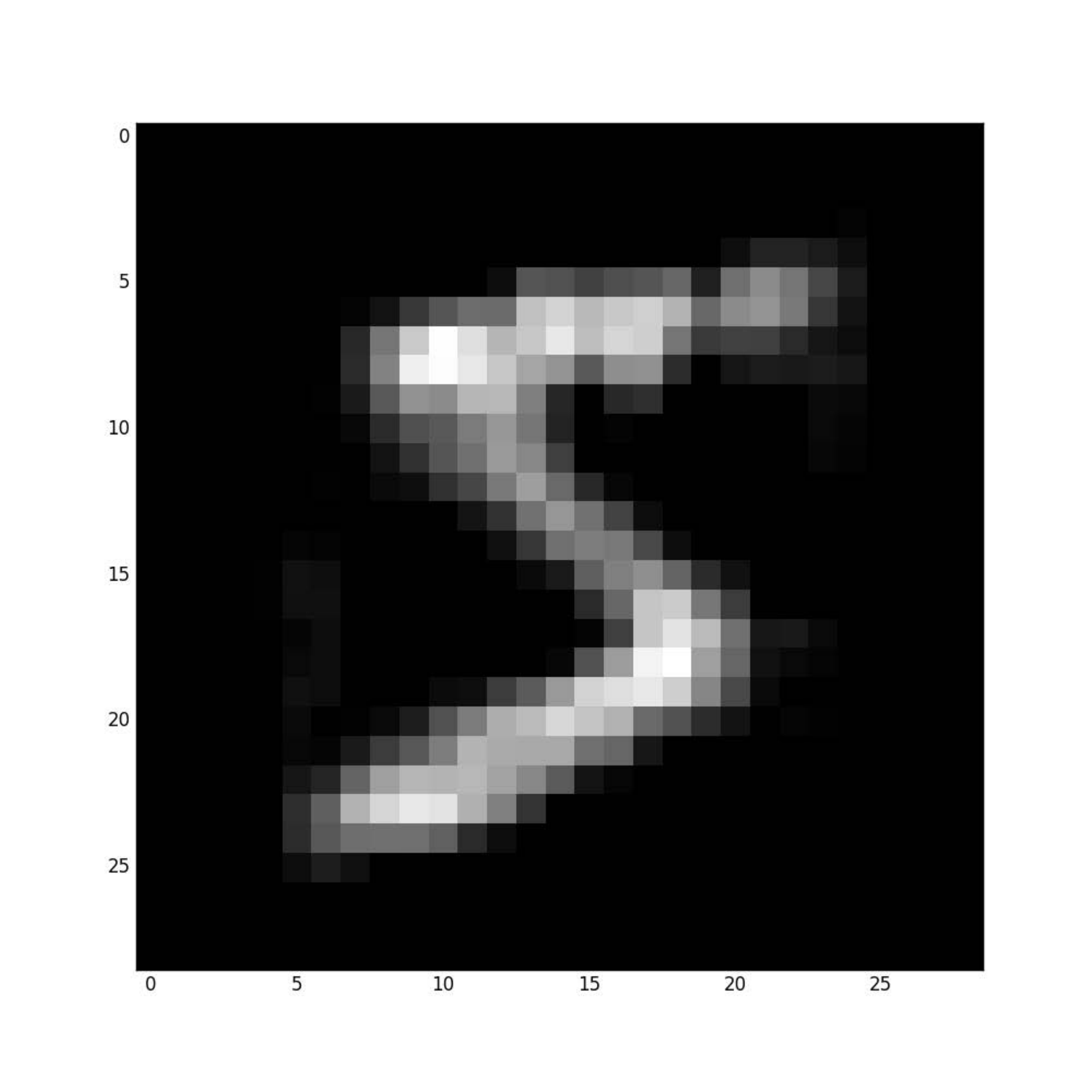}}\\
\subfigure[in.~]{\includegraphics[width=0.083\textwidth]{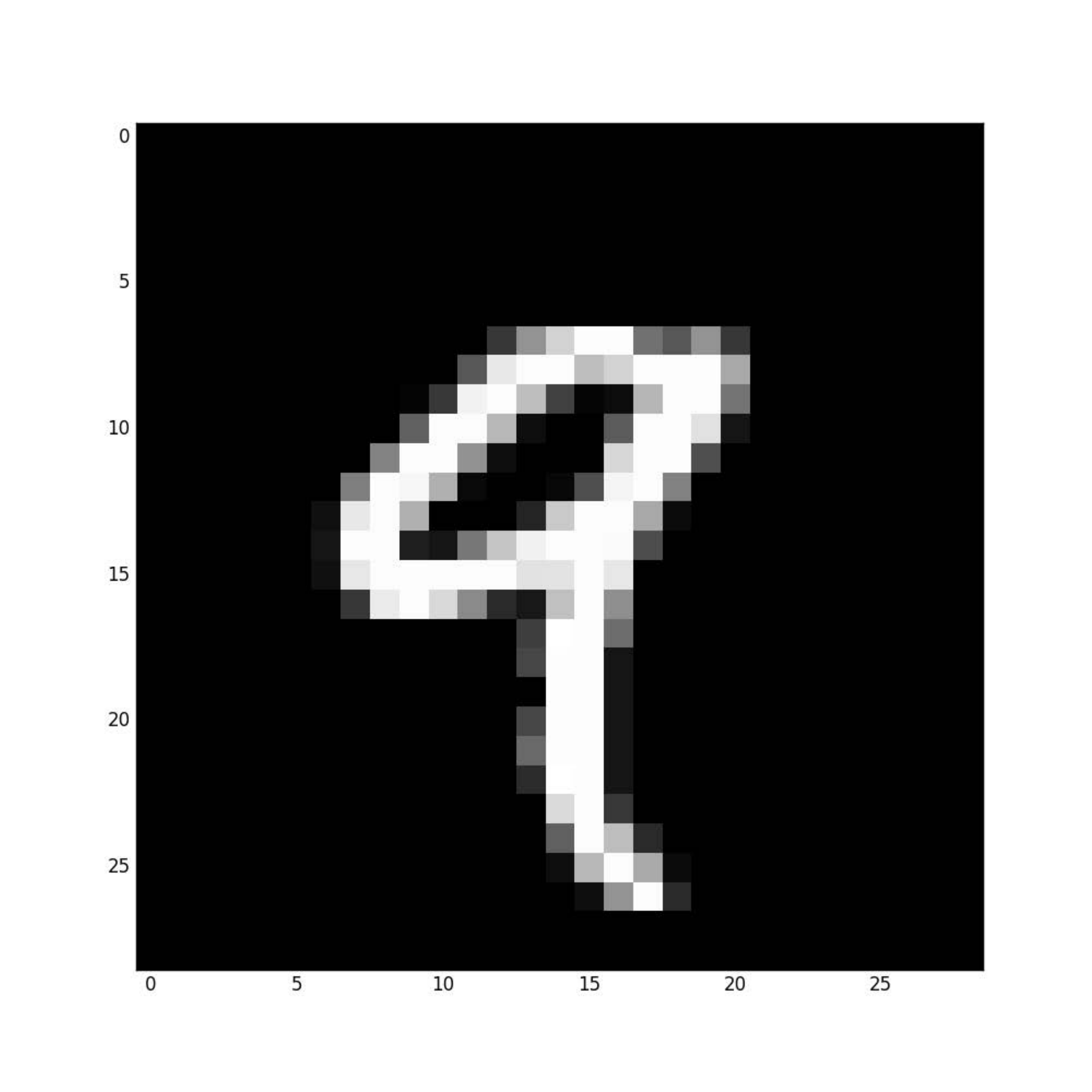}}
\subfigure[c1~]{\includegraphics[width=0.083\textwidth]{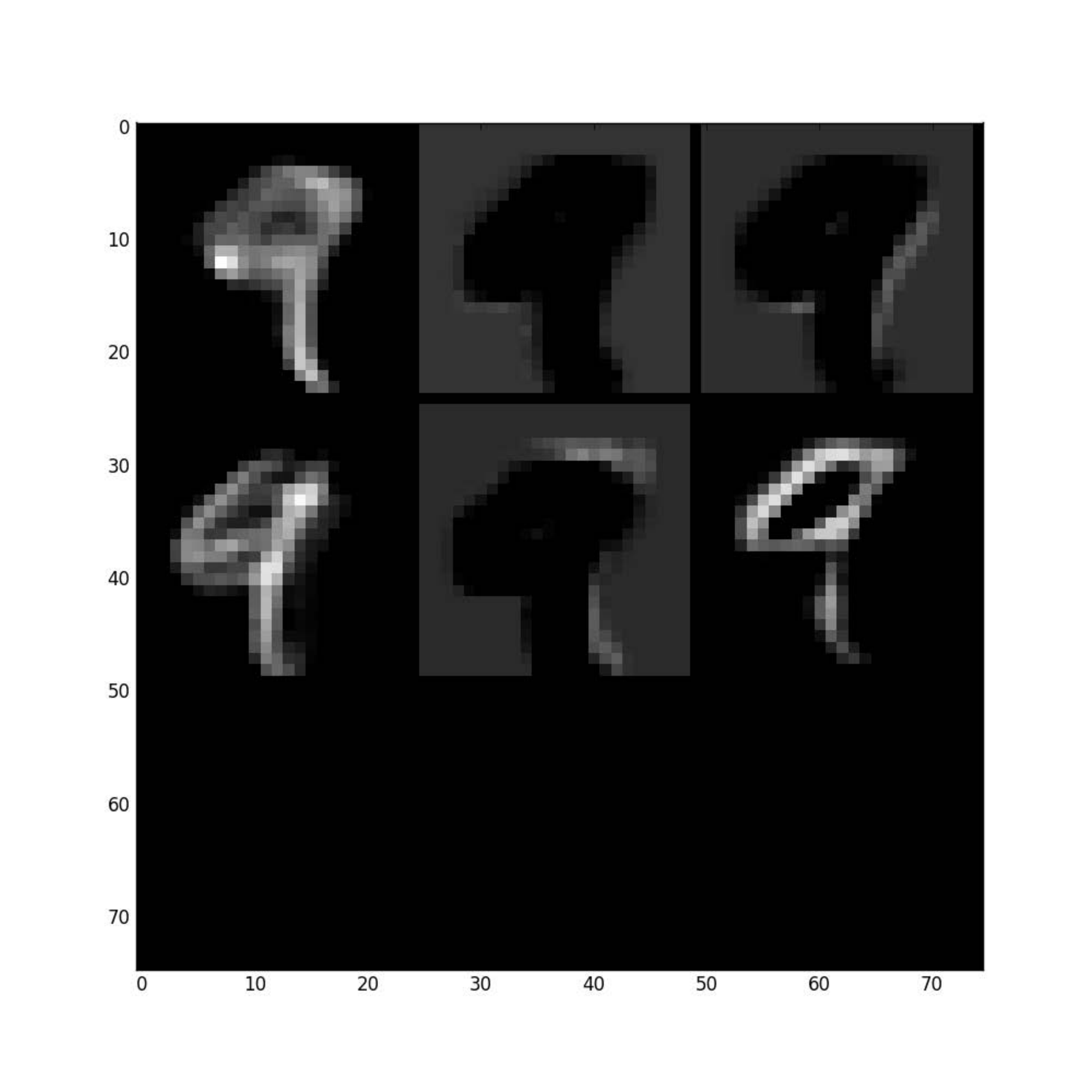}}
\subfigure[p1~]{\includegraphics[width=0.083\textwidth]{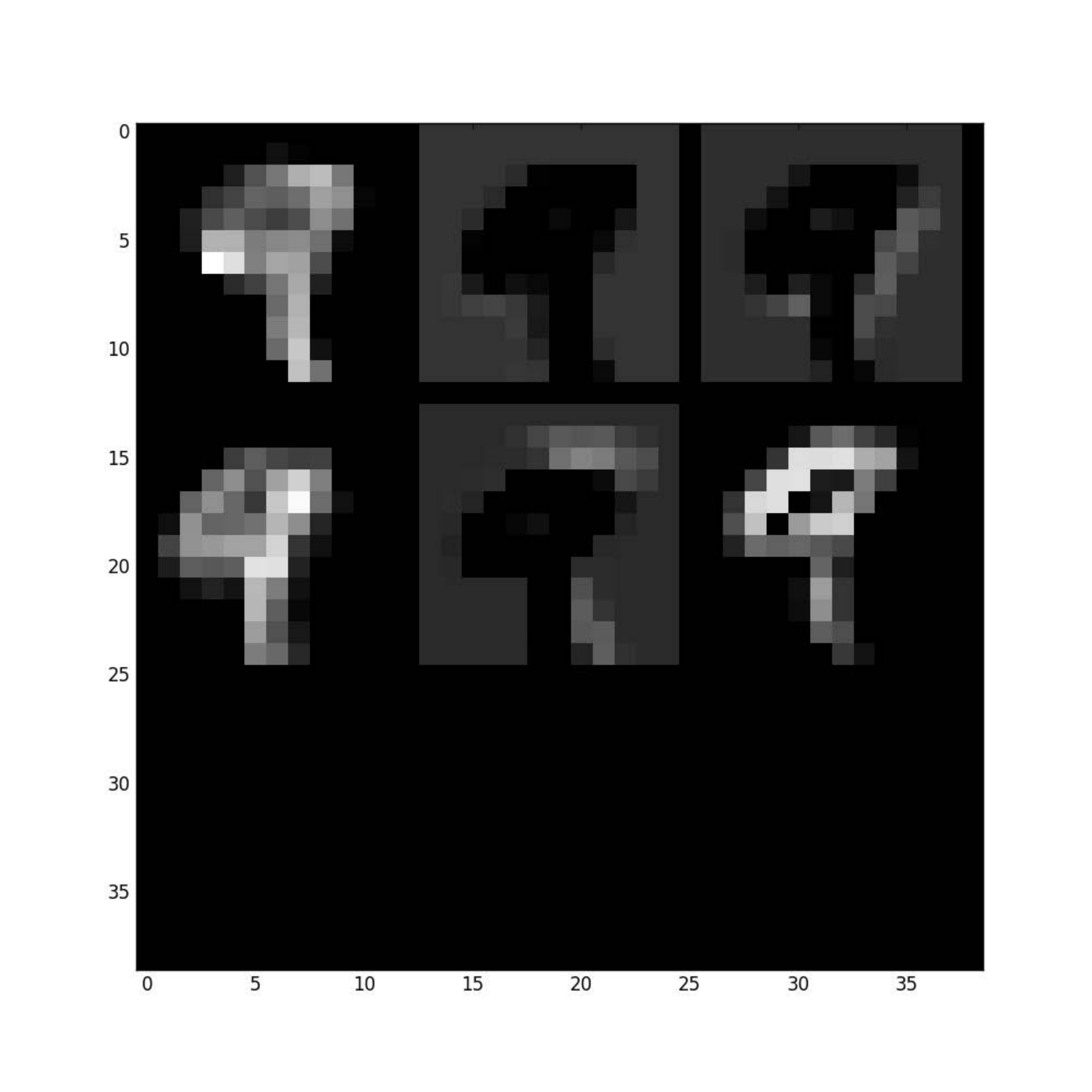}}
\subfigure[c2~]{\includegraphics[width=0.083\textwidth]{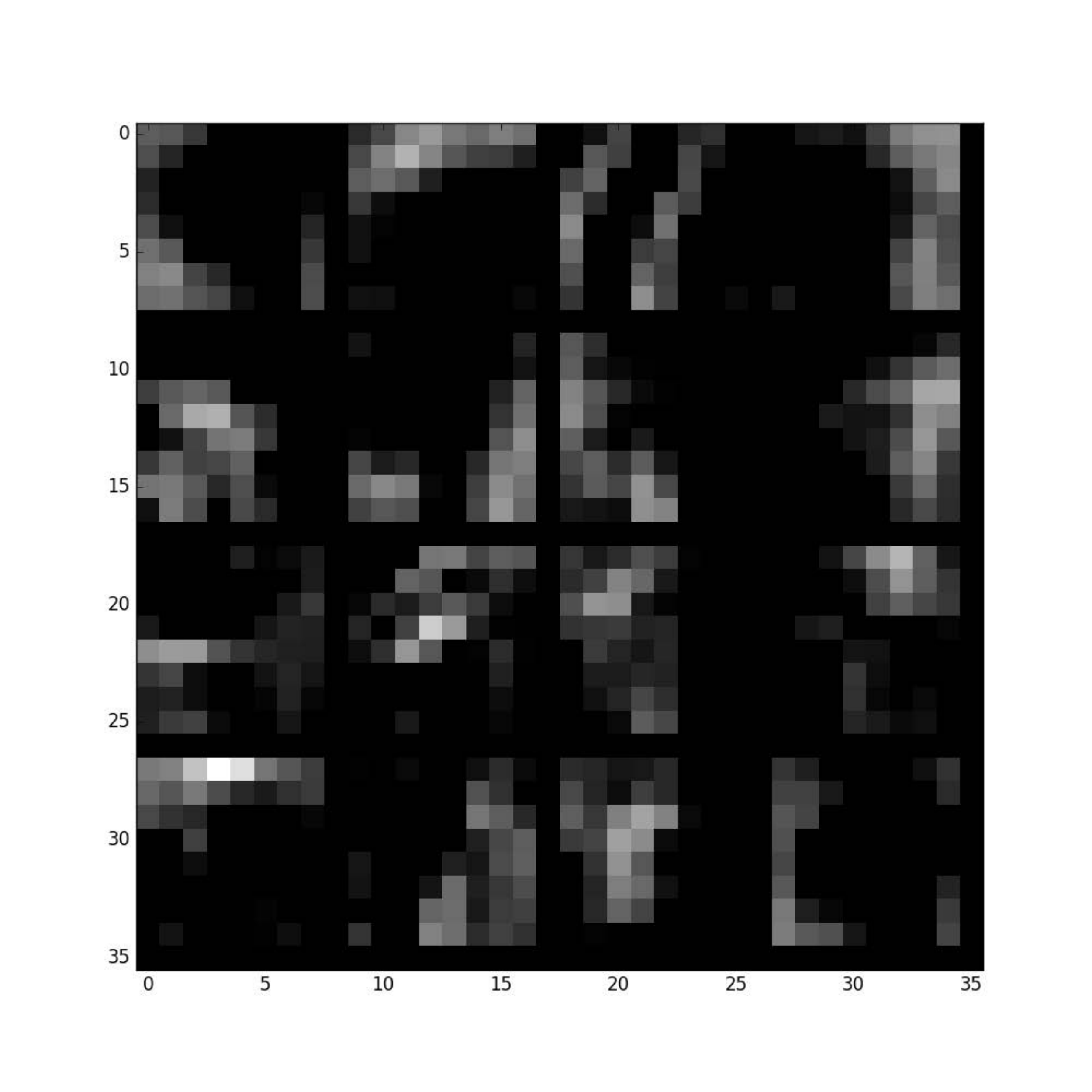}}
\subfigure[p2~]{\includegraphics[width=0.083\textwidth]{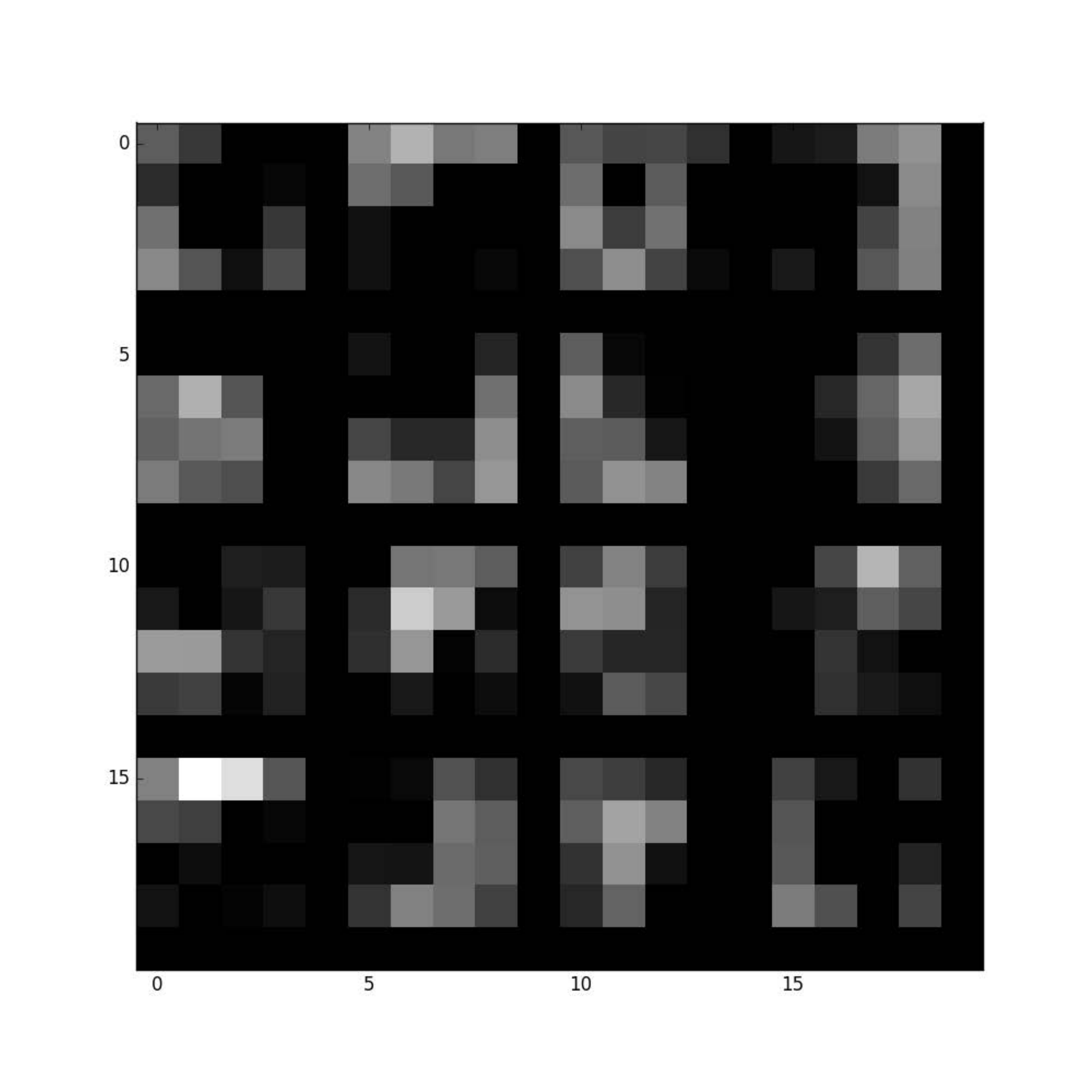}}
\subfigure[feat.~]{\includegraphics[width=0.083\textwidth]{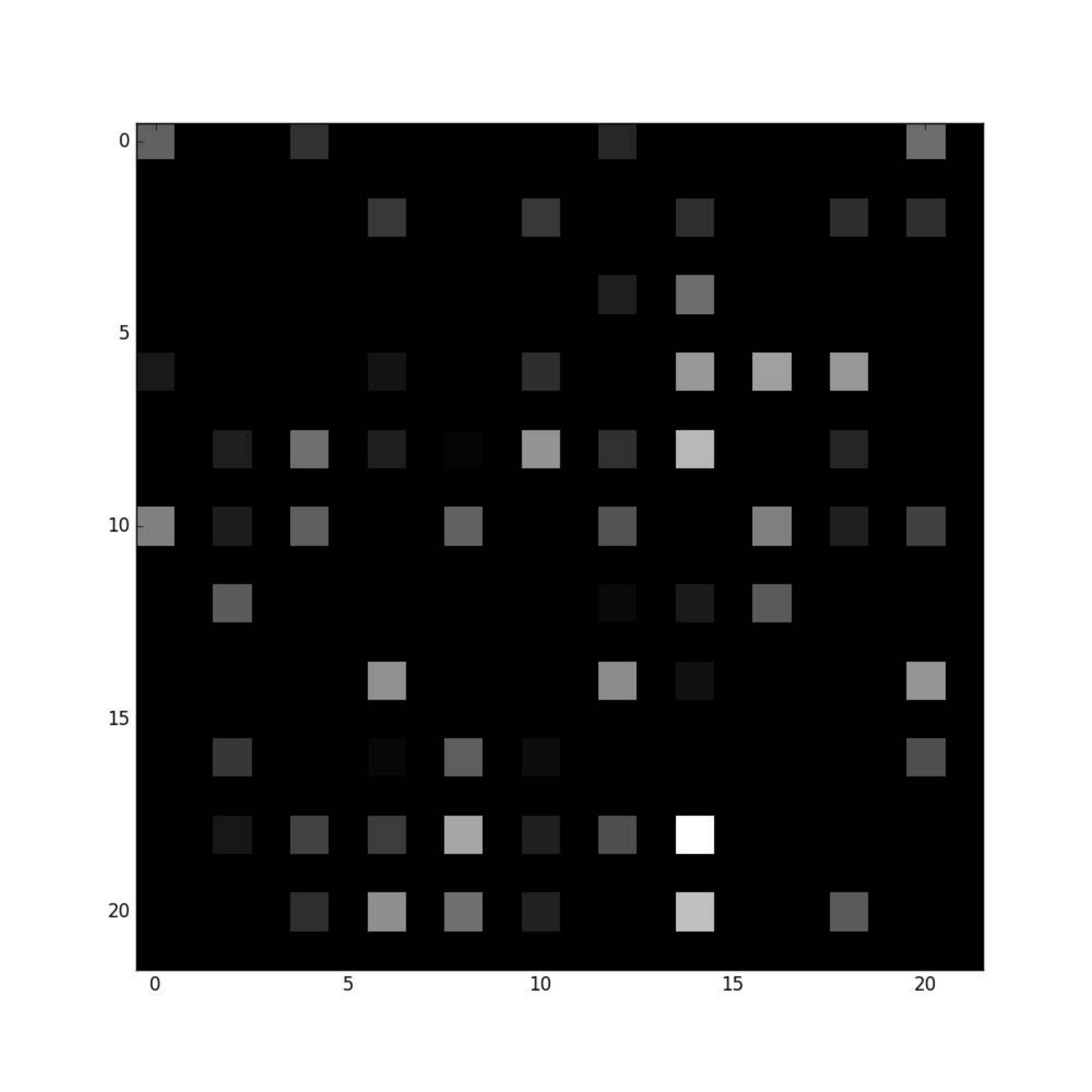}}
\subfigure[d\_p2~]{\includegraphics[width=0.083\textwidth]{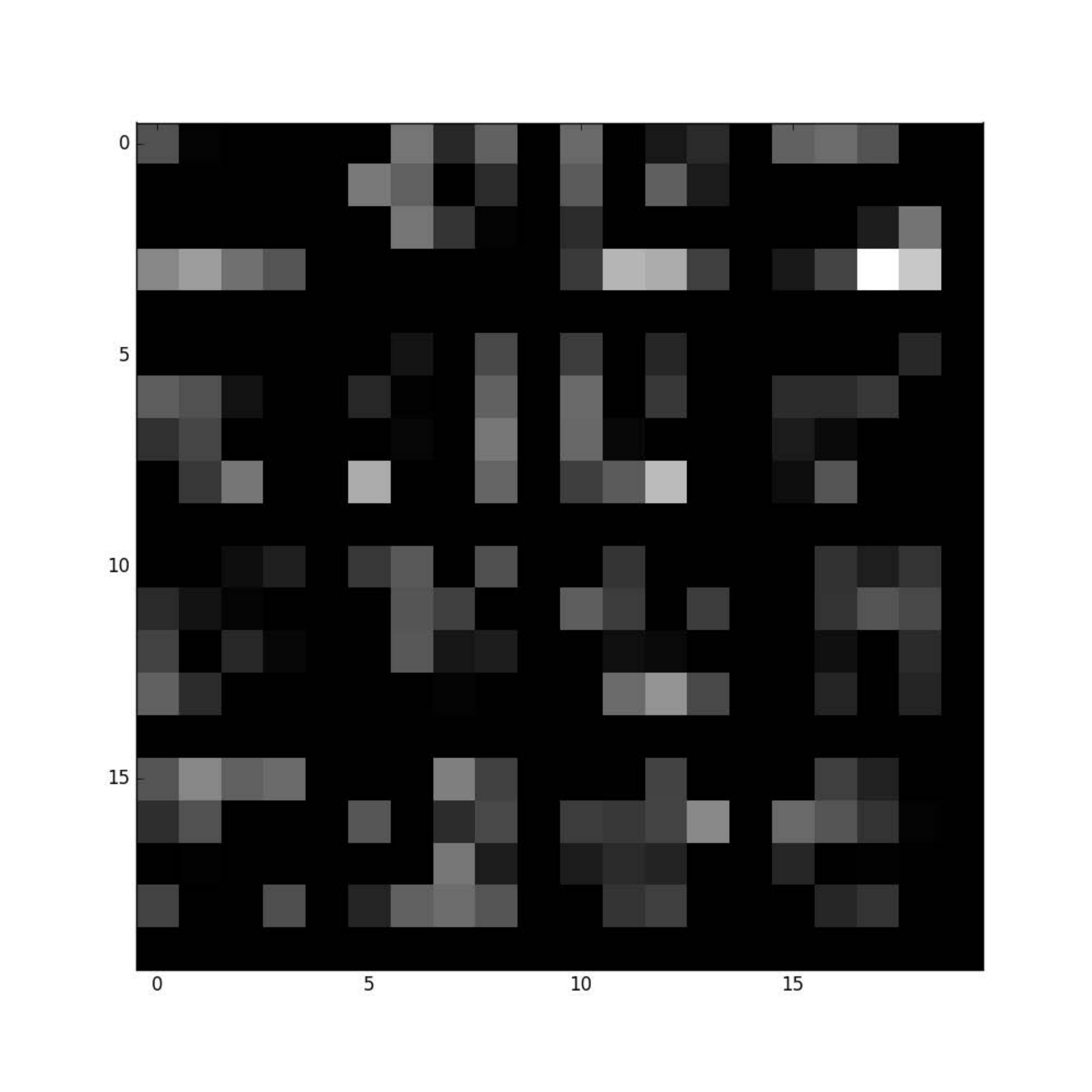}}
\subfigure[d\_c2~]{\includegraphics[width=0.083\textwidth]{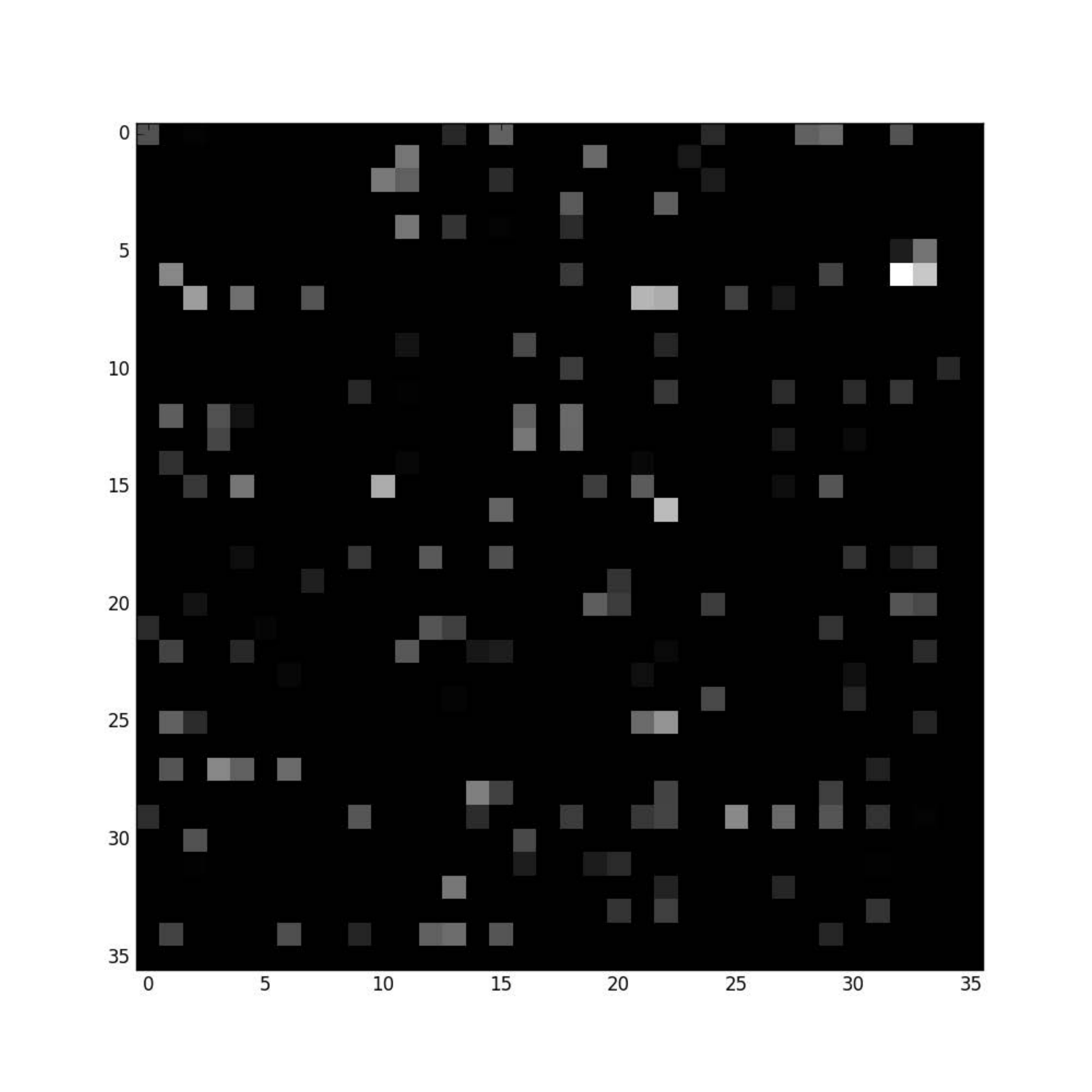}}
\subfigure[d\_p1~]{\includegraphics[width=0.083\textwidth]{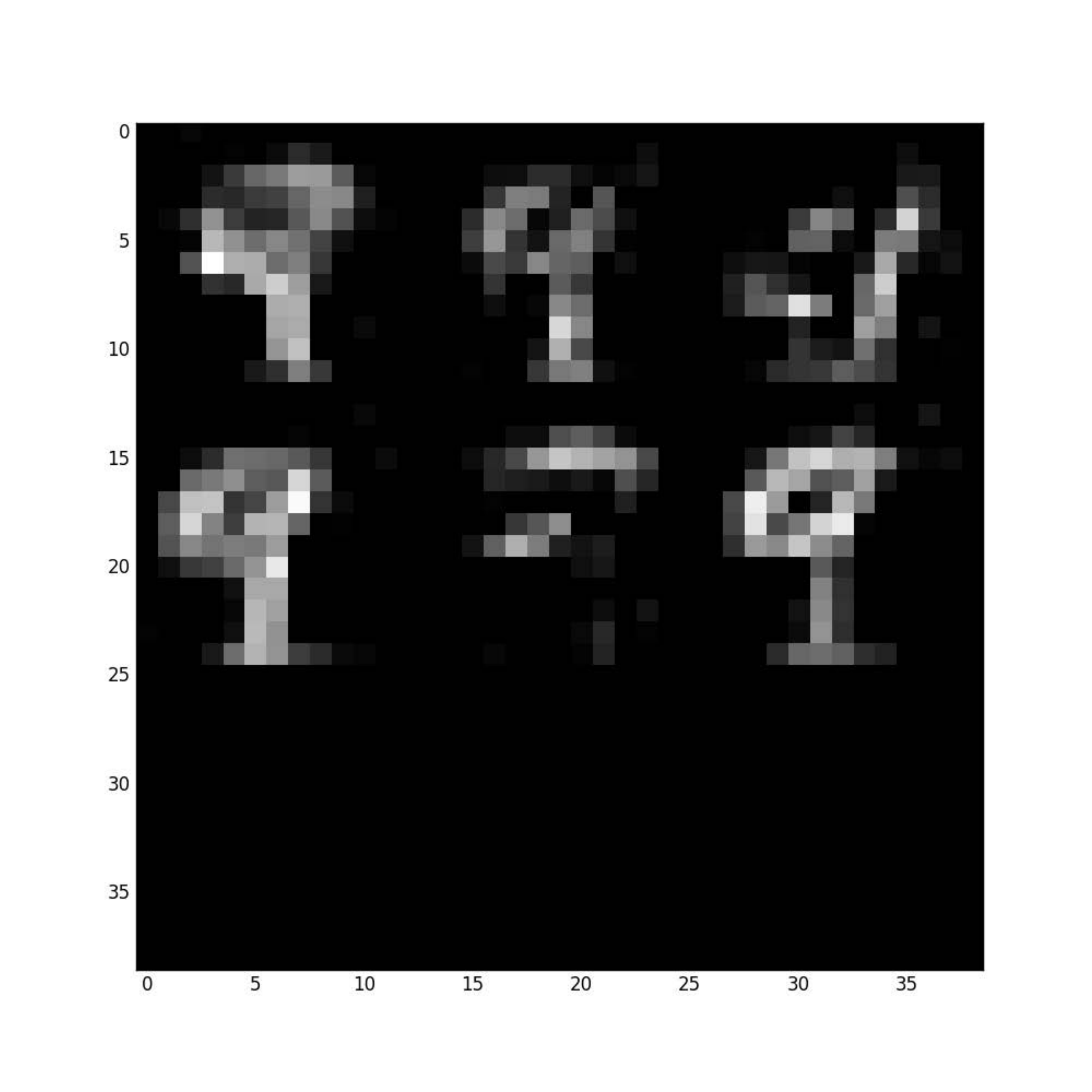}}
\subfigure[d\_c1~]{\includegraphics[width=0.083\textwidth]{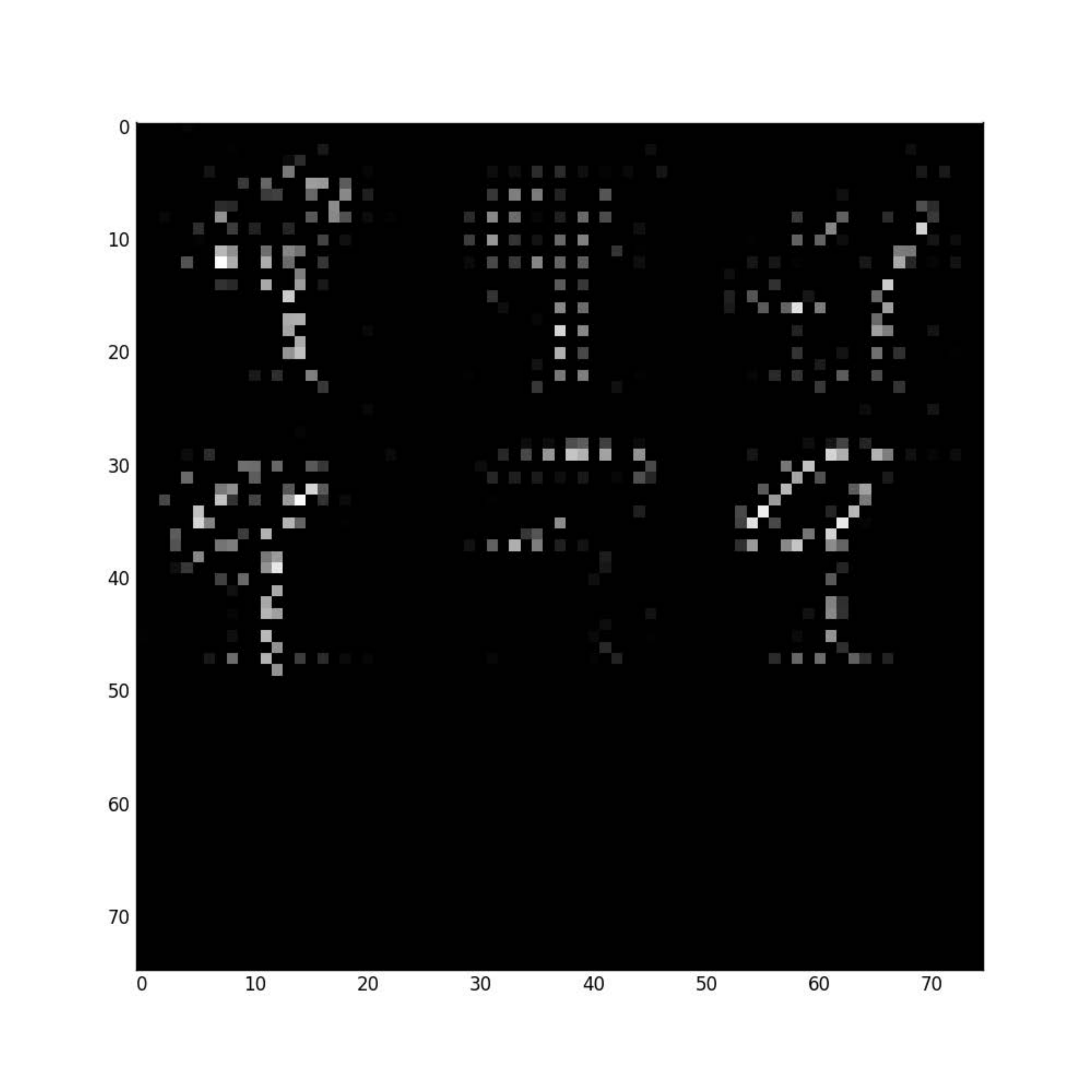}}
\subfigure[out.~]{\includegraphics[width=0.083\textwidth]{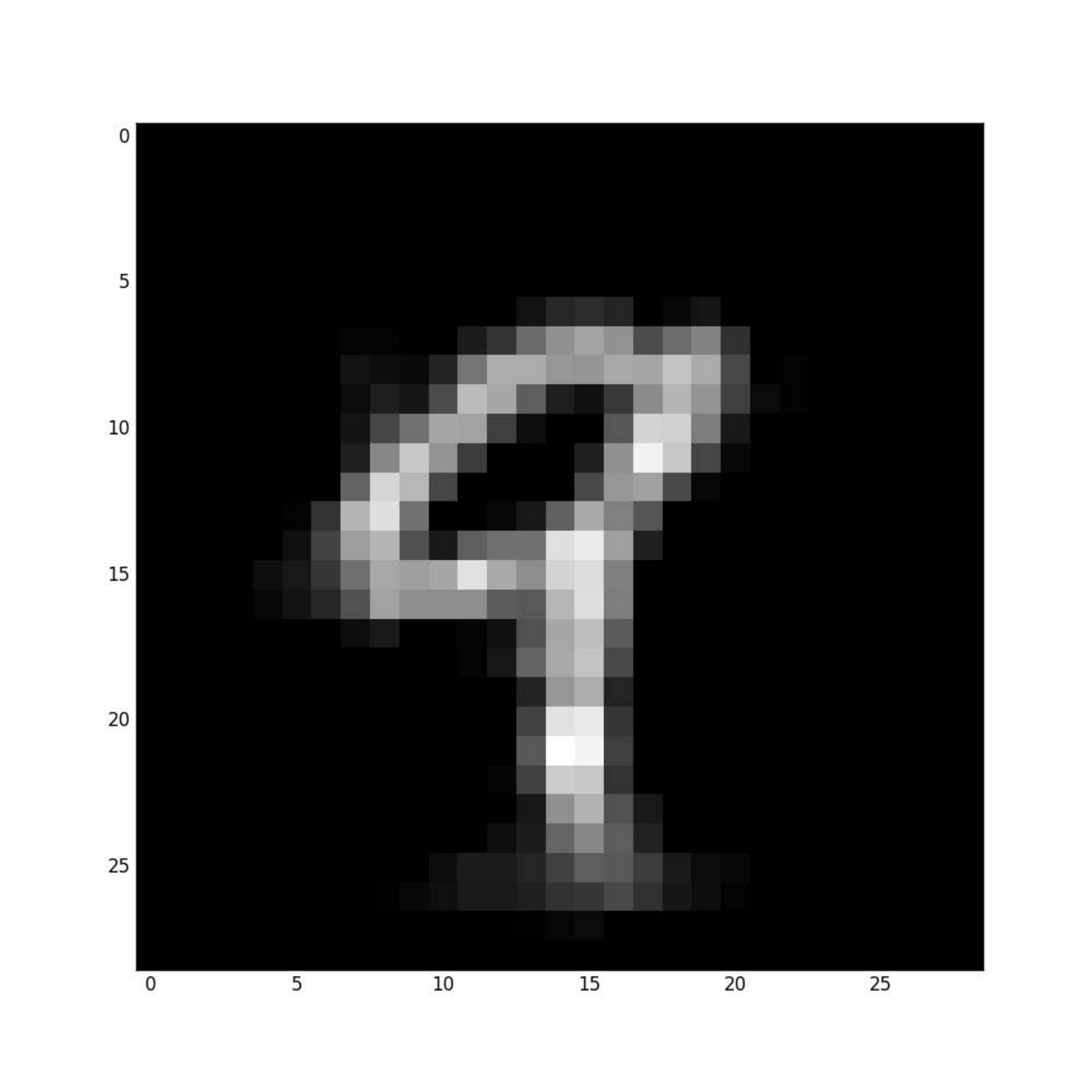}}
\caption{Visualization of the inner activations with respect to digits 1, 5 and 9.}
\label{fig:vis-activation}
\end{figure}

In Fig. \ref{fig:vis-activation}, we visualize the inner activations of FCAE on the MNIST dataset with
three digits: 1, 5, and 9. As shown in the figure, the activations in the feature layer are very sparse.
Besides, the deconvolutional layer gradually recovers details of the pooled feature maps and finally gives a
rough description of the original image. This indicates that FCAE can learn clustering-friendly features and
keep the key information for image reconstruction.

\begin{figure}[!htb]
\centering
\begin{minipage}{0.22\linewidth}
\includegraphics[width=\textwidth]{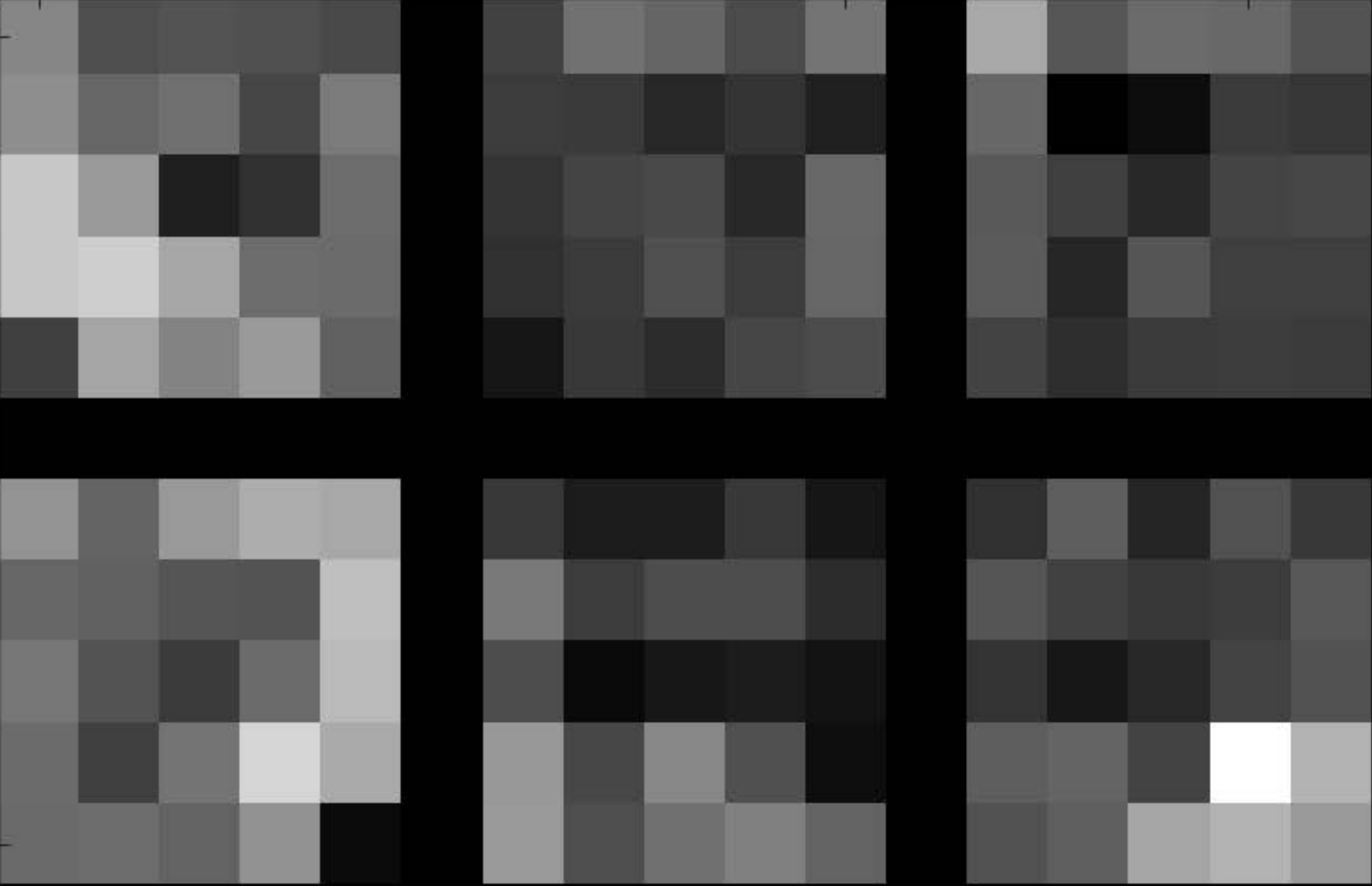}
\end{minipage}\quad
\begin{minipage}{0.22\linewidth}
\includegraphics[width=\textwidth]{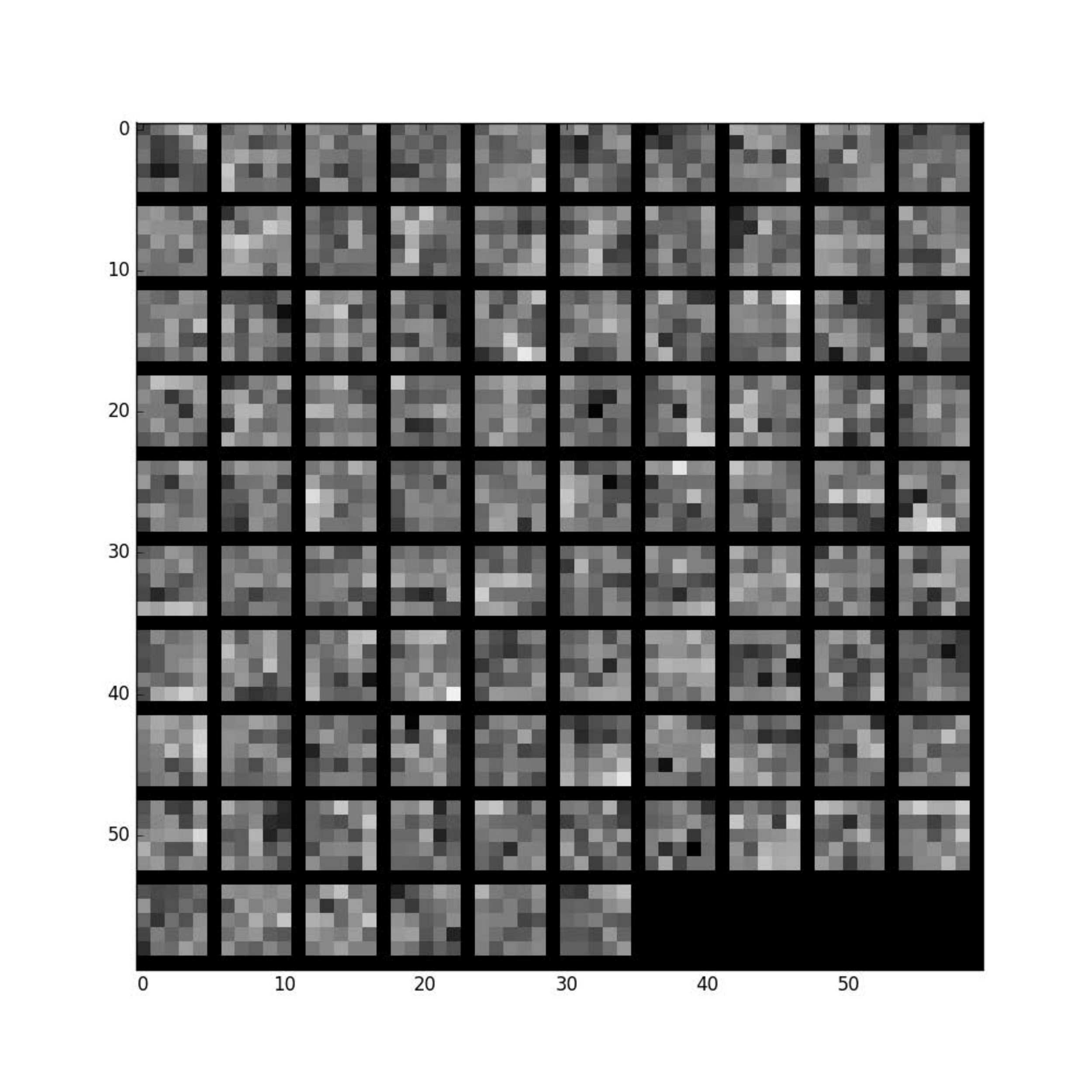}
\end{minipage}\quad
\begin{minipage}{0.22\linewidth}
\includegraphics[width=\textwidth]{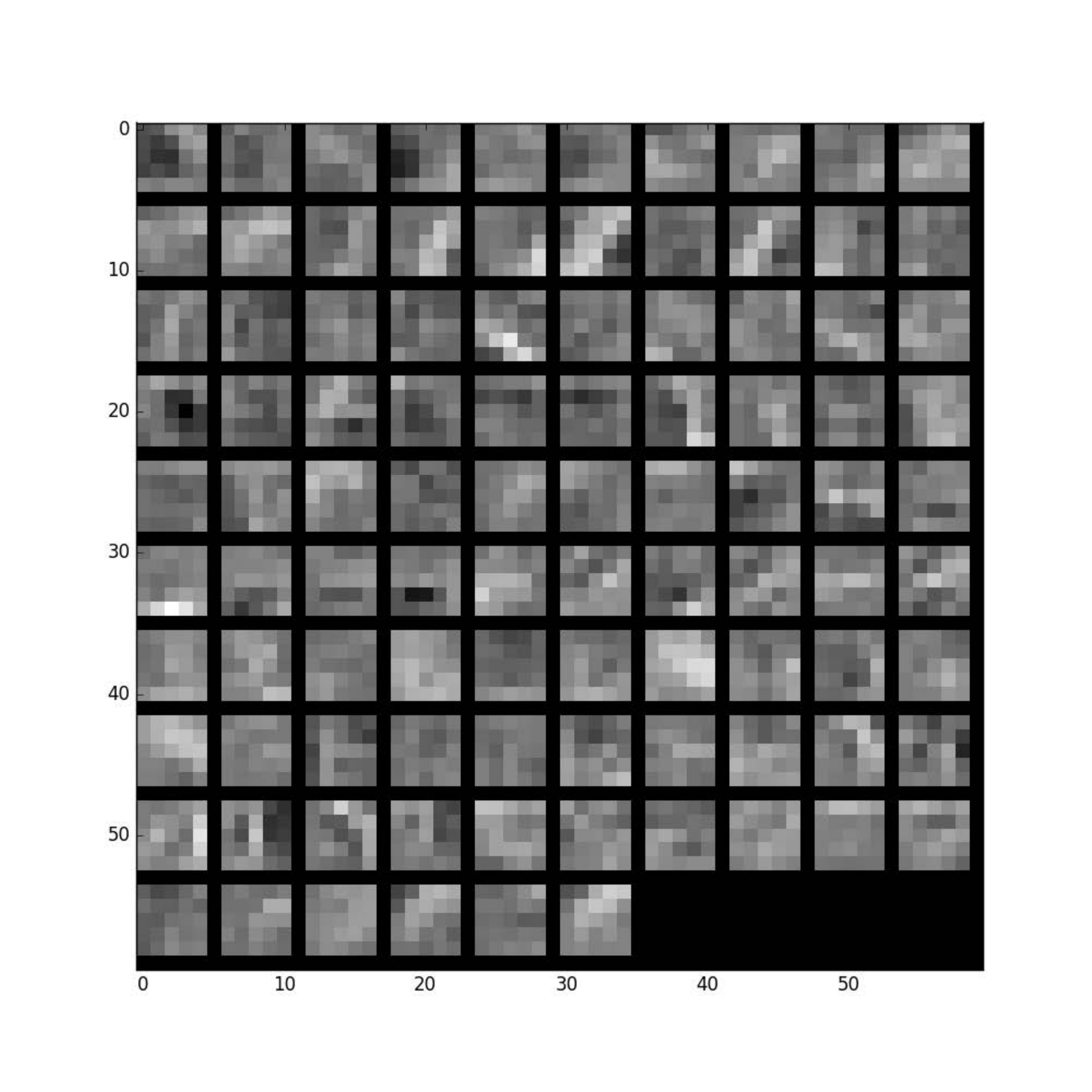}
\end{minipage}\quad
\begin{minipage}{0.22\linewidth}
\includegraphics[width=\textwidth]{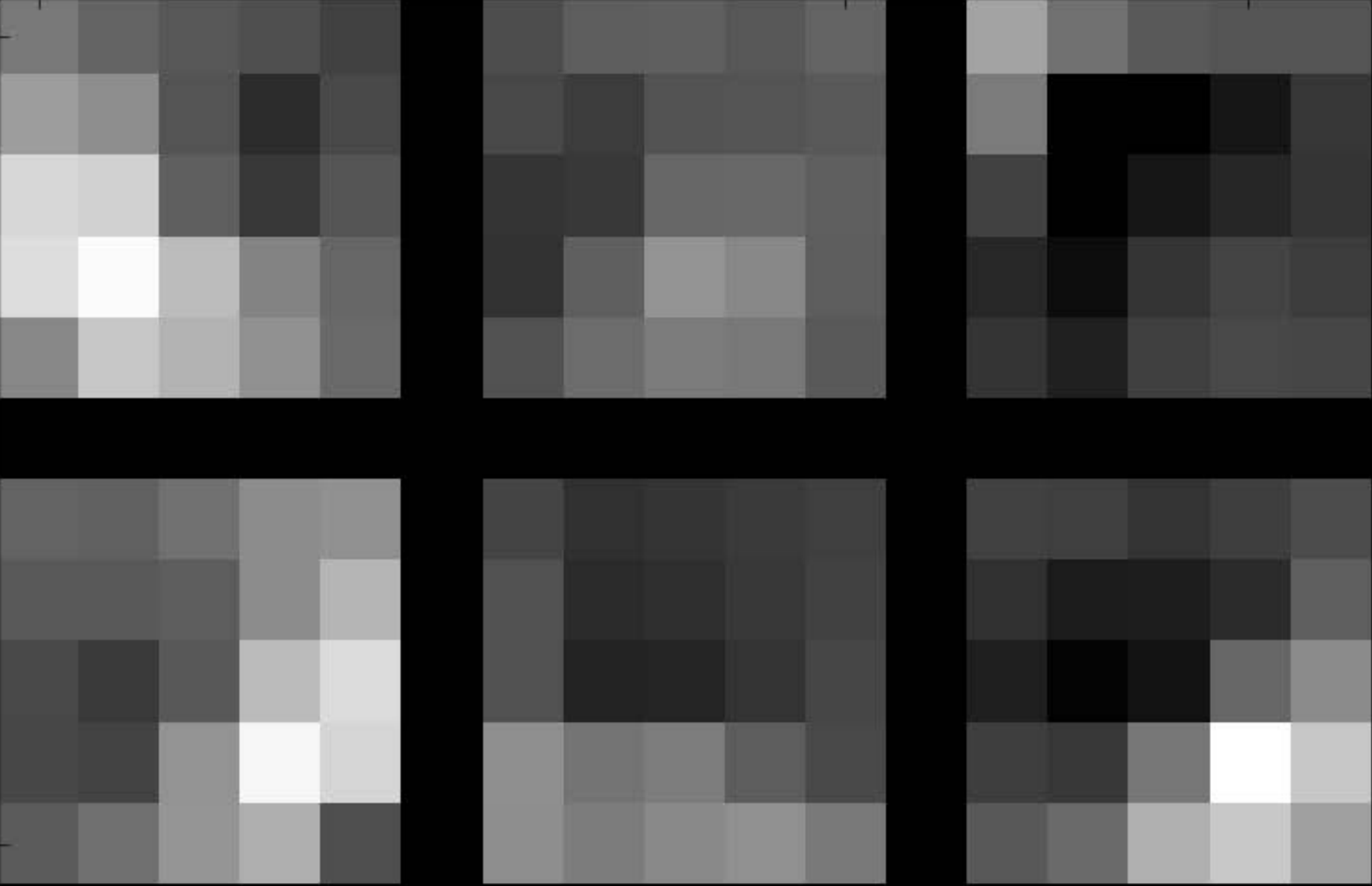}
\end{minipage}
\text{~~~~~~~~~~~~~~~~~~~~~~~~~~~~~~~~~~~~~~~~~~~~~~~~~~~~~~~}
\text{
~~~~~~~~conv1 ~~~~~~~~~~~~~~~ conv2 ~~~~~~~~~~~~~~ d\_conv2 ~~~~~~~~~~~~~~ d\_conv1~~~~~~~
}
\caption{Visualization of the learned FCAE filters.}\label{fig:vis-weights}
\end{figure}

Fig. \ref{fig:vis-weights} visualizes the learned filters of FCAE on the MNIST dataset. It is observed
in \cite{Masci2011SCAE} that the stacked convolutional auto-encoders trained on noisy inputs (30\% binomial noise)
and a max-pooling layer can learn localized biologically plausible filters. However, even without adding noise,
the learned deconvolutional filters in our architectures are non-trivial Gabor-like filters which are visually the
nicest shapes. This is due to the use of max-pooling and unpooling operations. As discussed in \cite{Masci2011SCAE},
the max-pooling layers are elegant way of enforcing sparse codes which are required to deal with the over-complete representations of convolutional architectures.

\subsubsection{Monitoring the learning process}\label{sec:process}

\begin{figure}[!htb]
\centering
\subfigure{
\includegraphics[width=0.18\linewidth]{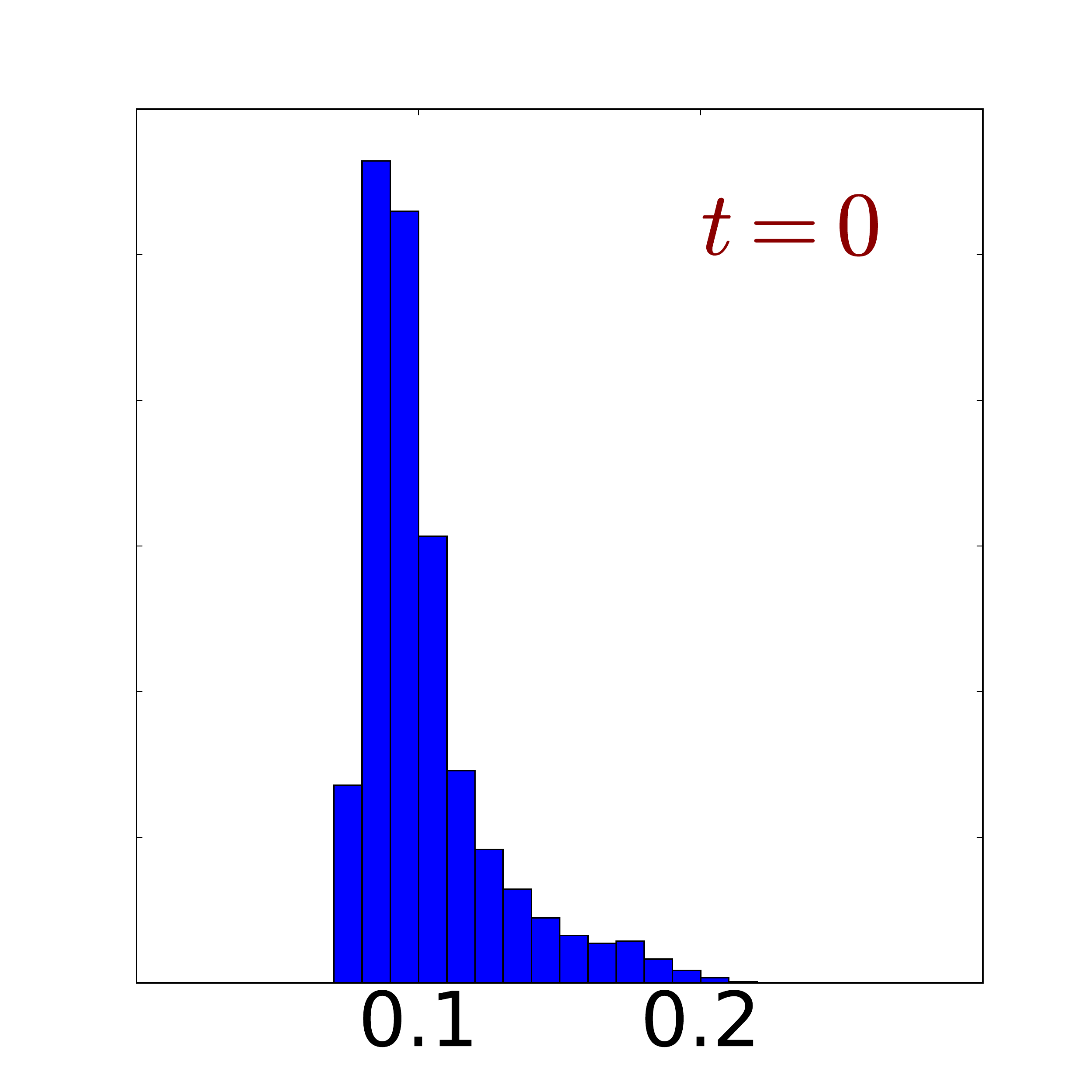}
\includegraphics[width=0.18\linewidth]{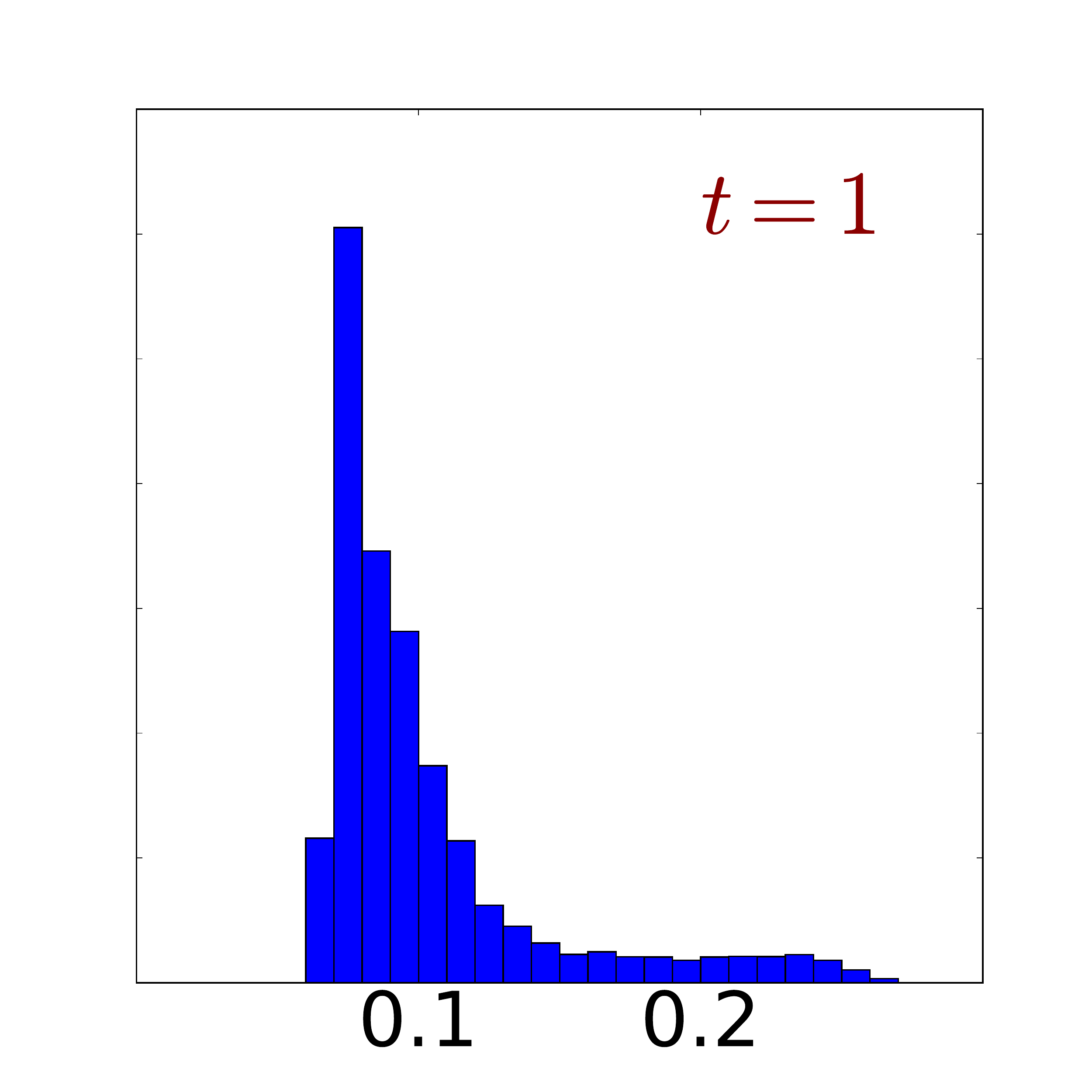}
\includegraphics[width=0.18\linewidth]{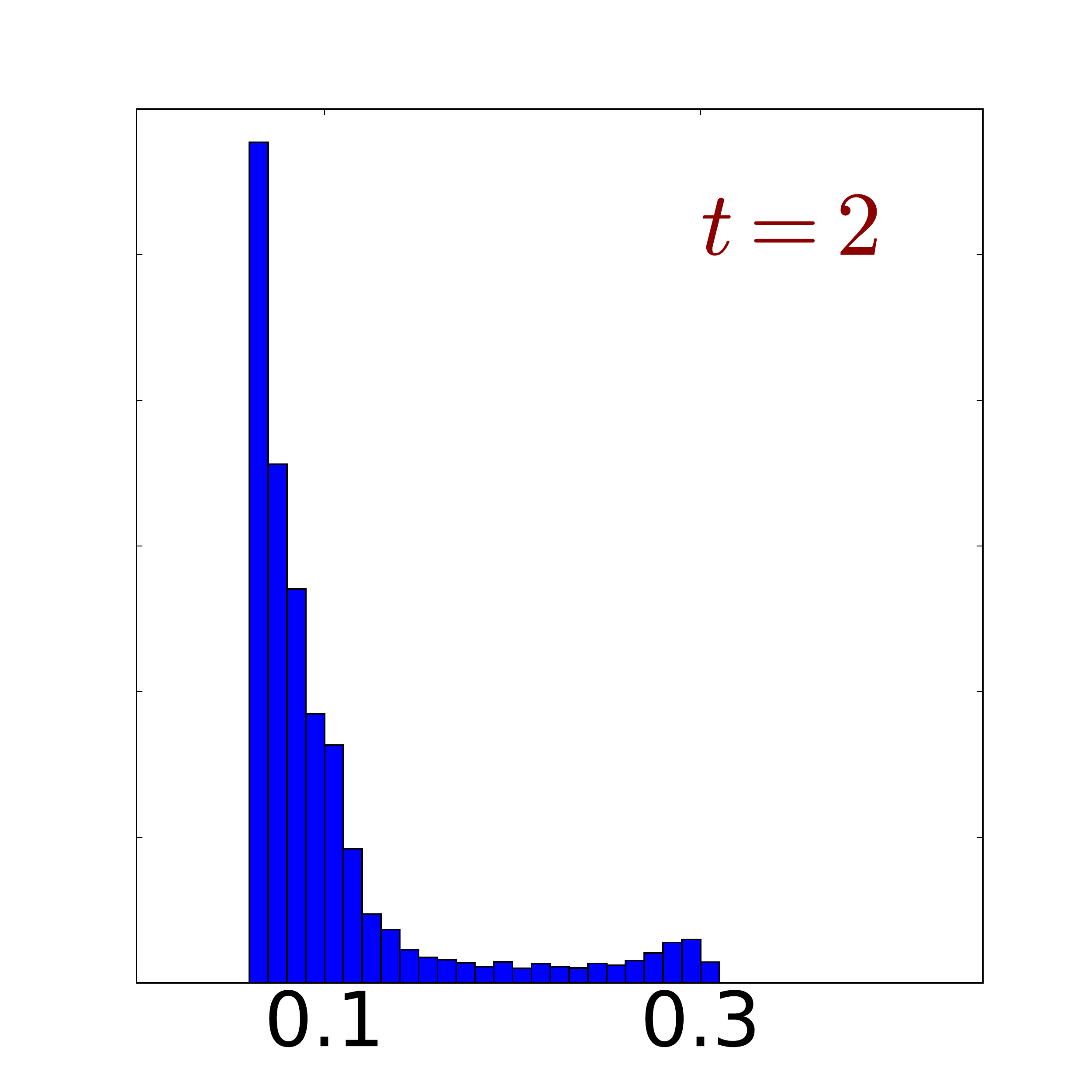}
\includegraphics[width=0.18\linewidth]{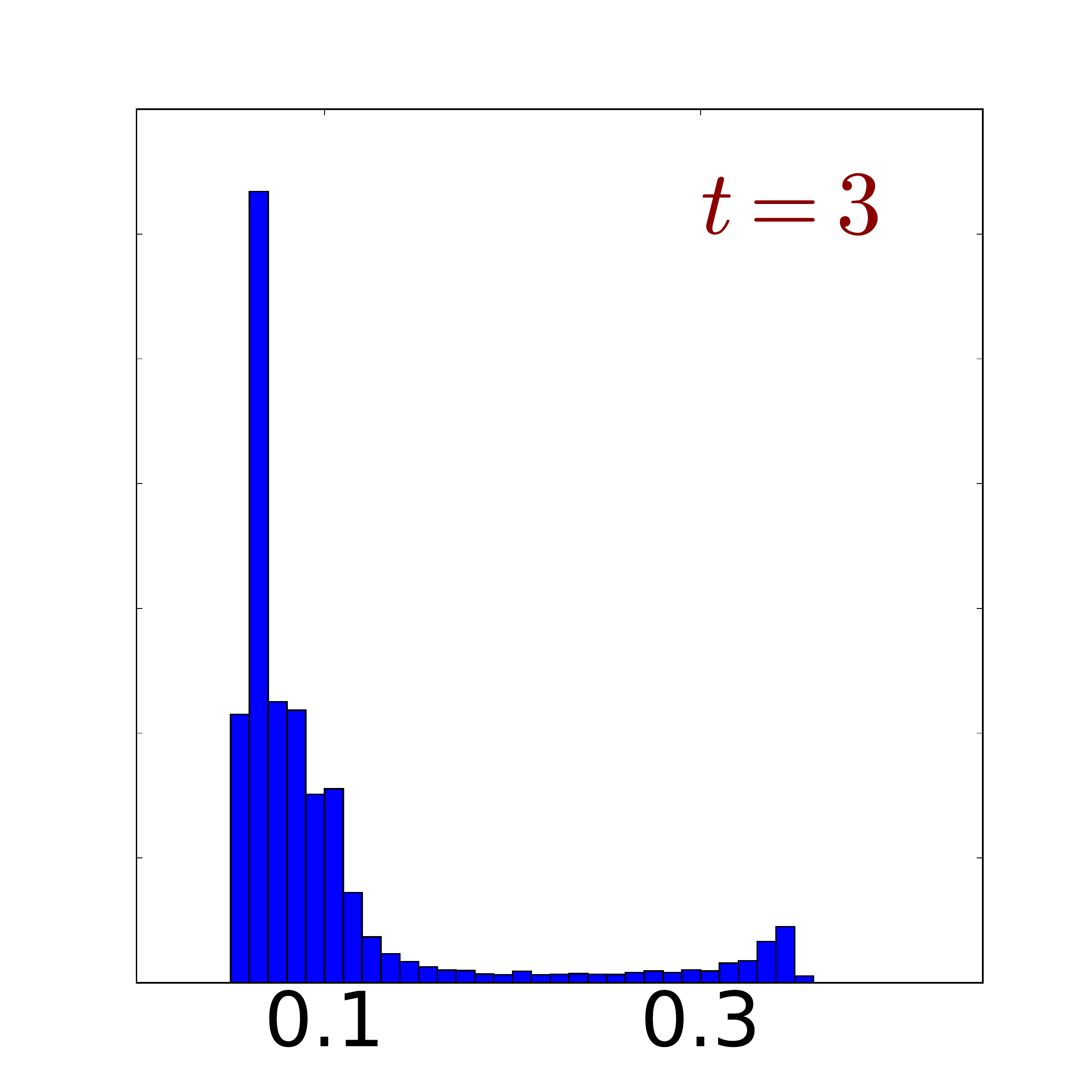}
\includegraphics[width=0.18\linewidth]{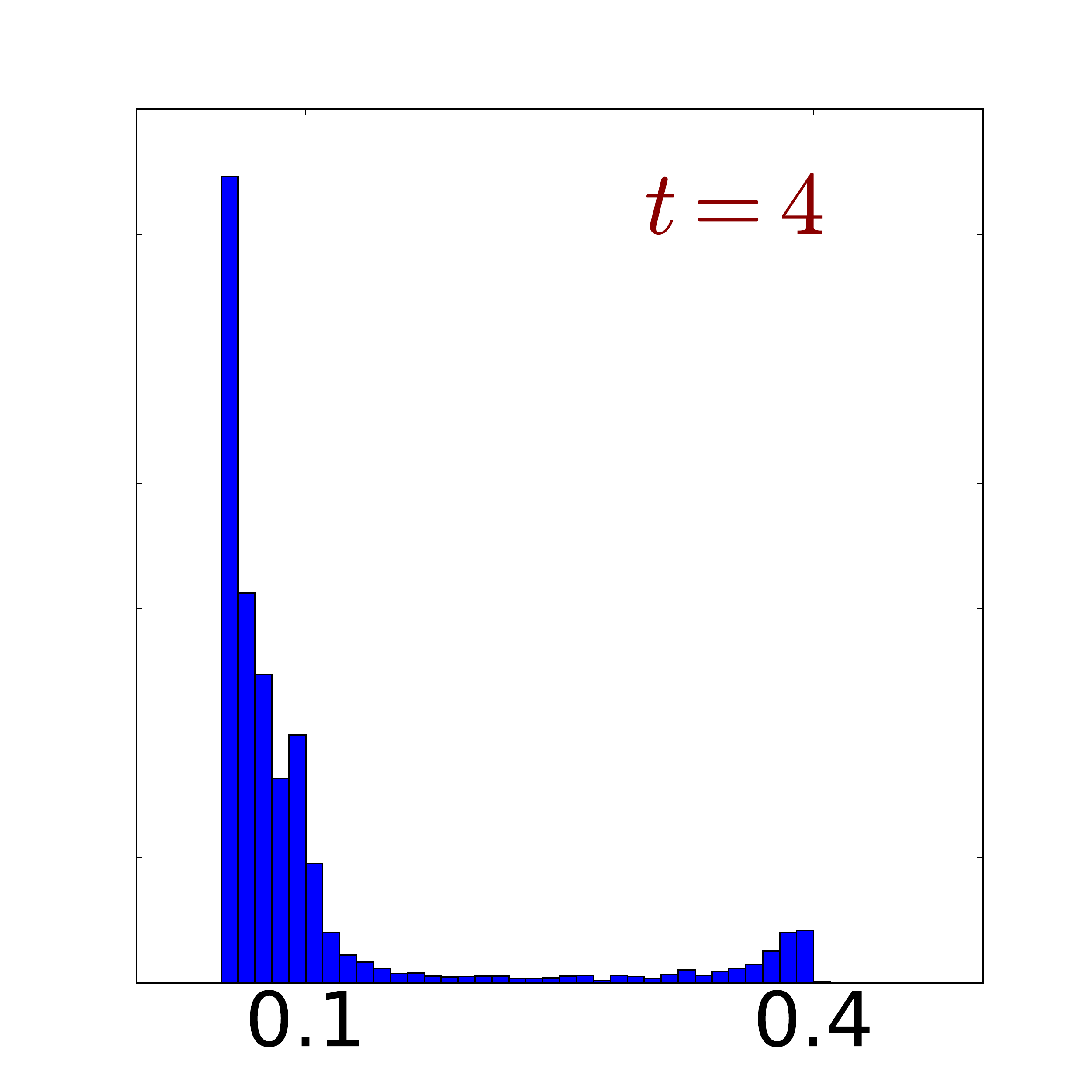}
}\\
\subfigure{
\includegraphics[width=0.18\linewidth]{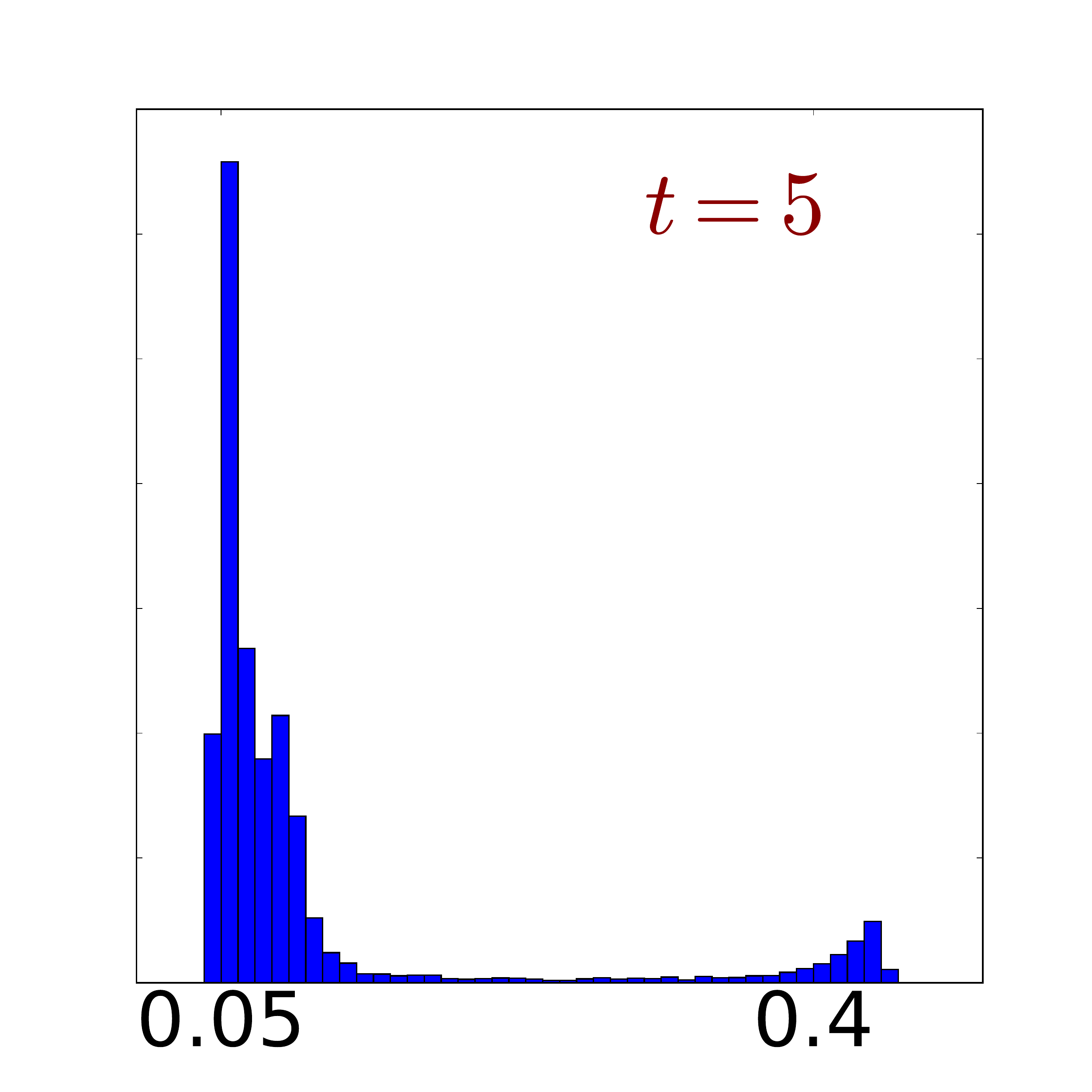}
\includegraphics[width=0.18\linewidth]{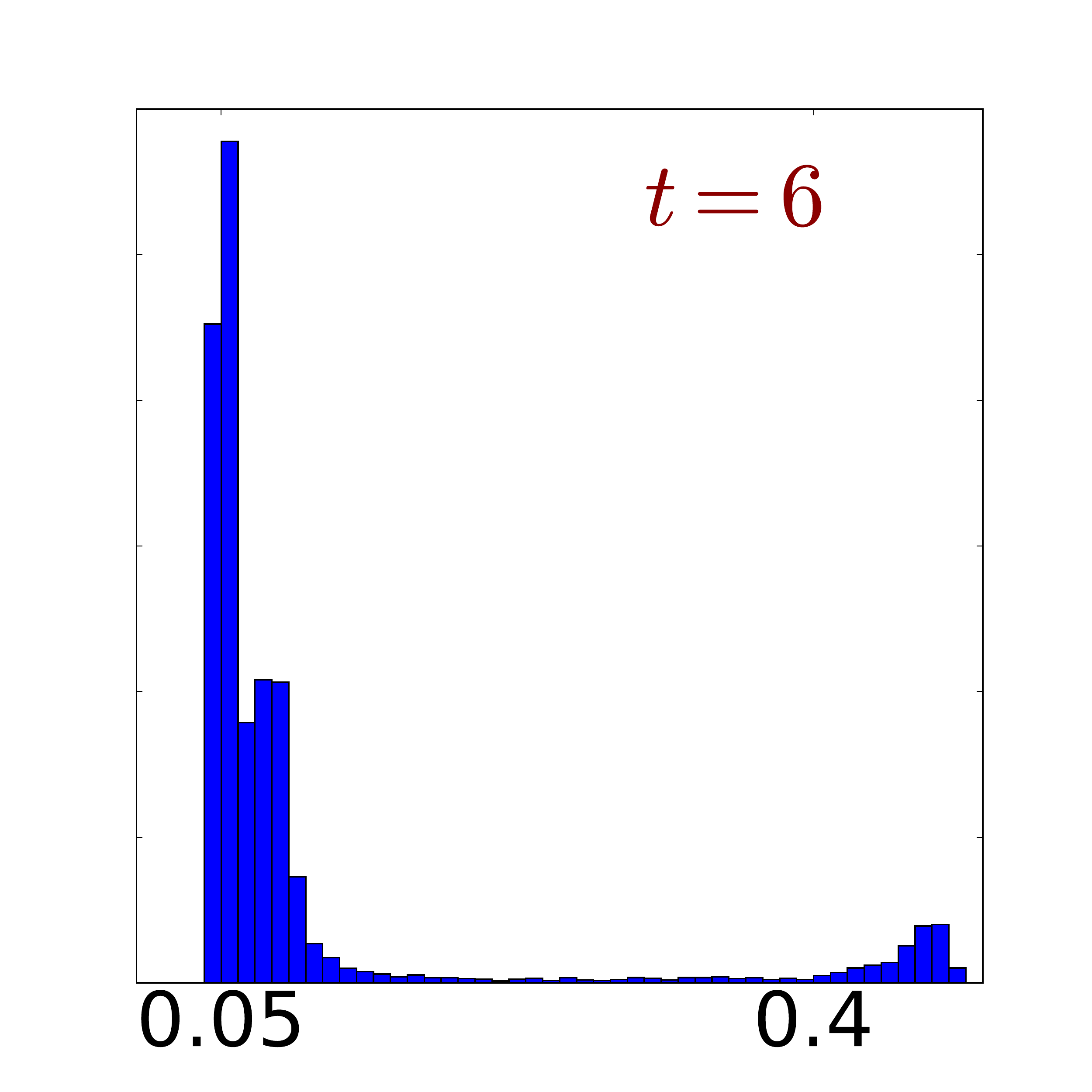}
\includegraphics[width=0.18\linewidth]{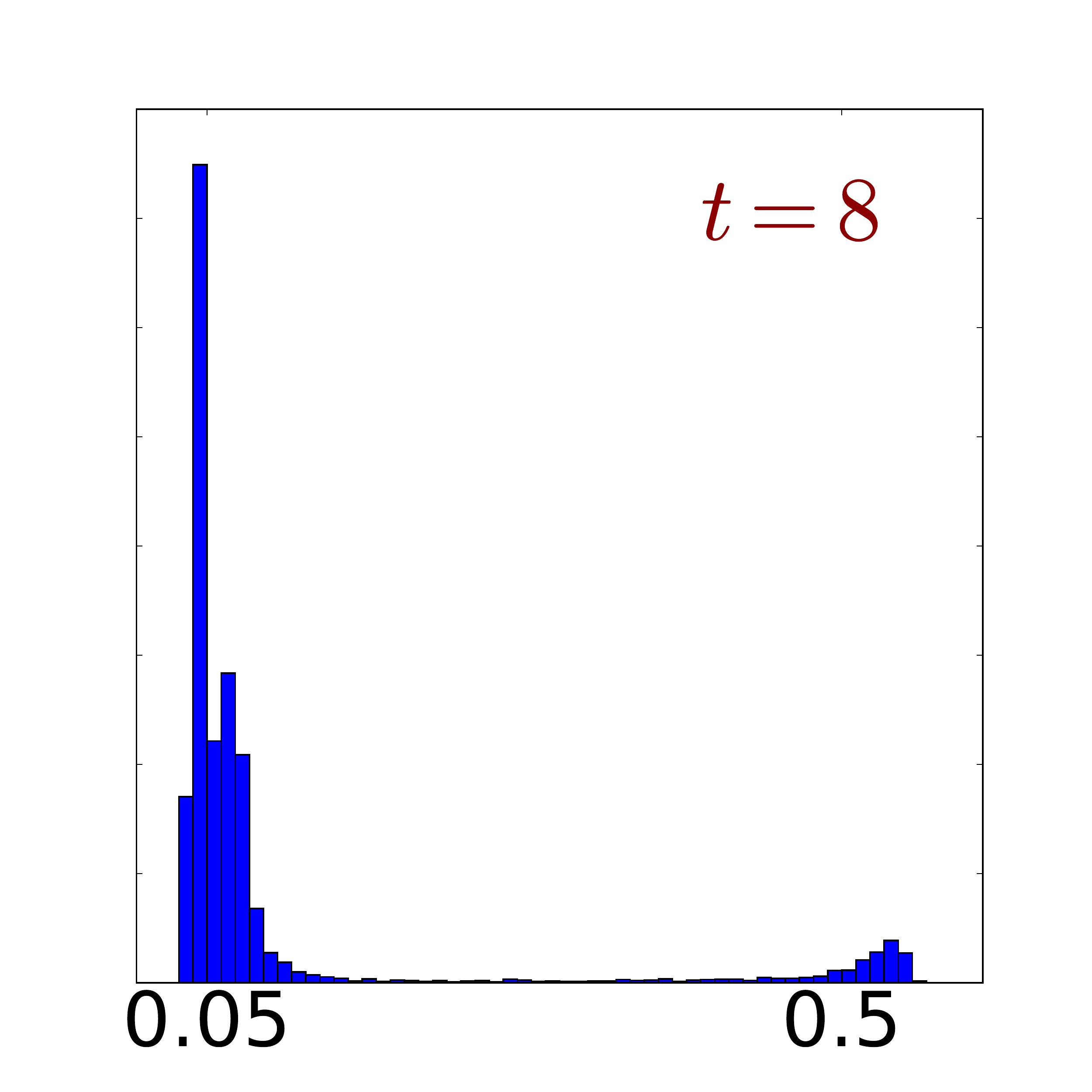}
\includegraphics[width=0.18\linewidth]{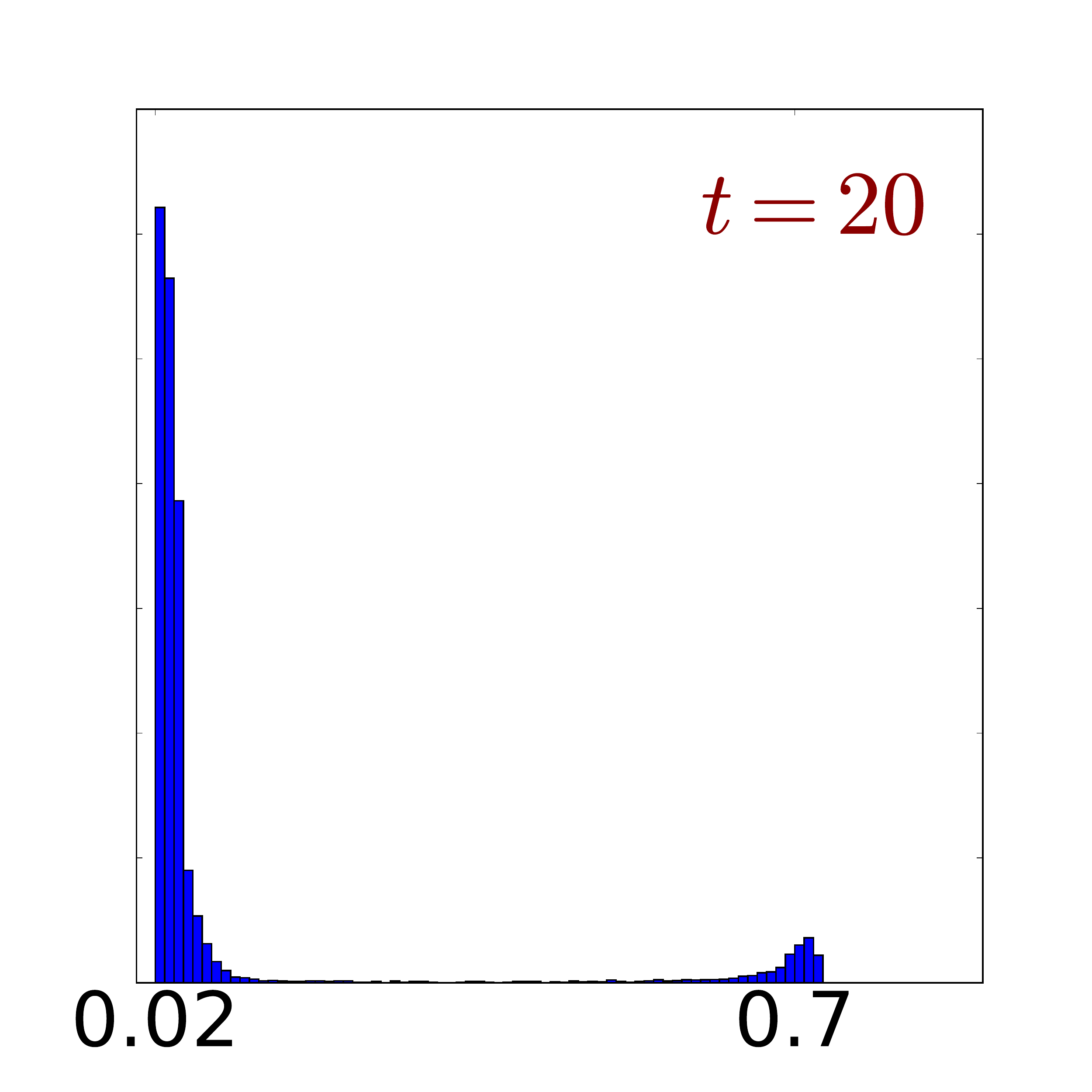}
\includegraphics[width=0.18\linewidth]{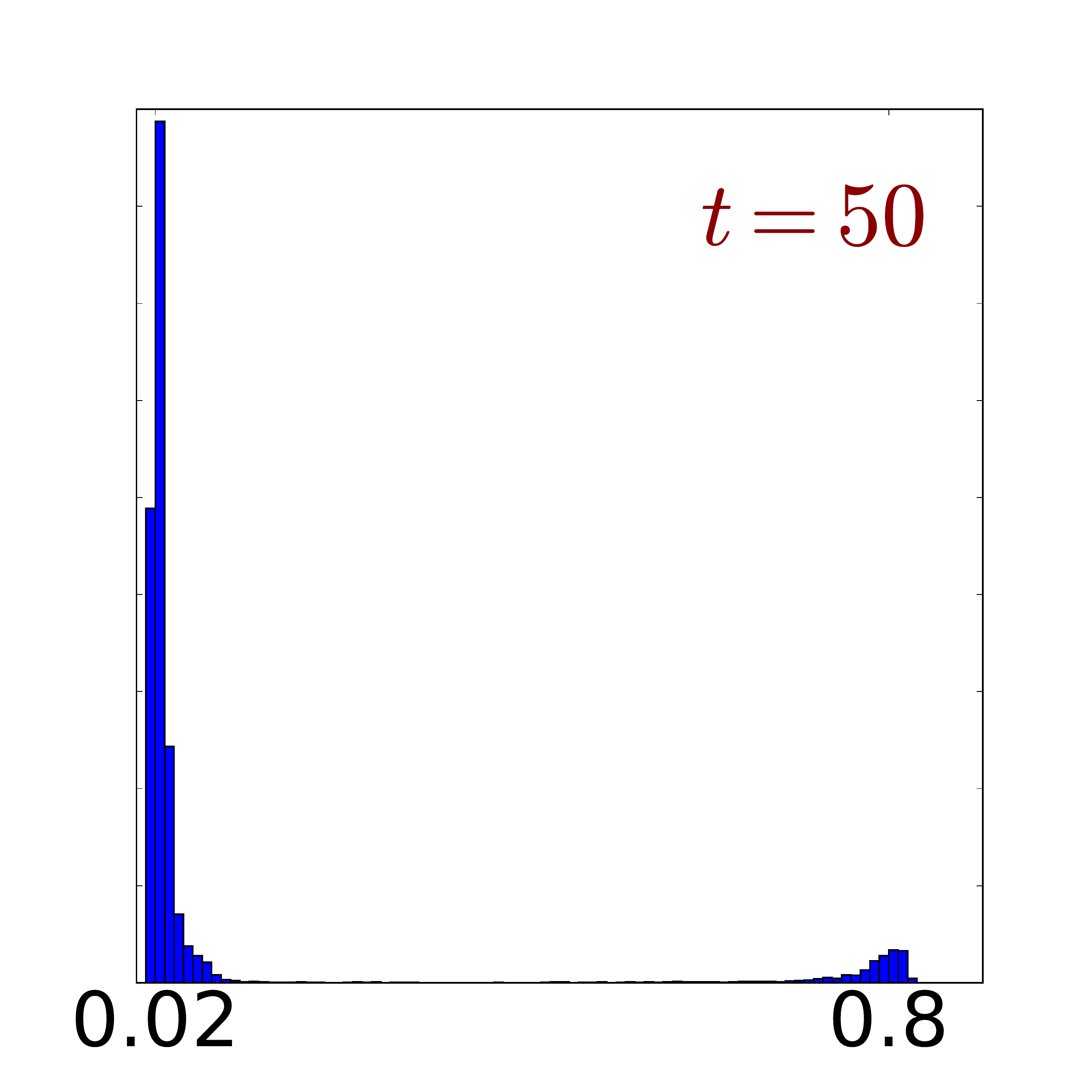}
}
\caption{Visualization of the soft score frequency hists with respect to the first cluster
at different learning stages.}\label{fig:vis-hist}
\end{figure}

We use frequency hist of the soft assignment scores to monitor the learning process of DBC.
Fig. \ref{fig:vis-hist} shows the hists of scores on the MNIST test dataset (a subset of the MNIST dataset
with $10000$ samples). The scores are assigned to the first cluster at different learning epochs.
At early epochs ($t\leq 4$), most of the scores are near $0.1.$ This is a random guess probability because
there are $10$ clusters. As the learning procedure goes on, some higher score samples are discriminatively boosted
and their scores become larger than others. As a result, the cluster tends to ``believe'' in these higher score
samples and consequently make scores of the others to be smaller (approximating zero). Finally, the scores
assigned to the cluster become two-side polarized. Samples with very high scores ($s_{ij}\approx 0.8$) are thought
to definitely belong to the first cluster and the others with very small scores ($s_{ij}\approx 0.02$)
should belong to other clusters.

\subsubsection{Embedding learned features in a low dimensional space}

\begin{figure}[!htb]
\centering
\subfigure[(a) epoch 0]{
\includegraphics[height=2.3cm]{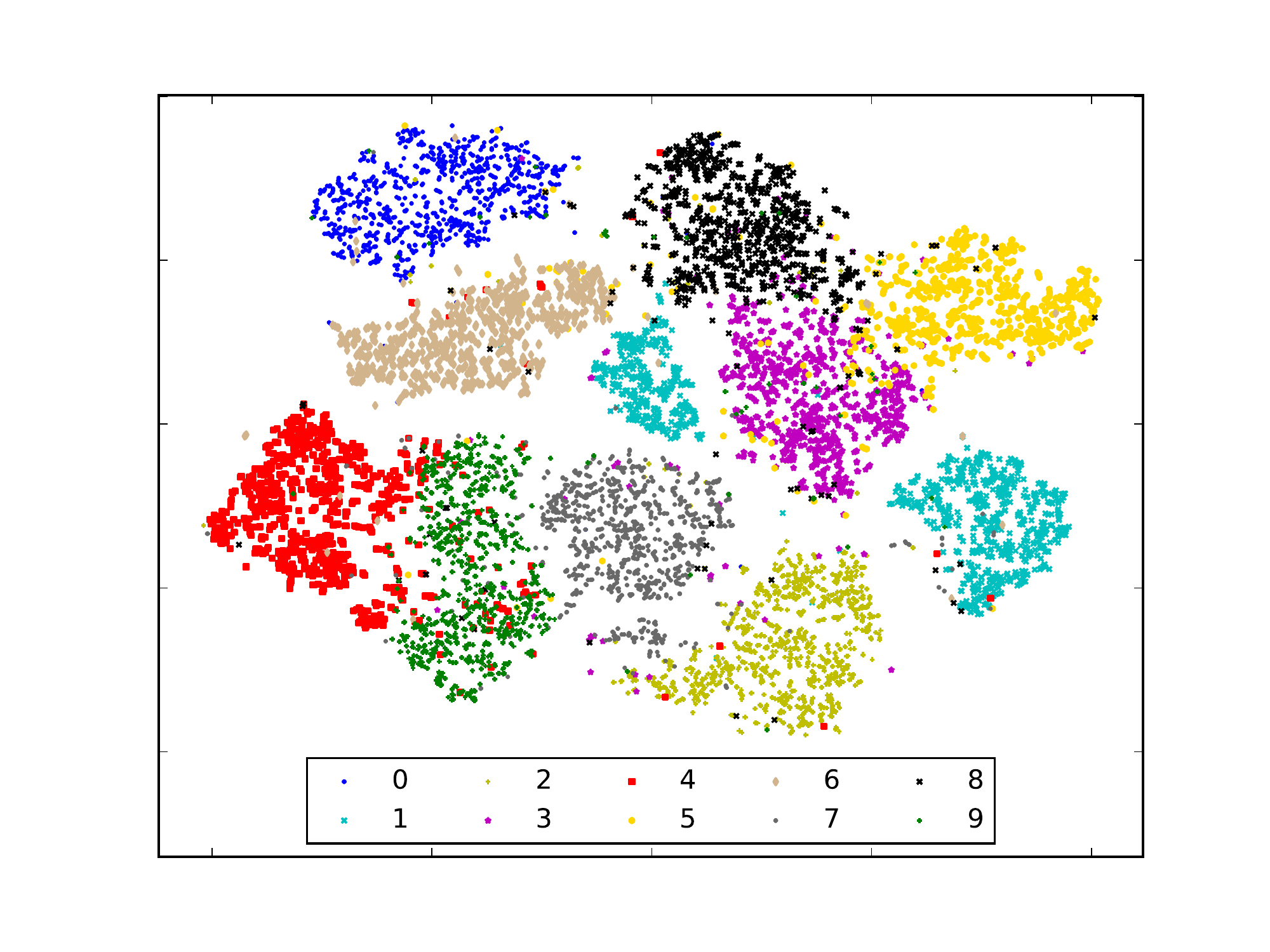}
}
\subfigure[(b) epoch 2]{
\includegraphics[height=2.3cm]{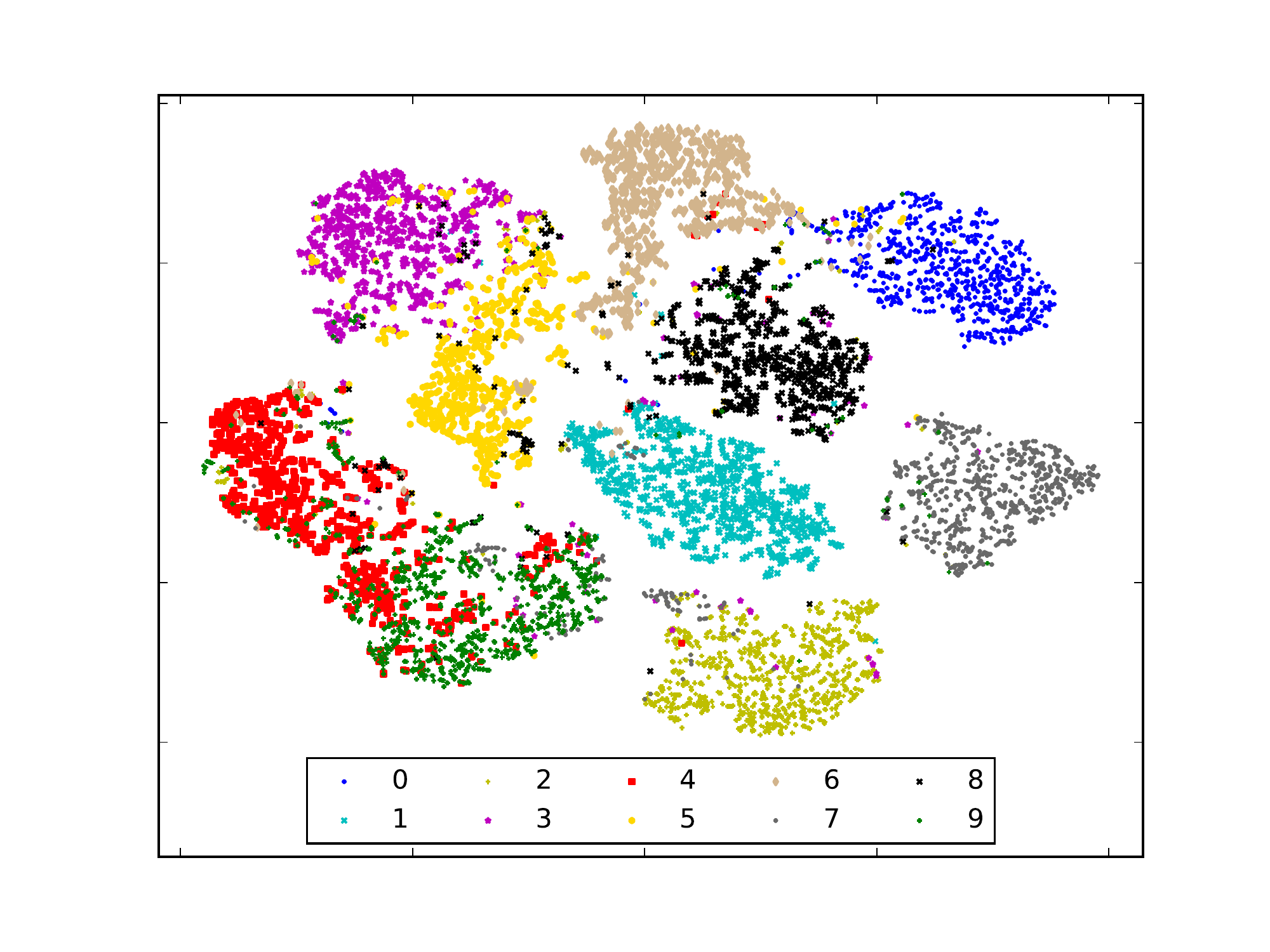}
}
\subfigure[(c) epoch 5]{
\includegraphics[height=2.3cm]{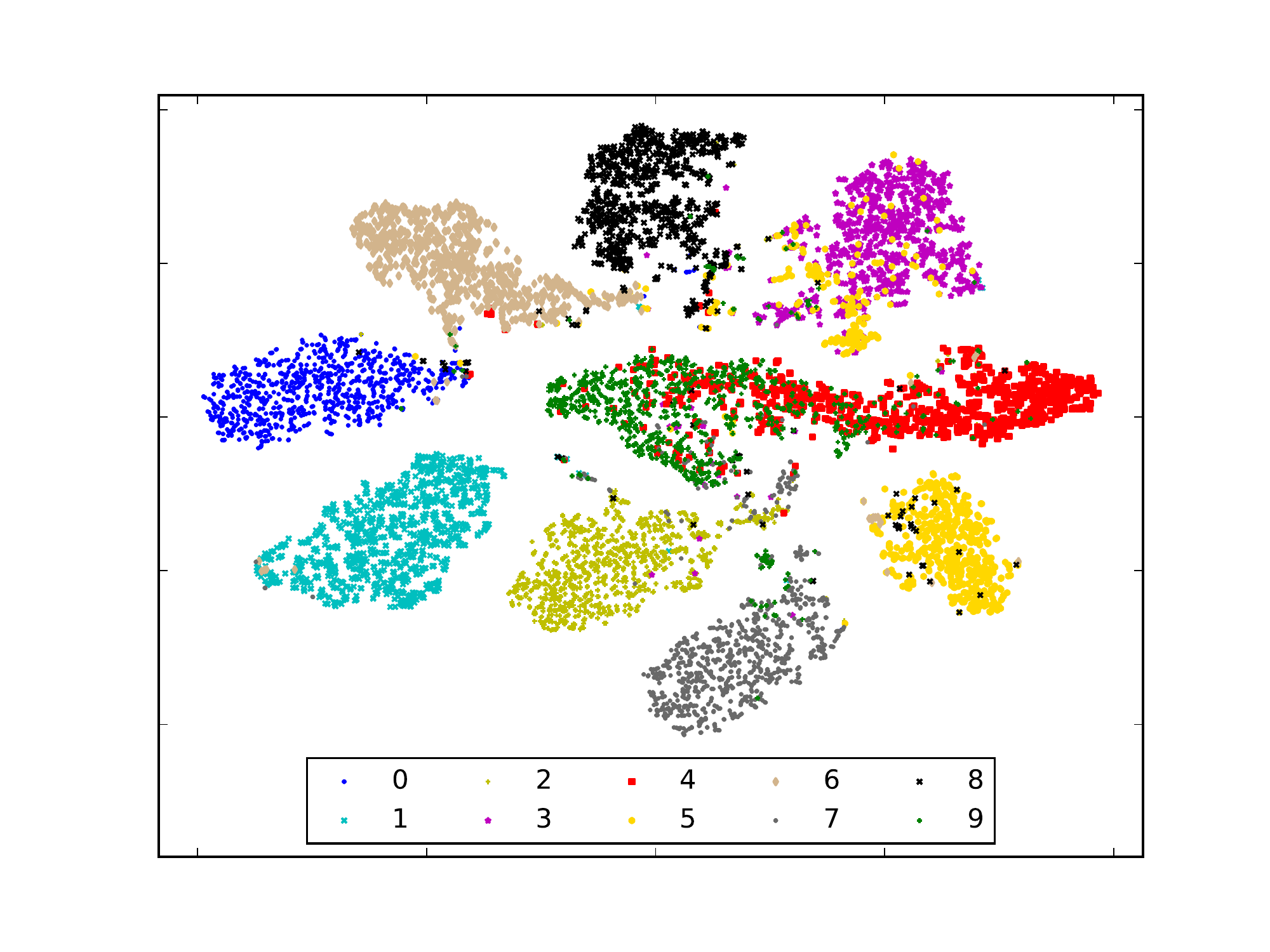}
}
\subfigure[(d) epoch 50]{
\includegraphics[height=2.3cm]{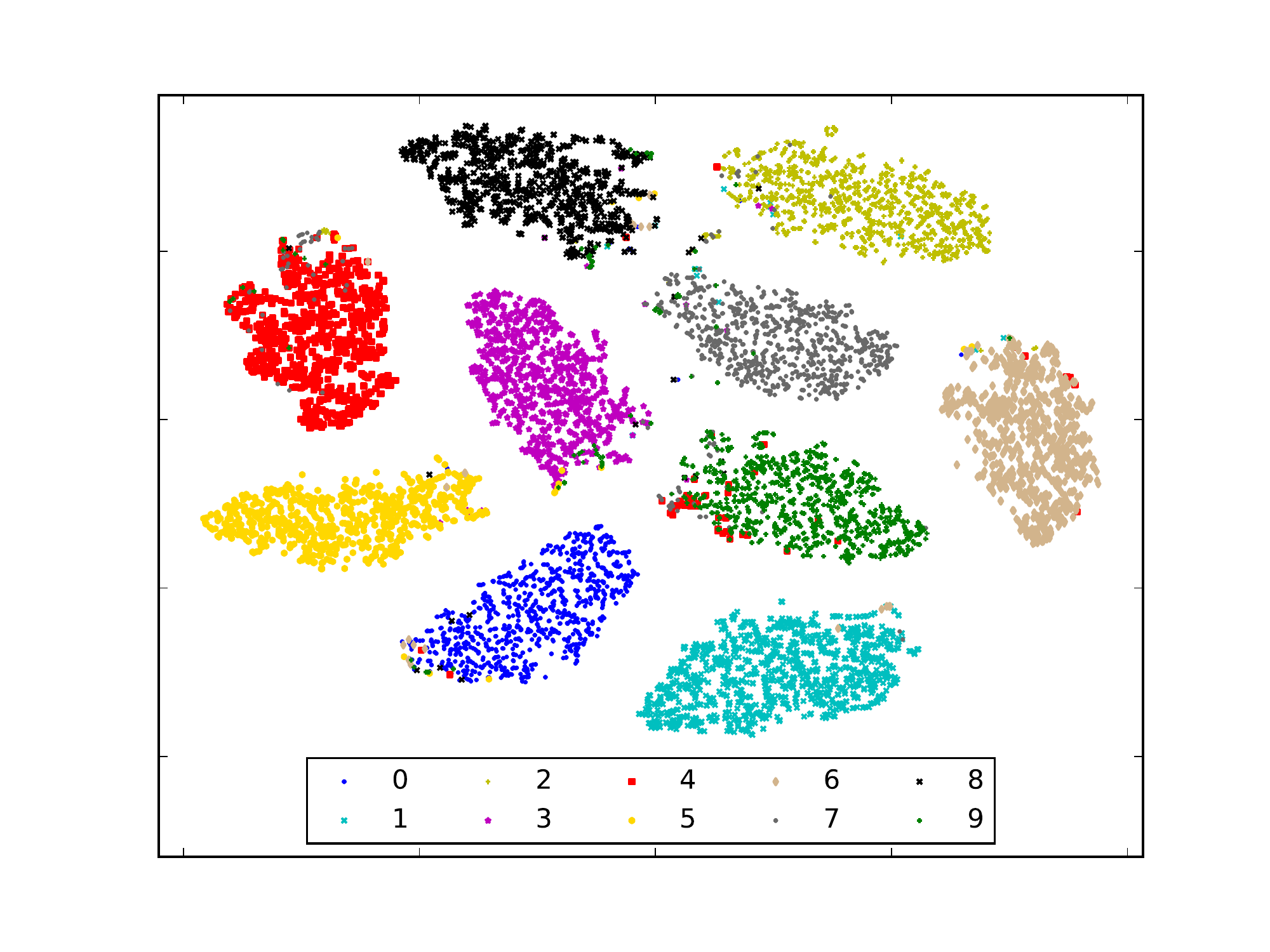}
}
\caption{Visualization of the embedded features in a two-dimensional space with $t$-SNE.}
\label{fig:vis-tsne}
\end{figure}

We visualize the distribution of the learned features in a two-dimensional space with $t$-SNE \cite{Maaten2008tSNE}.
Fig. \ref{fig:vis-tsne} shows the embedded features of the MNIST test dataset at different epochs.
At the initial epoch, the features learned with FCAE are not very discriminative for clustering.
As shown in Fig. \ref{fig:vis-tsne}(a), the features of digits 3, 5, and 8 are closely related.
The same thing happened with digits 4, 7, and 9. At the second epoch, the distribution of the learned features
becomes much compact locally. Besides, the features of digit 7 become far away from those of digits 4 and 9.
Similarly, the features of digit 8 get far away from those of digits 3 and 5. As the learning procedure goes on,
the hardest digits (4 v.s. 9, 3 v.s. 5) for categorization are mostly well categorized after enough discriminative
boosting epochs. The observation is consistent with the results showed in Subsection \ref{sec:process}.

\subsubsection{Visualization of falsely categorized examples}

\begin{figure}[!htb]
\centering
\includegraphics[width=0.7\linewidth]{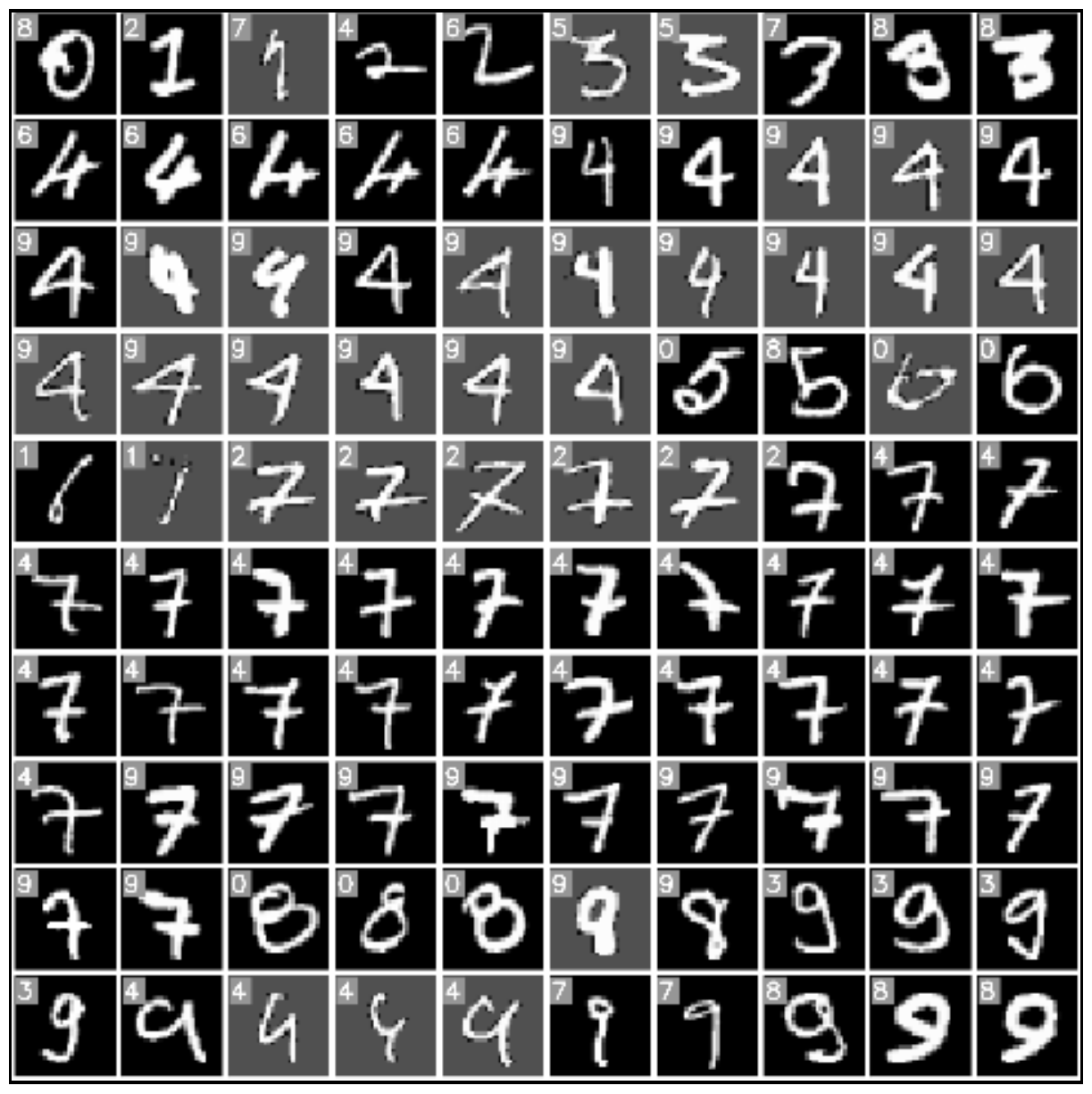}
\caption{Visualization of falsely categorized high score samples (top-100).
The number in the top-left corner is the clustering label which is generated by the Hungarian algorithm.
}\label{fig:vis-false}
\end{figure}

In Fig. \ref{fig:vis-false}, we show the top $100$ falsely categorized examples whose maximum soft assignment scores
are over $0.6.$ It can be observed that it is very hard to distinguish between some ground truth digits 4, 7 and 9
even with human experience. Lots of digits 7 are written with transverse lines in their middle space and would be
thought to be ambiguous for the clustering algorithm. Besides, some ground truth images are themselves confusing,
such as those showed with the gray background.

\subsection{Discussions}

In this section, we make some ablation studies on the learning process with respect to different boosting
factors ($\alpha$), different normalization methods ($n_j$) and different initialization models generated by FCAE.

\subsubsection{Impact of the boosting factor $\alpha$}

Fig. \ref{fig:ablations}(a) shows the ACC and NMI curves, where $\alpha$ equals to $1.5,2,4.$ With a small $\alpha$
($\alpha=1.5$), the learning process is very slow and takes very long time to terminate. On the contrary,
when the factor is set to be very large ($\alpha=4$), the learning process is very fast at the initial stages.
However, this could result in falsely boosting some scores of the ambiguous samples. As a consequence,
the model learned too much from some false information so the performance is not so satisfactory.
With a moderate boosting factor ($\alpha=2$), the ACC and NMI curves grow reasonably and progressively.

\subsubsection{Impact of the balance normalization}

In DEC \cite{Xie2015DEC}, the authors pointed out that the balance normalization plays an important role in preventing
large clusters from distorting the hidden feature space. To address this issue, we compare three normalization strategies:
1) the constant normalization for comparison, that is, $n_j=1$, 2) the normalization by dividing the sum of the original
soft assignment score per cluster, that is, $n_j=\sum_i s_{ij}$, which is adopted in DEC, and 3) the normalization by
dividing the sum of the boosted soft assignment score per cluster, that is, $n_j=\sum_i s_{ij}^\alpha$.
Fig. \ref{fig:ablations}(b) presents the value curves of ACC and NMI against the epoch with these settings.
Initially, the normalization does not affect ACC and NMI very much. However, the constant normalization can easily
get stuck at early stages. The normalization by dividing $n_j=\sum_i s_{ij}$ has certain power of preventing
the distortion. Our normalization strategy gives the best performance compared with the previous methods.
This is because our normalization directly reflects the changes of the boosted scores.

\subsubsection{Impact of the FCAE initialization}

\begin{figure}[!thb]
\centering
\subfigure[(a) boosting factor]{
\includegraphics[width=0.3\linewidth]{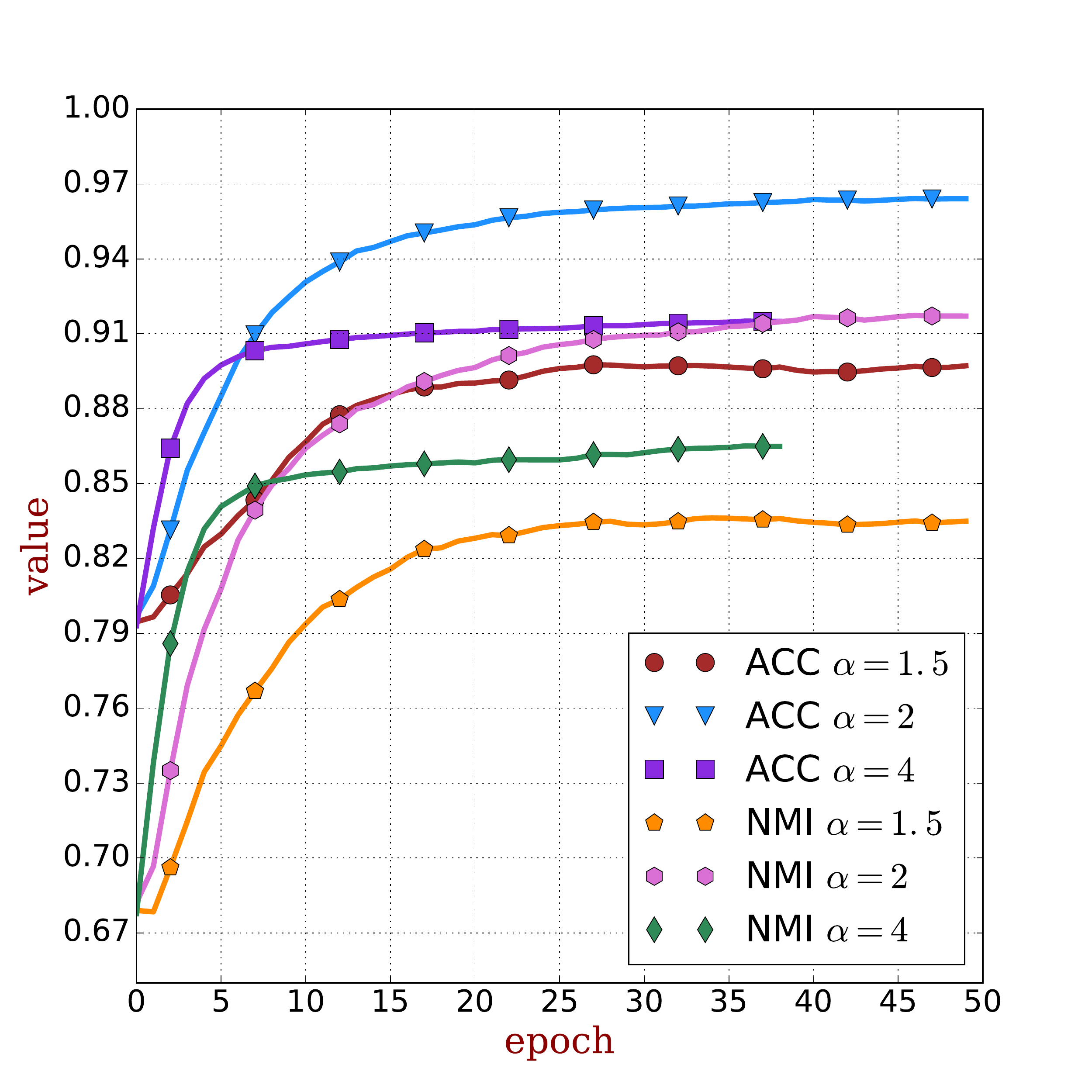}
}
\subfigure[(b) balance normalization]{
\includegraphics[width=0.3\linewidth]{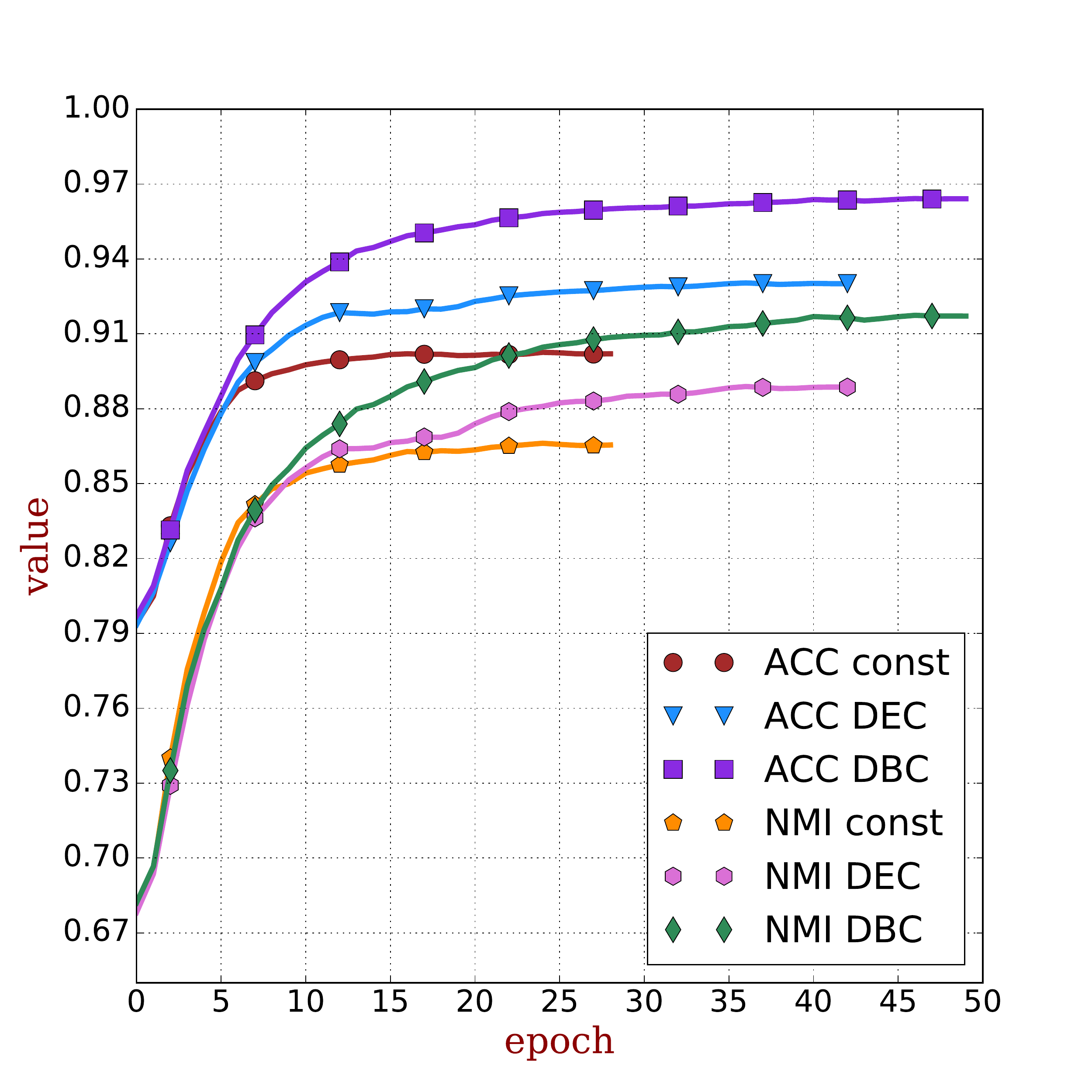}
}
\subfigure[(c) FCAE initialization]{
\includegraphics[width=0.3\linewidth]{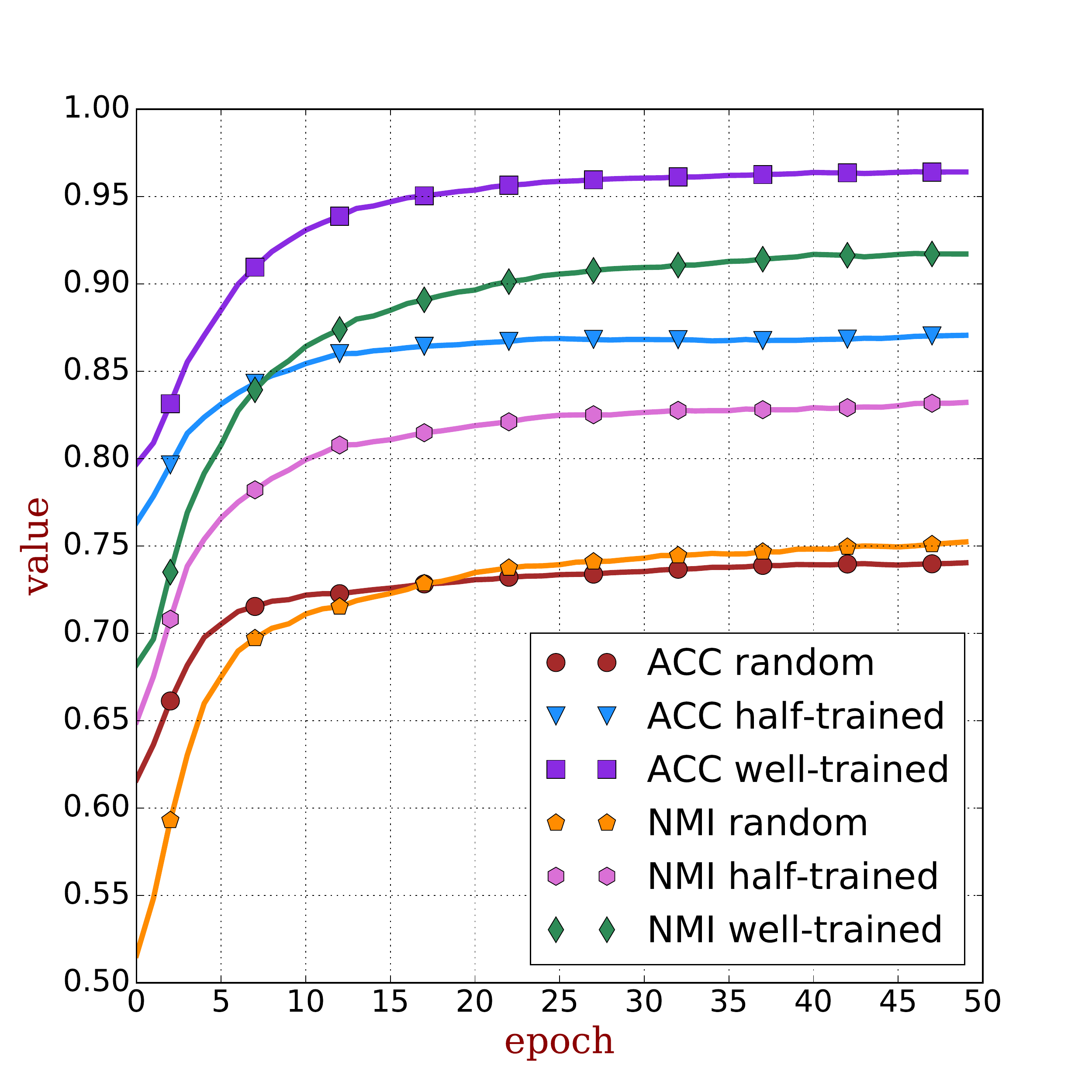}
}
\caption{Ablation studies with respect to the boosting factor $\alpha$, the balance normalization factor $n_j$
and the FCAE initialization models.}\label{fig:ablations}
\end{figure}

To investigate the impact of the FCAE initialization on DBC, we compare the performance of DBC with three different initialization models: 1) the random initialization, 2) the initialization with a half-trained FCAE model, and
3) the initialization with a sufficiently trained FCAE model. The comparison results are shown in
Fig. \ref{fig:ablations}(c). As illustrated in the figure, DBC performs greatly based on all the models
even when the initialization model is randomly distributed. However, if the FCAE model is not sufficiently
trained, the resultant DBC model will be suboptimal.

\section{Conclusions and future works}
\label{sec:conclusion}

In this paper, we proposed FCAE and DBC to deal with image representation learning and image clustering, respectively. Benchmarks on several visual datasets show that our methods can achieve superior performance than the analogous methods.
Besides, the visualization shows that the proposed learning algorithm can implement the idea proposed in
Section \ref{sec:dbc}. Some issues to be considered in the future include: 1) adding suitable constraints on FCAE
to deal with natural images, and 2) scaling the algorithm to deal with large-scale datasets such as the ImageNet dataset.

\section*{Acknowledgement}

This work was supported in part by NNSF of China under grants 61379093, 61602483 and 61603389.
We thank Shuguang Ding, Xuanyang Xi, Lu Qi and Yanfeng Lu for valuable discussions.


\appendix

\section*{Appendix}

\noindent \textbf{A. Derivation of (\ref{eq:gradient-z}).}

We use the chain rule for the deduction. First, we set
\begin{equation}
u_{ij} = 1 + \frac{||z_i-\mu_j||^2}{v}.
\end{equation}
Then it follows that
\begin{equation}
\frac{\partial u_{ij}}{\partial z_i} = \frac{2}{v}(z_i-\mu_j).
\end{equation}
Now set
\begin{equation}
q_{ij} = u_{ij}^{-\frac{1+v}{2}},
\end{equation}
so
\begin{eqnarray}
\frac{\partial q_{ij}}{\partial z_i}&=&\frac{\partial q_{ij}}{\partial u_{ij}}\cdot
\frac{\partial u_{ij}}{\partial z_i} \nonumber \\
   &=& (-\frac{1+v}{2})u_{ij}^{-\frac{3+v}{2}}\cdot \frac{2}{v}(z_i-\mu_j) \nonumber \\
   &=& (-\frac{1+v}{v})u_{ij}^{-1}q_{ij}\cdot (z_i-\mu_j).
\end{eqnarray}
Further, let
\begin{equation}
s_{ij} = \frac{q_{ij}}{\sum_{j'}q_{ij'}}.
\end{equation}
Then we have
\begin{eqnarray}
\frac{\partial s_{ij}}{\partial z_i} &=& \frac{\frac{\partial q_{ij}}{\partial z_i}\cdot \sum_{j'}q_{ij'}
  -q_{ij}\cdot\sum_{j'}\frac{\partial q_{ij'}}{\partial z_i}}{(\sum_{j'}q_{ij'})^2} \nonumber \\
  &=& \frac{(-\frac{1+v}{v})u_{ij}^{-1}q_{ij}\cdot (z_i-\mu_j) \cdot \sum_{j'}q_{ij'} - q_{ij}\cdot\sum_{j'}(-\frac{1+v}{v})u_{ij'}^{-1}q_{ij'}\cdot (z_i-\mu_{j'})}{(\sum_{j'}q_{ij'})^2}  \nonumber \\
  &=& (-\frac{1+v}{v})\frac{q_{ij}}{\sum_{j'}q_{ij'}} \frac{u_{ij}^{-1}\cdot (z_i-\mu_j) \cdot \sum_{j'}q_{ij'}
  - \sum_{j'}u_{ij'}^{-1}q_{ij'}\cdot (z_i-\mu_{j'})}{\sum_{j'}q_{ij'}} \nonumber \\
  &=& (-\frac{1+v}{v})s_{ij} (u_{ij}^{-1}\cdot (z_i-\mu_j) - \sum_{j'}u_{ij'}^{-1}s_{ij'}\cdot (z_i-\mu_{j'})).
\end{eqnarray}
Combine the above expressions to get the required result
\begin{eqnarray}
\frac{\partial L}{\partial z_i}&=&-\sum_{j}\frac{r_{ij}}{s_{ij}}\cdot\frac{\partial s_{ij}}{\partial z_i}\\
  &=& \frac{1+v}{v}\sum_j \frac{r_{ij}-s_{ij}}{u_{ij}}(z_i-\mu_j) \nonumber \\
  &=& \frac{1+v}{v}\sum_j (r_{ij}-s_{ij})\frac{z_i-\mu_j}{1 + ||z_i-\mu_j||^2/v}
\end{eqnarray}

\noindent \textbf{B. Derivation of (\ref{eq:gradient-mu}).}

(\ref{eq:gradient-mu}) can be derived similarly by exchanging $\mu$ and $z$ in the above derivations
of (\ref{eq:gradient-z}).




\end{document}